\documentclass{article}

\usepackage{fullpage}
\usepackage{natbib}
\usepackage{authblk}
\usepackage{graphicx} %
 \usepackage{booktabs} 
\usepackage{bm}
\usepackage[utf8]{inputenc}
\usepackage[T1]{fontenc}
\usepackage{graphicx,amsmath,amsfonts,amssymb,bm,url,color,latexsym,amsthm}
\usepackage{natbib}
\usepackage[colorlinks=true,
linkcolor=blue,
urlcolor=blue,
citecolor=blue]{hyperref}
\usepackage{wrapfig}
\usepackage[small,bf]{caption}
\usepackage[outdir=./]{epstopdf}
\usepackage{subcaption}
\usepackage{bbm}
\usepackage{microtype}
\usepackage[toc, page]{appendix}
\usepackage[ruled]{algorithm2e}
\usepackage{algpseudocode}
\usepackage{xr-hyper}
\usepackage[capitalize]{cleveref}

\usepackage{verbatim}
\usepackage{framed}
\usepackage{comment}
\usepackage{tikz}
\usepackage{fge}
\usepackage{mathdots}
\usepackage{enumitem}

\newtheorem{theorem}{Theorem}
\newtheorem{lemma}{Lemma}
\newtheorem{corollary}{Corollary}
\newtheorem{proposition}{Proposition}
\newtheorem{definition}{Definition}
\newtheorem{assumption}{Assumption}

\newtheorem{example}{Example}

\crefname{assumption}{Assumption}{Assumptions}

\usepackage{array} 
\newcolumntype{M}[1]{>{\centering\arraybackslash}m{#1}}

\newcommand{\mb}{\mathbf}

\newcommand{\bb}{\mathbb}

\newcommand{\eps}{\varepsilon}

\newcommand{\E}{\bb E}

\newcommand{\norm}[2]{\left\| #1 \right\|_{#2}}
\newcommand{\abs}[1]{\left| #1 \right|}
\newcommand{\innerprod}[2]{\left\langle #1,  #2 \right\rangle}

\newcommand{\R}{\bb R}

\newcommand{\Sp}{\bb S}
\newcommand{\Ba}{\bb B}

\renewcommand{\P}{\mathbb{P}}
\newcommand{\EE}{\mathbb E}
\newcommand{\RR}{\mathbb R}

\DeclareMathOperator{\diag}{diag}

\DeclareMathOperator{\rank}{rank}

\DeclareMathOperator{\grad}{grad}

\DeclareMathOperator*{\argmin}{argmin}
\DeclareMathOperator*{\argmax}{argmax}

\DeclareFontFamily{U}{mathx}{\hyphenchar\font45}
\DeclareFontShape{U}{mathx}{m}{n}{
      <5> <6> <7> <8> <9> <10>
      <10.95> <12> <14.4> <17.28> <20.74> <24.88>
      mathx10
      }{}
\DeclareSymbolFont{mathx}{U}{mathx}{m}{n}
\DeclareFontSubstitution{U}{mathx}{m}{n}
\DeclareMathAccent{\widecheck}{0}{mathx}{"71}

\newcommand{\wh}{\widehat}

\newcommand{\wt}{\widetilde}

\newcommand{\bY}{\mb Y}
\newcommand{\bX}{\mb X}
\newcommand{\bA}{\mb A}
\newcommand{\bq}{\mb q}
\newcommand{\ba}{\mb a}

\newcommand{\bI}{\mathbf{I}}
\newcommand{\bP}{\mb P}
\newcommand{\bu}{\mb u}
\newcommand{\bv}{\mb v}
\newcommand{\bD}{\mb D}
\newcommand{\bU}{\mb U}

\newcommand{\by}{\mb y}
\newcommand{\bx}{\mb x}

\newcommand{\bZ}{\mb Z}

\newcommand{\bLambda}{\mb \Lambda}

\newcommand{\bR}{\mb R}

\newcommand{\bM}{\mb M}
\newcommand{\bQ}{\mb Q}

\newcommand{\btheta}{\bs \theta}

\newcommand{\cC}{\mathcal C}

\newcommand{\cI}{\mathcal I}
\newcommand{\cP}{\mathcal P}

\newcommand{\bbO}{\mathbb{O}}

\newcommand{\T}{\top}

\newcommand{\sgn}{\textrm{sgn}}
\newcommand{\tr}{{\rm tr}}
\newcommand{\op}{{\rm op}}

\newcommand{\Var}{{\rm Var}}

\newcommand{\rII}{{\rm II}}
\newcommand{\rIII}{{\rm III}}
\newcommand{\rIV}{{\rm IV}}

\newcommand{\bs}{\boldsymbol}

\newcommand{\GMed}[1]{\operatorname{GM}\!\left(#1\right)}

\title{Federated Independent Component Analysis via Spectral Alignment and Robust Aggregation}

\author[1]{Xin Bing\thanks{xin.bing@utoronto.ca}}
\author[2]{Dian Jin\thanks{dianjin@nus.edu.sg}}
\author[3]{Yuqian Zhang\thanks{yqz.zhang@rutgers.edu}}

\affil[1]{Department of Statistical Sciences, University of
Toronto, }
\affil[2]{Department of Statistics and Data Science, National University of Singapore,}
\affil[3]{Department of Electrical and Computer Engineering, Rutgers University, New Brunswick}

\hypersetup{pdftitle={Federated Independent Component Analysis via Spectral Alignment and Robust Aggregation},pdfauthor={Xin Bing, Dian Jin, Yuqian Zhang}}
\begin{document}

\maketitle

\tableofcontents
\bigskip

\begin{abstract}
This paper studies robust estimation for Independent Component Analysis (ICA) in the federated learning setting, where data are locally distributed across clients and may exhibit substantial heterogeneity. The goal is to recover a common global mixing matrix by aggregating local estimators computed by individual clients. The main difficulty is that local ICA estimators are identifiable only up to sign flips and column permutations and may have highly heterogeneous estimation quality. We propose a novel three-step aggregation method that first aligns the permutations across all local estimators via a particular spectral clustering approach, then aligns the signs within each estimated cluster via another tailored spectral approach, and finally applies the geometric median for robust aggregation. The proposed estimator is shown to remain accurate even when a substantial fraction of local estimators are of low quality or inconsistent, as long as each cluster contains a majority of accurate estimators. This contrasts with its analogue based on simple averaging, whose performance is determined by the worst local estimator. In the homogeneous setting, the robust estimator is also shown to be minimax optimal, up to a logarithmic factor, both when the number of clients remains fixed and when it diverges. The theoretical findings are corroborated by simulation studies and a real-data analysis.
Code is available at \url{https://github.com/Jindiande/Federated_ICA_experiments}.
\end{abstract}

\section{Introduction}

Independent Component Analysis (ICA) \citep{Hyvarinen97} is a classical framework for recovering latent source signals from observed mixtures. Beyond recovering the latent sources themselves, many applications are particularly interested in the underlying mixing structure, as it characterizes how the latent sources jointly generate the observed data. Estimating this structure is therefore of both statistical and scientific interest and has made ICA a useful tool in a broad range of applications, including blind source separation \citep{cardoso1993blind,choi2005blind,correa2007performance,sawada2019review}, biomedical signal processing \citep{jung2001imaging,ungureanu2004independent,james2005independent}, and image steganography \citep{Gonzalez2001watermarking,bounkong2003ica,kumar2010performance}, just to name a few. 

In the standard ICA model, each observation $\mb y_i\in\mathbb R^r$ is written as $\mb y_i=\mb A\mb x_i$, where $\mb A\in\mathbb R^{r\times r}$ is the mixing matrix and $\mb x_i\in\mathbb R^r$ contains mutually independent, non-Gaussian latent components. The latter distributional assumption on $\bx_i$ ensures model identifiability, and a large body of ICA methods recover $\mb A$ by exploiting the non-Gaussianity of the latent components, for example through excess kurtosis or skewness \citep{hyvarinen1997fast}. Importantly, even under these conditions, the matrix $\mb A$ is identifiable only up to sign flips and permutations of its columns \citep{comon1994independent,hyvarinen2000independent}. We henceforth refer to this inherent non-identifiability as the {\em signed-permutation ambiguity}. We provide a brief review of ICA algorithms in \cref{sec_method_ICA}.

In many modern settings, however, the data used for ICA are inherently distributed across multiple clients, institutions, or devices and cannot be pooled at a central server due to privacy, communication, or data-ownership constraints. This naturally leads to a federated learning setting \citep{mcmahan2017communication}, where each client observes only its own local mixed signals, while the central server aims to recover a common global mixing structure without accessing the raw data. Unlike classical distributed estimation settings that often assume identically distributed local data, federated learning commonly involves heterogeneous, non-i.i.d. data across clients. Federated estimation for ICA therefore presents two fundamental and coupled challenges: cross-client alignment due to the signed-permutation ambiguity inherent in ICA, and robust aggregation under heterogeneous local data and estimation quality. We discuss these two challenges in turn.

The signed-permutation ambiguity inherent in ICA poses a fundamental challenge for aligning local estimators across clients. To see this, even when two clients target the same mixing matrix $\bA^\star$, their local estimators may be close to its columns only under different and unknown sign flips and column permutations. More generally, when two clients return local estimators $\bA^{(1)}$ and $\bA^{(2)}$ to be aggregated at the central server, they may recover different subsets of the columns of the common matrix $\mb A^\star$, in addition to the aforementioned signed permutations. Consequently, simple averaging-based methods such as FedAvg \citep{mcmahan2017communication,khaled2020tighter,haddadpour2019convergence} are generally inapplicable. Successful aggregation first requires identifying which local columns across different clients correspond to the same column of $\mb A^\star$ and then resolving their sign ambiguity. This challenge becomes increasingly difficult as the number of clients $K$ grows, since the number of local columns that must be aligned also increases.

Even after resolving the signed-permutation ambiguity, data heterogeneity across clients can render averaging-based aggregation procedures highly sensitive to data heterogeneity. In the federated ICA setting considered here, each client $k$ observes a local mixing matrix whose columns form a client-specific subset of those in a common global mixing matrix $\mb A^\star$. Moreover, the distributions of the latent sources $\bx^{(k)}$, noise levels, and local sample sizes may all vary across clients. Such heterogeneity can lead to substantial variation in estimation quality across clients, making simple averaging particularly sensitive to poorly estimated or even inconsistent local components.

Various robust aggregation methods have been developed for distributed and federated learning, beginning with \citet{feng2015distributed}, who proposed a distributed robust learning framework based on the geometric median with theoretical guarantees, followed by substantial advances in \citep{blanchard2017machine,chen2017distributed,yin2018byzantine,ghosh2019robust,yin2019defending,Pillutla2022Robust,Su2024Global}. More recently, \citet{tian2024towards} studied unsupervised federated learning for heterogeneous mixture models in the presence of adversarial clients, while \citet{tian2026robust} developed robust unsupervised multi-task and transfer learning methods for Gaussian mixture models, including alignment procedures to address label ambiguity across tasks. Another related work on Byzantine-tolerant distributed learning of finite mixture models is \citet{zhang2026byzantine}. They consider a setting in which a fraction of local clients may return arbitrarily erroneous, potentially adversarial, estimates. Their method filters unreliable local mixture estimates by comparing their estimated densities and then combines the remaining estimates in a way that accounts for label switching across machines. These approaches, however, are not directly applicable to ICA, where local estimators are subject to both permutation and sign ambiguities.

From a communication perspective, many existing federated learning methods require multiple rounds of communication between local clients and the central server to exchange parameter estimates or gradients. Such iterative procedures are generally less communication-efficient than the {\em one-shot} federated learning setting, which we focus on in this paper. The latter refers to the setting where each client communicates its locally computed estimator to the central server only once, and the server performs aggregation without further client-server interaction \citep{liu2024fedlpa,zhao2023non,jhunjhunwala2024fedfisher,tang2024fusefl,rjoub2022one,tao2024task,wang2024one}. %

Related one-shot distributed estimation methods have been developed for principal component analysis (PCA) \citep{fan2019distributed,chen2022analog}. However, the aggregation problem in PCA is fundamentally different from that considered here. In distributed PCA, the target is typically the principal subspace, which is identifiable without fixing a particular orthonormal basis and can therefore be aggregated through rotation-invariant representations such as projection matrices. In contrast, ICA aims to recover the individual mixing directions, which are identifiable only up to signed permutations. Thus, local ICA estimators must first be matched and sign-aligned across clients before they can be meaningfully aggregated, a component-wise alignment problem that does not arise in distributed PCA.

In this paper we develop  a robust one-shot  aggregation procedure for estimating the common mixing matrix $\mb A^\star$ from heterogeneous local ICA estimators. At each client $k$, the local ICA model involves an unknown subset of columns of $\mb A^\star$, and the client returns an estimator of these columns, up to an unknown signed permutation and potentially substantial estimation error. Using these local estimators, the central server resolves their signed-permutation ambiguity and robustly aggregates the aligned estimators, without any further communication with the clients. Our main contributions are twofold.

\begin{itemize}[leftmargin = 1.5em]
    \item \textbf{A three-stage robust aggregation algorithm.} 
    We propose Spectral-Robust-Federated ICA (SRF-ICA), a robust one-shot aggregation method tailored to the signed-permutation ambiguity and client heterogeneity in federated ICA; see \cref{sec:problem_method}. The central server starts by pooling the columns of the local estimators across clients and constructs a sign-invariant affinity matrix using their absolute pairwise inner products. This construction allows local atoms corresponding to the same latent component to be identified as such, even when they have opposite signs relative to their respective true atoms. Based on this affinity matrix, we apply spectral $k$-means clustering to partition the local atoms into $r$ estimated clusters, thereby resolving the permutation ambiguity; see \cref{sec_method_kmeans}.  Because this clustering step is sign-invariant, atoms with opposite signs may belong to the same cluster. We therefore introduce in \cref{sec_method_sign} a within-cluster sign-alignment procedure based on the leading singular vector of the atoms in each cluster. 
    Once the signed-permutation ambiguity is resolved, various aggregation procedures can in principle be applied. In the final step of SRF-ICA, we propose to aggregate the sign-aligned atoms within each cluster at the central server using the geometric median, as described in \cref{sec_method_GM}. Such aggregation provides robustness to heterogeneous local estimation quality.

    \item \textbf{Theoretical guarantees for spectral alignment and robust aggregation.}
    We establish theoretical guarantees in \cref{sec_theory} for all three stages of SRF-ICA. For the spectral clustering stage, we show in \cref{lem:spectral_kmeans} of \cref{sec_theory_km} that the sign-invariant affinity matrix preserves the true clustering structure up to a perturbation controlled by the local estimation errors, which in turn yields a cluster-wise misclustering bound for spectral $k$-means. For the sign-alignment stage, we show in \cref{prop:sign_alignment} that the misclustering errors from the preceding step have a limited effect on sign alignment, and that every sufficiently accurate and correctly clustered atom is aligned to a common orientation. Together, these two results allow us to transfer the estimation-error quantiles under the oracle signed-permutation alignment to those of the atoms aligned using the estimated permutations and signs, as established in \cref{lem:mis_class} of \cref{sec_theory_GM}. Combined with a robustness bound for the geometric median characterized by the estimation-error quantiles, our main result is stated in \cref{thm:recover} of \cref{sec:combined}: the estimation error of the final SRF-ICA estimator is controlled by majority quantiles of the local ICA estimation errors, rather than by their average or worst-case error. Consequently, SRF-ICA remains accurate under substantial client heterogeneity as long as, for each global component, more than half of its local estimators are reliable.  In contrast, the error of the simple averaging estimator, even after resolving the signed-permutation ambiguity, can be determined by the largest local estimation error; see \cref{thm:kmeans_center}. We further specialize to specific ICA estimators in \cref{cor:dictionary} of \cref{sec_theory_ICA} and derive explicit rates that account for heterogeneous sample sizes across clients. Beyond robustness, we show in \cref{thm:optimal_aggre_rate} of \cref{sec:optimal_aggregate_rate_ica} that SRF-ICA is also minimax optimal, up to logarithmic factors, in the homogeneous ICA setting, where clients have comparable sample sizes and share the same target mixing matrix. A key technical lemma underlying this result  establishes a new concentration inequality for the geometric median of independent random vectors that are only approximately spatially centered and subject to perturbations, which is of independent interest.
\end{itemize}
In \cref{sec:exp}, we conduct simulation studies and a real data analysis to corroborate our theoretical findings, demonstrating the robustness of SRF-ICA under client heterogeneity and its statistical optimality under homogeneity. All the proofs are deferred to the Appendix.\\

\noindent{\em Notation.} Throughout the paper, bold lowercase letters such as $\mb x$ denote vectors, while bold uppercase letters such as $\mb X$ denote matrices. For a matrix $\mb X$, its $(i,j)$-th entry is denoted by $x_{ij}$, while $\mb x_j$ denote its  $j$-th column. For any positive integer $n\in\mathbb N$, we define $[n]=\{1,2,\ldots,n\}$. For a vector $\mb x$, we use  $\|\mb x\|$ and $\|\mb x\|_4$ to denote its $\ell_2$- and $\ell_4$-norms. For a matrix $\mb X$,  we use $\norm{\mb X}{\op}$ and $\norm{\mb X}{F}$ to denote its operator and Frobenius norms, respectively. We write $\|\bX\|_{2,\infty} = \max_j\|\bx_j\|$ for the maximum column-wise $\ell_2$-norm. We use $\Sp^{d-1}$ to denote the set of all vectors in $\RR^d$ with unit $\ell_2$ norm. We use $\bbO_{d_1,d_2}$ to denote the set of  all $d_1\times d_2$ matrices with orthonormal columns. 
  For two nonnegative sequences $a_n$ and $b_n$, we write $a_n\lesssim b_n$ if $a_n\le Cb_n$ for some constant $C>0$, and $a_n\asymp b_n$ if both $a_n\lesssim b_n$ and $b_n\lesssim a_n$.

\section{Federated ICA: problem setup and aggregation challenges}\label{sec:background}

We consider a federated learning environment for Independent Component Analysis (ICA) where the system comprises $K$ clients and a central server.   Each client $k\in [K]$ has access to its local data stored in an $r\times N_k$ matrix $\mb Y^{(k)} = (\by^{(k)}_1, \ldots, \by^{(k)}_{N_k})$ which is assumed to be generated from
\begin{align}\label{def:model}
    \mb Y^{(k)}  = \mb A^{(k)\star} \mb X^{(k)\star}.
\end{align}
The matrix $\mb A^{(k)\star} \in \bb R^{r\times r_k}$ contains a subset of columns of a global matrix $\mb A^\star\in \mathbb R^{r\times r}$, where $r_k \le r$.
Both the subset and its size $r_k$ are allowed to vary across clients.
The matrix $\mb X^{(k)\star} \in \bb R^{r_k\times N_k}$ represents the source signals in each local dataset, which may have different distributions and sample sizes $N_k$. Without loss of generality, we can assume $\rank(\mb A^{(k)\star}) = r_k$, as otherwise $r_k$ can be reduced to satisfy this condition. We further assume throughout our theoretical analysis that the global mixing matrix 
\begin{equation}\label{ass_orthogonal}
    \mb A^\star\in\bbO_{r,r}
\end{equation}
so that each local mixing matrix $\mb A^{(k)\star}$, consisting of a subset of its columns, belongs to $\bbO_{r,r_k}$. When $r_k=r$ for all clients, this condition is without loss of generality under the standard whitening preprocessing for ICA \citep{hyvarinen2000independent,jin2021unique}. For the more general setting with $r_k<r$, we impose orthogonality to simplify the theoretical analysis. The proposed aggregation procedure itself does not require this condition and can be applied to more general mixing matrices.

The goal of federated learning in this context is for each client to compute a local estimator of the mixing matrix $\mb A^{(k)\star}$, after which the central server aggregates these estimators to obtain a robust estimator of the global matrix $\mb A^\star$. This global estimator can then be used in downstream tasks, such as better recovering each individual source signal $\mb X^{(k)\star}$. We start by first reviewing the existing ICA algorithms which can be used by each client to obtain local estimators. 

\subsection{Review on the FastICA algorithm}\label{sec_method_ICA}
In the ICA literature, the primary objective under model \eqref{def:model} is to recover the mixing matrix $\mb A^{(k)\star}$ and to extract the original source signals $\mb X^{(k)\star}$. 
However,  for general matrix decomposition of the form in \eqref{def:model}, one can only recover $\bA^{(k)\star}$ up to an $r_k \times r_k$ non-singular matrix $\bR$, since $(\bA^{(k)\star} \bR)(\bR^{-1} \bX^{(k)\star})$ yields the same model. The premise of ICA is that when the components of $\bX^{(k)\star}$ are independent realizations from non-Gaussian distributions, the general non-singularity ambiguity can be reduced to a smaller set consisting of signed permutation matrices, that is, sign flipping and permutation of the columns. 

There exists a vast literature on ICA, varying in how the non-Gaussianity of $\mb X^{(k)\star}$ is modeled. We refer to \cite{hyvarinen2000independent} for a comprehensive review. Among the many ICA algorithms, FastICA is arguably the most popular, relying on non-zero excess kurtosis as a measure of non-Gaussianity to recover the independent components. %
Its symmetric version \citep{hyvarinen2000independent} estimates  $\bA^{(k)\star} \in \bbO_{r,r_k}$ with orthonormal columns by solving the following optimization problem:
\begin{align}\label{eq:ICA_form}
    \wt{\mb A}^{(k)}~ :=~ \argmax_{\bQ \in \bbO_{r,r_k}} ~ \sum_{i=1}^{N_k}\bigl\|\bQ^\top \by^{(k)}_i\bigr\|_4^4. 
\end{align}
In practice, this is typically done via a fixed-point iteration method \citep{hyvarinen1997fast}. 
Despite its popularity, its theoretical guarantees remain limited, primarily due to the non-convex orthogonality constraint $\bQ \in \bbO_{r,r_k}$. Recently, \cite{auddy2023Large,bing2025optimal} showed that deflation-type FastICA, which solves a variant of the above problem one column at a time, provably estimates the columns of $\bA^{(k)\star}$, up to a signed permutation, at the minimax-optimal error rate. More details on these results and the required modeling conditions are provided in \cref{sec_theory_ICA}.  Beyond FastICA, other approaches, such as tensor-based methods \citep{anandkumar2014analyzing,ma2016polynomial}, also come with theoretical guarantees for global recovery and are computationally feasible. Nevertheless, these methods do not always guarantee statistical optimality. 

Both of our method and most of the theory we derive apply to general ICA estimators from each client. Specialization to the FastICA-type estimators is only considered in \cref{sec:optimal_aggregate_rate_ica}.

\subsection{Challenges of aggregation over clients}

Adopting any of the FastICA algorithms mentioned above, we assume that  each client $k\in [K]$  has computed its local estimator  
$$
\wt{\mb A}^{(k)} = \bigl[\wt\ba_1^{(k)},\ldots, \wt\ba_{r_k}^{(k)}\bigr]\in \RR^{r\times r_k}
$$
with $\|\wt\ba_i^{(k)}\| = 1,$ for each $i\in [r_k]$. Note that we do not necessarily require the columns of the local estimator to be orthogonal. Since $\wt{\mb A}^{(k)}$ is expected to be close to the true $\mb A^{(k)\star}$  up to some  $r_k\times r_k$ signed permutation matrix, we define the permutation map that  best aligns its columns as 
\begin{equation}\label{def_sigma_star}
    \sigma^\star(k,\cdot)
    ~\in~ 
    \argmin_{\sigma: [r_k]\to [r]}~ 
    \sum_{i=1}^{r_k}
    \min_{\xi\in\{\pm1\}}
    \bigl\|
        \wt{\mb a}^{(k)}_i
        -
        \xi\,\mb a^\star_{\sigma(i)}
    \bigr\|^2 
\end{equation}
as well as the best sign alignment of 
each $i\in [r_k]$ as  
\begin{equation}\label{def_xi_star}
    \xi^\star_{(k,i)} = \argmin_{\xi\in\{\pm1\}}~ 
    \bigl\|
        \wt{\mb a}^{(k)}_i
        -
        \xi\,\mb a^\star_{\sigma^\star(k,i)}
    \bigr\|^2.
\end{equation}
With the above definition, we have
\begin{equation}\label{ambiguity}
    \wt{\mb A}^{(k)} \approx  \bA^\star \bZ^{(k)} \bs\Xi^{(k)}.
\end{equation}
where $\bZ^{(k)}$ is an $r\times r_k$ assignment matrix with its $i$-th column equal to the canonical basis vector $\mb e_{\sigma^\star(k,i)}$ in $\RR^r,$
and $\mb\Xi^{(k)}$ is an $r_k\times r_k$ diagonal matrix with its $i$-th diagonal entry equal to $\xi^\star_{(k,i)}$.

In our one-shot federated learning setting, we aim to pool local estimators $\wt{\mb A}^{(k)}$ across all $k\in [K]$ clients in the central server to obtain improved estimation of $\bA^*$. Since different clients target different columns of $\bA^\star$, we first note that  $\sigma^\star(k,\cdot)$ in \eqref{def_sigma_star}, for $k\in [K]$, induces a partition of all $R:=  \sum_{k=1}^K r_k$ local estimators into $r$ clusters. Specifically, let
$$
\cI:= \{
(k,i): k\in [K], i \in [r_k]
\}
$$ denote the index set of all columns of the local estimators with $|\cI| = R$. We let
\begin{equation}\label{def_clusters_star}
    \cC_a^\star
    :=
    \left\{
    (k, i) \in \mathcal I : \sigma^\star(k, i) = a
    \right\},\quad a\in[r],
\end{equation}
denote the indices among them corresponding to estimators of $\ba_a^\star$, the $a$-th column of $\bA^\star$, and write  $n_a := |\cC_a^\star|$ for its cardinality. Clearly, the larger $n_a$ is, the more clients are able to estimate $\ba_a^\star$. Further define 
\begin{align}\label{def_barn}
    n_{\max} = \max_{a\in [r]} n_a, \qquad n_{\min} = \min_{a\in [r]} n_a,\qquad \bar n = R/r.
\end{align}
We assume that all columns  in  $\mb A^{\star}$ are estimated at least by one client:
\begin{equation}\label{ass_cover}
    \bigcup_{(k,i)\in \mathcal I}\sigma^\star(k,i) = [r].
\end{equation}
Otherwise,  only the columns of $\mb A^\star$ that are contained in $\{\sigma^\star(k,i): (k,i)\in \mathcal I\}$ can be recovered.  Under \eqref{ass_cover}, there are still two main challenges to combine the local estimators into a single $r \times r$ matrix: 
\begin{enumerate}[noitemsep]
    \item[(1)] the assignment matrix $\bZ^{(k)}$ and the sign matrix $\mb\Xi^{(k)}$ in \eqref{ambiguity} are unknown and may differ across clients, rendering simple aggregation methods such as averaging meaningless; and 
    \item[(2)] the quality of local estimators varies across clients, necessitating robust aggregation other than averaging, after adjusting both ambiguities. 
\end{enumerate}
Our proposed procedure in the next section addresses these two challenges separately. We conclude this section by illustrating the first challenge with a simple example, which will also serve as a running example throughout our discussion of the proposed methodology. 

\begin{example}[Signed permutation ambiguity]\label{example}
Consider the case $r=3$ and let the global matrix be
$
\mb A^\star = [\ba_1^\star ,\ba_2^\star ,\ba_3^\star ]\in\bbO_{r,r}.
$
Suppose there are three clients with their true local matrices as
\[
\mb A^{(1)\star}=[\ba_1^\star ,\ba_2^\star ],\qquad
\mb A^{(2)\star}=[\ba_2^\star ,\ba_3^\star ],\qquad
\mb A^{(3)\star}=[\ba_1^\star ,\ba_3^\star ].
\]
Under the ICA model, each local mixing matrix is identifiable only up to a signed permutation. Therefore, their local estimators may take the form
\[
\wt{\mb A}^{(1)}\approx [\ba_2^\star ,-\ba_1^\star ],\qquad
\wt{\mb A}^{(2)}\approx [-\ba_3^\star ,\ba_2^\star ],\qquad
\wt{\mb A}^{(3)}\approx [\ba_3^\star ,\ba_1^\star ].
\]
These three estimators can each recover two columns of the global matrix $\mb A^\star$, but up to {\em different} $2\times 2$ signed permutation matrices. To illustrates the ambiguity arising from both sign and permutation, first note that the same global atom may appear with different signs across clients: for instance, $\ba_1^\star $ appears as $-\ba_1^\star $ in client $1$ but as $\ba_1^\star $ in client $3$, and similarly $\ba_3^\star $ appears with opposite signs in clients $2$ and $3$. Second, even after ignoring signs, the order of the local columns is inconsistent across clients. Consequently, there is no canonical column-wise correspondence among the local estimators, and simple averaging across clients is not meaningful before resolving the signed permutation ambiguity.
\end{example}

\section{The proposed method}\label{sec:problem_method}

In this section we propose a three-step algorithm that aggregates local estimators from all clients. The first step resolves the permutation ambiguity via  a spectral clustering on a chosen affinity matrix which takes into account the sign ambiguity. The sign ambiguity is resolved in the second step where we use another tailored spectral approach based on the previous clustering results. After resolving both ambiguities, we propose to aggregate within each cluster in the last step by using the geometric median.
  
\subsection{Resolve the permutation ambiguity}\label{sec_method_kmeans}

To resolve the permutation ambiguity among the local estimators, we first concatenate their columns into an $r\times R$ matrix
\begin{equation}
    \wt{\mb A}_c := \bigl[\wt{\mb A}^{(1)},\ldots,\wt{\mb A}^{(K)}\bigr]
    ~ \overset{\eqref{ambiguity}}{\approx} ~
    \bigl[\mb A^{\star}\bZ^{(1)}\mb\Xi^{(1)},\ldots,\mb A^{\star}\bZ^{(K)}\mb\Xi^{(K)}\bigr]
    ~=:
    \bA^\star \bZ \mb\Xi,
    \label{def_tilde_A_c}
\end{equation}
where $\mb\Xi = \diag(\mb\Xi^{(1)},\ldots,\mb\Xi^{(K)})$ is the combined $R\times R$ diagonal sign matrix encoding the sign ambiguity of the local estimators, and $\bZ = [\bZ^{(1)},\ldots,\bZ^{(K)}]$ is the $r\times R$ concatenated assignment matrix encoding the permutation ambiguity. Resolving the permutation ambiguity therefore amounts to estimating $\bZ$, or equivalently, its induced partition $\sigma^\star$ defined in \eqref{def_sigma_star}.

If there were no sign ambiguity, i.e., $\mb\Xi=\bI_R$, the columns of $\wt{\mb A}_c$ could be directly clustered to recover the assignment structure encoded by $\bZ$. In the presence of general $\mb\Xi$, however, local estimators corresponding to the same column of $\mb A^\star$ may have opposite signs and hence appear as two distinct groups. Consequently, direct clustering may produce as many as $2r$ clusters even in the noiseless case; see \cref{example}. To eliminate this nuisance sign ambiguity while preserving the assignment structure, we exploit the fact that sign flips do not affect the absolute values of pairwise inner products. This motivates a spectral preprocessing step based on a sign-invariant affinity matrix, followed by clustering to estimate $\sigma^\star$.

Concretely, we propose to first form the $R\times R$ Gram-type matrix $\bM$ that satisfies
\begin{equation}\label{def_M_mat}
\sqrt{\bar n}~ \mb M:=  \bigl|\wt{\mb A}_c^\top \wt{\mb A}_c\bigr|  \approx  \abs{\mb\Xi\bZ^\T \bA^{\star\top}\bA^\star\bZ\mb\Xi} =  \abs{\mb\Xi\bZ^\T \bZ\mb\Xi} = \bZ^\T \bZ .  
\end{equation}
Here $\sqrt{\bar n}$ is a normalization scalar and the absolute values in the above display are defined entry-wise. 
The term $\bZ^\T \bZ$ in \eqref{def_M_mat} reveals the clustering structure. By noting that it has rank $r$, we propose to estimate it by the best rank-$r$ approximation $ \bM^{(r)}$ of $\bM$. Specifically, by eigen-decomposing 
$
    \bM = \sum_{j=1}^R d_j \bu_j \bu_j^\T,
$
we have 
\begin{align}\label{def:M_r}
        \bM^{(r)} = \bU_r \bD_r \bU_r^\T \in \RR^{R\times R}
\end{align}
with  $\mb U_r = [\bu_1,\ldots, \bu_r] \in \bbO_{R,r}$ and $\bD_r = \diag(d_1,\ldots,d_r)$. In principle, one could apply the $k$-means algorithm to partition the columns of $\bM^{(r)}$ into $r$ clusters. However, since $R$ can be large, we use a simple trick to reduce both computational and memory costs: $k$-means clustering of the columns of $\bM^{(r)}$ is equivalent to clustering the columns of 
\[
\mb Q := \mb U_r^\top \mb M  = \bD_r \bU_r^\T \in \mathbb R^{r \times R}.
\]
We provide its proof in \cref{lem:rank_kmeans} for completeness. 
Denoting by $\bq_{(k,i)}$ the column of $\bQ$ corresponding to the $i$-th atom of the $k$-th client, we propose to solve the following $k$-means problem:
\begin{align}\label{kmeans_spec}
    \wt{\bs \theta}_1,\ldots,\wt{\bs \theta}_r,\wt \sigma
    ~ = \argmin_{\substack{\btheta_1,\ldots,\btheta_r\in\bb R^r,~\sigma:\mathcal I \to [r]}}
    \sum_{(k,i)\in \mathcal I}
    \bigl\|\bq_{(k,i)}-\btheta_{\sigma(k,i)}\bigr\|^2.
\end{align}
The resulting clustering map $\wt \sigma:\cI\to [r]$ yields the clusters $\wt\cC_1,\ldots, \wt\cC_r$, given by 
\begin{align}\label{def_clusters}
       \wt \cC_a
   =
   \left\{
        (k, i) \in \mathcal{I} : \wt\sigma(k,i) = a
    \right\},
   \qquad \forall ~ a\in [r].
\end{align}
We analyze the misclustering error rate of $\wt\sigma$ in \cref{lem:k_means,lem:spectral_kmeans} of \cref{sec_theory_km}. The following continuation of \cref{example} illustrates the above spectral clustering procedure.

\begin{example}\label{example_sign}
    Continuing \cref{example}, we have the concatenated matrix in \eqref{def_tilde_A_c} as
    $$
        \wt \bA_c = \bigl[\wt \ba_1^{(1)}, \wt \ba_2^{(1)},\wt \ba_1^{(2)}, \wt \ba_2^{(2)},\wt \ba_1^{(3)}, \wt \ba_2^{(3)}\bigr]  ~ \approx  ~  \bigl[\ba_2^\star ,-\ba_1^\star, -\ba_3^\star ,~ \ba_2^\star, ~ \ba_3^\star ,~ \ba_1^\star \bigr]
    $$ so that 
    \[
        \wt{\mb A}_c^\top \wt{\mb A}_c  \approx \begin{bmatrix}
            1 & 0 & 0 & 1 & 0 & 0 \\
            0 & 1 & 0 & 0 & 0 & -1\\
            0 & 0 & 1 & 0 & -1 & 0 \\
            1 & 0 & 0 & 1 & 0 & 0 \\
            0 & 0 & -1 & 0 & 1 & 0\\
            0 & -1 & 0 & 0 & 0 & 1
        \end{bmatrix},\qquad   \bigl|\wt{\mb A}_c^\top \wt{\mb A}_c\bigr|  \approx \begin{bmatrix}
            1 & 0 & 0 & 1 & 0 & 0 \\
            0 & 1 & 0 & 0 & 0 & 1\\
            0 & 0 & 1 & 0 & 1 & 0 \\
            1 & 0 & 0 & 1 & 0 & 0 \\
            0 & 0 & 1 & 0 & 1 & 0\\
            0 & 1 & 0 & 0 & 0 & 1
        \end{bmatrix}.
    \]
    Applying the spectral clustering to $|\wt{\mb A}_c^\top \wt{\mb A}_c|$ gives 
    three clusters $\wt\cC_1 = \{(1,2), (3,2)\}$, $\wt \cC_2 = \{(1,1), (2,2)\}$ and $\wt \cC_3 = \{(2,1), (3,1)\}$, where, for each pair $(k,i)$, $k$ denotes the client index and $i$ denotes the column index within client $k$.  Note that without taking absolute values, the rows of $\wt{\mb A}_c^\top \wt{\mb A}_c$ form five clusters rather than three. In terms of the local estimators within each of the three clusters, we have  
    \[
        \bigl[\wt{\mb a}^{(1)}_1, ~ \wt \ba_2^{(2)}\bigr] \approx \bigl[\ba_2^\star, ~\ba_2^\star\bigr],\qquad  \bigl[\wt{\mb a}^{(1)}_2, ~ \wt \ba_2^{(3)} \bigr] \approx \bigl[-\ba_1^\star, ~ \ba_1^\star\bigr],\qquad  \bigl[\wt{\mb a}^{(2)}_1, ~ \wt \ba_1^{(3)} \bigr] \approx \bigl[-\ba_3^\star, ~ \ba_3^\star\bigr].
    \]
    Thus, each cluster contains estimators of the same true atom, but possibly with different signs. Directly aggregating the estimators within each cluster without first aligning their signs can therefore lead to cancellation and result in a poor estimate. This motivates the sign-alignment procedure introduced in the next section.
\end{example}

\subsection{Resolve the sign ambiguity}\label{sec_method_sign}
After resolving the permutation ambiguity via \eqref{def_clusters}, a sign ambiguity remains within each estimated cluster, as illustrated by the simple  \cref{example_sign}. More generally, for $\wt\cC_a$ given in \eqref{def_clusters}, collect its local estimators  as
\begin{equation}\label{sign_ambiguity}
    \wt{\mb A}_{\wt\cC_a} ~:=~
    \bigl[ \wt{\mb a}^{(k)}_i : (k,i)\in\wt\cC_a \bigr]
    ~\approx~
    \bigl[ \xi^\star_{(k,i)}~  \ba_{a'}^\star : (k,i)\in\wt\cC_a \bigr]
    \in \RR^{r\times |\wt\cC_a|}.
\end{equation}
When $\wt\sigma$ is close to $\sigma^\star$, the approximation above is expected to hold for some $a'\in[r]$, where $\xi^\star_{(k,i)}$, defined in \eqref{def_xi_star}, denotes the oracle sign associated with the local estimator $\wt\ba_i^{(k)}$.
While the use of absolute inner products in constructing the affinity matrix $\mb M$ ensures that local estimators with opposite signs are assigned to the same cluster, this sign invariance creates an additional issue for subsequent aggregation. In particular, directly aggregating the columns of $\wt{\mb A}_{\wt\cC_a}$ may lead to cancellation between estimators close to $\mb a^\star_{a'}$ and those close to $-\mb a^\star_{a'}$. We therefore align the signs of the local estimators within each estimated cluster before aggregation.

A key observation for sign alignment is that the right-hand side of \eqref{sign_ambiguity} is a rank-one matrix whose leading left singular vector is parallel to $\ba^\star_{a'}$; see, also, \cref{example_sign}. This suggests using the leading singular vector of the left-hand side of \eqref{sign_ambiguity}, denoted by $\wt{\mb u}_a \in \Sp^{r-1}$, to align the signs of its columns.  Specifically, for each local estimator $\wt{\mb a}^{(k)}_i$ with $(k,i)\in\wt\cC_a$, we define
its sign-aligned version as
\begin{equation}\label{def:sign_aligned_atom}
    \widecheck{\mb a}^{(k)}_i
    ~ := ~ 
    \operatorname{sgn}
    \bigl(
        \langle
            \wt{\mb u}_a,\wt{\mb a}^{(k)}_i
         \rangle
    \bigr)~ 
    \wt{\mb a}^{(k)}_i,
\end{equation}
with $\operatorname{sgn}(0) \equiv 1$. In \cref{prop:sign_alignment}, we show that the above sign-aligned estimator 
successfully resolves the sign ambiguity by removing the dependence on $\xi^\star_{(k,i)}$ in \eqref{sign_ambiguity} such that $\widecheck{\mb a}^{(k)}_i$ is close to $\ba^\star_{a'}$ for all $(k,i)\in \wt\cC_a$. For instance, in \cref{example_sign}, for $a=1$, if we have $\wt \bu_a \approx \ba_1^\star$, then 
$$
    \widecheck \ba_2^{(1)} \approx  
    \operatorname{sgn}\bigl(\langle
            \ba_1^\star,\wt{\mb a}^{(1)}_2
         \rangle\bigr) \wt \ba_2^{(1)}\approx \ba_1^\star,
\qquad \widecheck \ba_2^{(3)} \approx \operatorname{sgn}\bigl(\langle
            \ba_1^\star,\wt{\mb a}^{(3)}_2
         \rangle\bigr)  \ba_2^{(1)}\approx \ba_1^\star,
$$
which have the same sign after alignment.

Our aggregation procedure in the next subsection thereby applies to the sign-aligned estimators $ \widecheck{\mb a}^{(k)}_i$ in \eqref{def:sign_aligned_atom} for all $(k,i)\in \wt\cC_a$ and $a\in [r]$.

\subsection{Geometric median for robust aggregation}\label{sec_method_GM}

Once resolving the signed permutation ambiguity, any aggregation procedure can in principle be applied to the sign-aligned local estimators in \eqref{def:sign_aligned_atom} within each cluster to form a single estimator of $\bA^\star$. The simplest is to average the local estimators within each estimated cluster, that is, for each $a\in [r]$,  
\begin{equation}\label{def_centers}
    \bar{\mb a}_a
     ~=~ {1\over |\wt\cC_a|}\sum_{(k,i)\in\wt\cC_a} \widecheck{\mb a}^{(k)}_i  ~=~ 
    \argmin_{\mb q\in\RR^r}
    \sum_{(k,i)\in\wt\cC_a}
    \bigl\|
        \mb q-\widecheck{\mb a}^{(k)}_i
    \bigr\|^2 .
\end{equation}
This leads to the averaged estimator $\bar \bA = [\bar \ba_1, \ldots, \bar \ba_r]$, whose estimation error is analyzed in \cref{thm:kmeans_center} of \cref{sec_theory_GM}. However, in the presence of heterogeneity, averaging is not robust to corrupted local estimators.

To improve robustness, we propose using the geometric median (GM) within each cluster instead of simple averaging. Specifically, for each $a\in[r]$, we define:
\begin{equation}\label{def_GM}
    \wh \ba_a
    ~ \in ~ 
    \argmin_{\mb q\in\RR^r}
    \sum_{(k,i)\in\wt\cC_a}
    \bigl\|
        \mb q-\widecheck{\mb a}^{(k)}_i
    \bigr\|.
\end{equation}
The GM is a well-known robust estimator due to its optimal breakdown point of $0.5$, meaning that up to 50\% of the local estimators  $\widecheck{\mb a}^{(k)}_i$ in $\wt \cC_a$ can be arbitrarily corrupted without substantially affecting the quality of the resulting estimator. See details in \cref{def:p_quantile} and \cref{lem:gm_robust}. In contrast, the averaged estimator $\bar \ba_a$ in \eqref{def_centers} has a breakdown point of zero, implying that even a single bad estimator in $\widecheck{\mb a}^{(k)}_i$  can severely distort the estimate. The robustness of the GM is formally established in \cref{thm:recover} of \cref{sec:combined}. 

Computationally, although the optimization problem in \eqref{def_GM} is convex, the intrinsic non-differentiability of its objective function at certain points renders the exact solution intractable \citep{cardot2013efficient}. Consequently, a $(1+\gamma)$-approximate solution is typically computed, which can be achieved in polynomial time for any arbitrary $\gamma>0$. Our theoretical analysis in the next section assumes that an exact solution to \eqref{def_GM} is obtained. \Cref{sec:approx_kmedian} extends our results to any $(1+\gamma)$-approximate solution.

For the reader’s convenience, our procedure stated in \cref{sec_method_kmeans,sec_method_sign,sec_method_GM} is summarized in \cref{alg:1} and termed Spectral Robust Federated ICA (SRF-ICA). Finally, although motivated by federated ICA, the proposed aggregation procedure and our subsequent analysis in \cref{sec_theory_km,sec_theory_GM} apply more generally to collections of matrix estimators that are identifiable up to signed permutations.

\begin{algorithm}[ht]
\caption{Spectral-Robust-Federated ICA (SRF-ICA)}
\label{alg:1}
\noindent\begin{minipage}{\dimexpr\linewidth-2\algomargin\relax}
\begin{algorithmic}[1]
\Require Local estimators $\wt{\mb A}^{(k)}\in \RR^{r\times r_k}$ for each $k\in[K]$.
\Ensure Robust estimator $\wh{\mb A}\in\RR^{r\times r}$.

\State Let $R=\sum_{k=1}^K r_k$ and 
$\wt{\mb A}_c$ as in \eqref{def_tilde_A_c}. Form 
$
    \mb M= 
    \sqrt{r/R} ~ |\wt{\mb A}_c^\top\wt{\mb A}_c| .
$

\State Compute the top-$r$ eigenvectors $\mb U_r$ of $\mb M$ and set 
$\mb Q=\mb U_r^\top\mb M$.

\State Run the $k$-means algorithm on the columns of $\mb Q$ as in \eqref{kmeans_spec} to obtain clusters
$\wt{\mathcal C}_1,\ldots,\wt{\mathcal C}_r$.

\State For each $a\in[r]$, let $\wt{\mb u}_a$ be the leading left singular vector of $\wt{\mb A}_{\wt\cC_a}$ in \eqref{def:sign_aligned_atom}, and compute $\widecheck{\mb a}^{(k)}_i$ as in \eqref{def:sign_aligned_atom}. 

\State For each $a\in[r]$, compute the geometric median $\wh \ba_a$ given in \eqref{def_GM}. 

\State \Return $\wh{\mb A}=[\wh\ba_1,\ldots,\wh\ba_r]$.
\end{algorithmic}
\end{minipage}
\end{algorithm}

\section{Theoretical results} \label{sec_theory}
In this section, we provide theoretical guarantees of our method in \cref{alg:1}. In \cref{sec_theory_km}, we analyze the solution of the $k$-means problem by deriving  upper bounds of its within-cluster misclustering rates. Such results are then used to provide guarantees for the sign-alignment step. In \cref{sec_theory_GM}, we establish theoretical guarantees for the geometric median in terms of the sample quantiles of the errors between the local estimators and the ground truth. In \cref{sec:combined}, we combine the analyses from \cref{sec_theory_km} and \cref{sec_theory_GM} to provide complete analysis for the proposed \cref{alg:1}. In \cref{sec_theory_ICA}, we specialize to the ICA model and provide explicit error rates. \cref{sec:optimal_aggregate_rate_ica} further discusses the optimality of the proposed estimator in the homogeneous setting.

\subsection{Theoretical guarantees for resolving signed permutation ambiguity}\label{sec_theory_km}

We first analyze the $k$-means clustering in \eqref{kmeans_spec} for resolving the permutation ambiguity. The results presented here assume that an optimal solution to \eqref{kmeans_spec} can be obtained; we extend the analysis to any $(1+\eps)$-approximate solution in \cref{sec:appro_kmeans}.

Recall that $\wt \sigma$ is the optimal permutation map obtained from \eqref{kmeans_spec}. Its overall misclustering error is
\begin{align}\label{def:spec_esti}
    \min_{\pi: [r]\to [r]}~ \sum_{k\in [K]}\sum_{i\in [r_k]} \mb 1\bigl\{\sigma^\star(k, i)\neq\pi(\wt{\sigma}(k, i))\bigr\}
\end{align}
with $\sigma^\star$ given in \eqref{def_sigma_star}.
Let $\pi^\star$ denote any permutation that minimizes  \eqref{def:spec_esti}. For each cluster $\cC_a^\star$ with $a \in [r]$, its {\em within-cluster} misclustering rate is
\begin{align}\label{def:s_max}
    s_a \left(\wt{\sigma}\right)~ :=~  \frac{1}{n_a} \sum_{k\in [K]}\sum_{i\in [r_k]}
    \mb 1\bigl\{\sigma^\star(k, i) = a,\ \pi^\star(\wt \sigma(k, i)) \ne a\bigr\}.
\end{align}
Here recall that $n_a = |\cC_a^\star|$. 
The analysis of $s_a(\wt\sigma)$ depends on how close the Gram-type matrix $\mb M^{(r)}$ in \eqref{def:M_r} is to its population counterpart. The latter,  in view of \eqref{def_M_mat}, is defined as
\begin{equation}\label{def_M_star}
\mb M^\star:= {1\over \sqrt{\bar n}}\,\mb Z^\top \mb Z \in \RR^{R\times R},
\end{equation}
with $\bZ$ given in \eqref{def_sigma_star}.
Denote the within-cluster and overall estimation errors of $\bM^{(r)}$ by, respectively,
\begin{equation}\label{def_eta_a}
    \eta_a:=\frac{1}{n_a}\sum_{(k,i)\in \cC_a^\star}\bigl\|\mb m^{(r)}_{(k,i)}-\mb m_{(k,i)}^\star\bigr\|^2,
    \qquad
    \eta:={1\over \bar n}\sum_{a=1}^r n_a \eta_a=\frac{1}{\bar n}\bigl\|\mb M^{(r)}-\mb M^\star\bigr\|_F^2.
\end{equation} 
The following lemma  bounds from above the within-cluster misclustering error $s_a(\wt\sigma)$ by the within-cluster estimation error $\eta_a$ for each $a\in [r]$. Its proof is stated in \cref{app_sec_proof_lem:kmeans}.

\begin{lemma}\label{lem:k_means}
     Under conditions \eqref{ass_orthogonal} and \eqref{ass_cover}, if $
     \eta \le n_{\min}^2 / (128 \bar n^2)
    $
    holds, then one has
    \begin{align}\label{up_bd_sm}
        s_a\left(\wt{\sigma}\right) ~ \le ~  8\eta_a  ~ (\bar n / n_{\min}),\qquad \forall ~ a\in [r].
    \end{align}
\end{lemma}

The result in \cref{lem:k_means} depends on $\eta_a$ and $\eta$, both of which are further related to the estimation errors of the local estimators. Specifically, for each client $k\in[K]$, recall the permutation map $\sigma^\star(k,\cdot)$ in \eqref{def_sigma_star}. For the $k$-th client, the $\ell_2$-norm estimation error of each estimated column $i\in [r_k]$ is defined as 
\begin{align}\label{eqn:def_data_A}
    \epsilon^{(k)}_{i}
    :=
    \min_{\xi\in\{\pm1\}}
    \bigl\|
        \wt{\mb a}^{(k)}_i
        -
        \xi\,\mb a^\star_{\sigma^\star(k,i)}
    \bigr\| =  
    \bigl\|
        \wt{\mb a}^{(k)}_i
        -
        \xi^\star_{(k,i)}\,\mb a^\star_{\sigma^\star(k,i)}
    \bigr\|.
\end{align}
For each $a\in [r]$, the {\em column-wise} error for estimating $\ba^\star_a$ is then defined as 
\begin{equation}\label{def_epsilon_ell}
    \epsilon_a := \frac{1}{n_a}\sum_{(k,i) \in \cC^\star_a }(\epsilon^{(k)}_{i})^2
\end{equation}
while the {\em overall} estimation error for $\bA^\star$ is defined as
\begin{align}\label{def_epsilon_all}
    \epsilon:=  {1\over \bar n}\sum_{a=1}^r n_a \epsilon_a  = \frac{1}{\bar n}\sum_{k\in [K]}\sum_{i\in [r_k]}(\epsilon^{(k)}_{i})^2.
\end{align}
The following lemma links $\eta$ to $\epsilon$, and is proved in \cref{app_sec_proof_lem:spec}.%

\begin{lemma}\label{lem:spectral_kmeans}
Under conditions \eqref{ass_orthogonal} and \eqref{ass_cover}, one has
\begin{align}\label{ieq:eta_upperbound}
    \eta \le 4\left(2 \sqrt{\epsilon ~ (n_{\max}/\bar n)}+\epsilon\right)^2.
\end{align}
Consequently, if there exists some sufficiently small constant $c>0$ such that
\begin{equation}\label{ieq:cond_epsilon}
\epsilon \le c ~ n_{\min}^2 /(\bar n n_{\max})
\end{equation}
holds, then the conclusion in \eqref{up_bd_sm} holds. 
\end{lemma}

The overall estimation error of $\bM^{(r)}$ is related to that of all local estimators as in \eqref{ieq:eta_upperbound}. This bound, together with condition \eqref{ieq:cond_epsilon}, implies $\eta \le n_{\min}^2 / (128 \bar n^2)$, so that one can invoke \cref{lem:k_means}. %
Condition \eqref{ieq:cond_epsilon} is the main requirement in our analysis, which imposes an upper bound on the overall estimation errors of all client estimators. The explicit expression of $c$ can be found in \eqref{def_cond_eps} of the proof. We emphasize that it does {\em not} imply all local estimators to be consistent when $n_{\min} \asymp n_{\max}$. We defer a detailed discussion to \cref{example}.
\\

The result in \cref{lem:spectral_kmeans} for resolving the permutation ambiguity is key to analyzing the sign adjustment procedure in \cref{sec_method_sign}. The following lemma provides the theoretical justification for Step~4 of \cref{alg:1}: the leading singular vector $\wt{\mb u}_a$ of each estimated cluster induces the correct sign adjustment for all sufficiently accurate atoms. Its proof is given in \cref{app_sec_proof_lem_sign}.

\begin{proposition}[Signed permutation alignment]
\label{prop:sign_alignment}
Grant conditions   \eqref{ass_orthogonal}, \eqref{ass_cover} and \eqref{ieq:cond_epsilon}. For any $a\in[r]$ and  any local estimator $\wt\ba_i^{(k)}$ with $(k,i) \in \cC^\star_{a}$, if 
\begin{equation}\label{cond_eps_ik}
    \epsilon_i^{(k)}
    <~  c_0 \sqrt{n_{\min}  / n_{\max}}
\end{equation}
holds for some universal constant $c_0>0$, then
\begin{enumerate}[noitemsep]
    \item[(i)] this local estimator is correctly clustered, that is, $(k,i) \in \wt \cC_{\pi^\star(a)}$ with $\pi^\star$ given in \eqref{def:spec_esti}, and
    \item[(ii)] its sign-aligned version $\widecheck{\mb a}^{(k)}_i$ given in \eqref{def:sign_aligned_atom} satisfies
\begin{align}\label{def:check_epsilon_sign}
    \bigl\|
        \widecheck{\mb a}^{(k)}_i
        -
        \varsigma_a\,\ba^\star_a
    \bigr\|=    
    \bigl\|
        \wt{\mb a}^{(k)}_i
        -
        \xi^\star_{(k,i)}\,\ba^\star_a
    \bigr\|= 
    \epsilon_i^{(k)}.
\end{align} 
where  $\varsigma_a
    :=
    \operatorname{sgn}\left( \left\langle
        \wt{\mb u}_a,\ba^\star_a \right\rangle\right)$
    and $\xi^\star_{(k,i)}$ is the oracle sign alignment  given in \eqref{def_xi_star}.
\end{enumerate} 
\end{proposition} 
Part~$(i)$ of \cref{prop:sign_alignment} states that any sufficiently accurate local estimator, in the sense of \eqref{cond_eps_ik}, will be correctly clustered in the first step, thereby resolving its permutation ambiguity. Similar to \eqref{ieq:cond_epsilon}, condition \eqref{cond_eps_ik} does not imply consistency of the individual estimator when $n_{\min} \asymp n_{\max}$. Part~$(ii)$ further strengthens this result by showing that the sign ambiguity of such an estimator can also be resolved in the second step. Moreover, as shown in \eqref{def:check_epsilon_sign}, the sign-alignment step in \eqref{def:sign_aligned_atom} only resolves the sign ambiguity without inflating the scale of the original estimation error. For future reference, we define the individual estimation error after permutation and sign alignment as
\[
    \widecheck{\epsilon}_i^{(k)}:= \bigl\|
        \widecheck{\mb a}^{(k)}_i
        -
        \varsigma_a\,\ba^\star_a
    \bigr\|,\qquad \text{for }(k,i) \in \cC_a^\star \text{ and }a\in [r].
\]

\subsection{Theoretical analysis of the robust aggregation via geometric median}\label{sec_theory_GM}

In this section, we first state a general result of the geometric median (GM)   and then apply it to our GM estimator in Step 5 of \cref{alg:1} after resolving both sign and permutation ambiguities. 

Let $ \bx_1, \ldots, \bx_N$ represent a collection of estimates of some target vector $\mb x^\star \in \RR^r$. Their GM is defined as
\begin{align}\label{eqn:def_gm}
     {\rm GM}(\bx_1,\ldots,\bx_N):=\argmin_{\mb y\in \mathbb{R}^r}\sum_{i=1}^N\|\mb y-\mb x_i\|.
\end{align}
Its estimation error is well understood in the literature (see, for instance, \citep{chen2017distributed}) and is characterized in terms of the quantiles of the estimation errors of the individual $\mb x_i$. 
To state the result, we first recall the following definition of sample quantiles.
\begin{definition}\label{def:p_quantile}
    Let $z_1,\ldots,z_N\in\RR$, and let $z_{(1)}\le \cdots \le z_{(N)}$ denote their order statistics. For any $p\in(0,1]$, define the $p$-th sample quantile as
    \[
    Q(p;\{z_1,\ldots,z_N\}) = z_{(\lceil pN\rceil)}.
    \]
\end{definition}

By \cref{def:p_quantile}, at least $pN$ of $z_1,\ldots,z_N$ are no greater than $Q(p;{z_1,\ldots,z_N})$. Moreover, it is easy to see that at least $(1-p)N$ are no less than it. 
Applying \citet[Lemma 1]{chen2017distributed} with $r = Q(p; {\|\bx_i-\bx^\star\|}_{i\in[N]})$ and $\alpha = 1-p$, and then taking the infimum over $p\in(1/2,1]$, gives the following result.
\begin{lemma}\label{lem:gm_robust}
The GM given by \eqref{eqn:def_gm} satisfies 
\begin{align}
     \bigl\|{\rm GM}(\bx_1, \ldots, \bx_N)-\mb x^\star\bigr\|  ~ \leq ~ \inf_{p\in(1/2, 1]}\, \frac{2p}{2p-1}Q\left(p; \{\|\bx_i-\bx^\star\|\}_{i\in [N]}\right).\nonumber
\end{align}
\end{lemma}

As shown in \cref{lem:gm_robust},  the error of the GM estimator can be upper bounded by the infimum of the product between the $p$-quantile of the individual estimation errors and the factor $(2p)/(2p - 1)$, over $p \in (1/2, 1]$. It in particular implies that the GM estimator remains consistent as long as more than half of the original estimators are consistent. We also remark that the infimum  in \cref{lem:gm_robust} is attained.

By this robustness property, our estimator $\wh\bA=[\wh\ba_1,\ldots,\wh\ba_r]$ of $\bA^\star$ in \eqref{def_GM}, obtained via geometric median aggregation, would be consistent if more than half of the estimators in each of the estimated clusters $\wt\cC_1,\ldots,\wt\cC_r$ are consistent. Recall the individual estimation errors $\widecheck{\epsilon}_i^{(k)}$  in \eqref{def:check_epsilon_sign} after sign and permutation alignment. For each $a\in[r]$, define the $p$-quantile of the estimation errors of all local sign-aligned estimators in $\wt\cC_a$ by
\begin{align}\label{def:check_bar_quantile}
        \widecheck{Q}_a(p) := Q_a\Bigl(p; \bigl\{\widecheck{\epsilon}_i^{(k)}: (k,i)\in \wt\cC_a\bigr\}\Bigr).
\end{align}
Its counterpart under oracle sign adjustment, with $\epsilon_i^{(k)}$ given in \eqref{eqn:def_data_A},  is
\begin{align}\label{def:bar_quantile}
        \wt Q_a(p) := Q_a\Bigl(p; \bigl\{\epsilon_i^{(k)}: (k,i)\in \wt \cC_a\bigr\}\Bigr).
\end{align}
The following lemma relates  $\wt Q_a(p)$ and $\widecheck{Q}_a(p)$.%
\begin{lemma}[Quantile preservation after sign alignment]
\label{lem:quantile_preservation_sign}
Under conditions   \eqref{ass_orthogonal}, \eqref{ass_cover} and \eqref{ieq:cond_epsilon}, for each $a\in[r]$ and for any $p\in(0,1)$ satisfying
$
    \wt Q_a(p)<c_0\sqrt{n_{\min} /n_{\max}},
$ we have
$
    \widecheck{Q}_a(p) \le \wt Q_a(p).
$ 
\end{lemma}

The proof of \cref{lem:quantile_preservation_sign} is given  in   \cref{sec:proof_quantile_preservation_sign} and is a direct consequence of \cref{prop:sign_alignment}. \cref{lem:quantile_preservation_sign} shows that the sample quantiles of the estimation errors within each estimated cluster do not increase after the sign-alignment step in \cref{sec_method_sign}. 

On the other hand, since both $\wt Q_a(p)$ and $\widecheck{Q}_a(p)$ still depend on the $k$-means clusters, we need to relate them to the original  quantile estimation error defined using the true clusters $\cC^\star_1,\ldots,\cC^\star_r$. Specifically, denote the $p$-quantile of the estimation errors of all local estimators, after both oracle sign and oracle permutation adjustments, by
\begin{equation}\label{def:p_quantile_i}
    Q_a(p) : = Q_a\Bigl(p; \bigl\{\epsilon_i^{(k)}: (k,i)\in\cC^\star_a\bigr\}\Bigr).
\end{equation} 
 The following lemma relates $\wt Q_a(p)$ with $Q_a(p)$. Its proof can be found in \cref{app_sec_proof_lem:mis_class}. Write for short 
 \begin{align}\label{def_delta_eps}
     \delta_\epsilon  ~ := ~ 32 {\bar n^2 \over n_{\min}^2}  \left(2 \sqrt{\epsilon ~ {n_{\max}\over \bar n}}+\epsilon\right)^2.
 \end{align}

\begin{lemma}[Quantile preservation after both sign and permutation alignments]\label{lem:mis_class} 
Grant conditions   \eqref{ass_orthogonal}, \eqref{ass_cover} and \eqref{ieq:cond_epsilon}. 
For each $a\in[r]$ and for any $p\in(1/2,1]$ such that
$
    Q_a(p)<c_0 \sqrt{n_{\min} / \bar n},
$
one has
\begin{align}\label{eqn:lem:mis_class:2}
        \widecheck Q_a(\bar p)
    ~ \le ~ 
    \wt Q_a(\bar p)
    ~ \le ~ 
    Q_a(p),\qquad \text{for all }~  \bar p \le p/(1+\delta_\epsilon).
\end{align}
\end{lemma}  

The key to proving \cref{lem:mis_class} is to track how the proportion of misclustered points in  $\wt \cC_a$ affects the quantiles of the corresponding estimation errors. \cref{lem:k_means} is used to show that this proportion is bounded by $\delta_\epsilon$, which is then used to establish \cref{lem:mis_class}. It can be seen that the error quantiles of $\{\widecheck{\epsilon}_i^{(k)}:(k,i)\in\wt\cC_a\}$ after permutation and sign alignment remain close to those of the oracle-aligned errors $\{\epsilon_i^{(k)}:(k,i)\in\cC^\star_a\}$, in the precise sense of \eqref{eqn:lem:mis_class:2}. This provides a key ingredient for analyzing our proposed estimator below.

\subsection{Theoretical guarantees for \cref{alg:1}}\label{sec:combined} 
Combining the results from \cref{sec_theory_km,sec_theory_GM}, we provide a complete analysis of \cref{alg:1} in this section. 
Recall the error bound of the geometric median (GM) in \cref{lem:gm_robust} and  $\delta_\epsilon$ from \eqref{def_delta_eps}. For any $a\in [r]$, let 
\begin{align}\label{def:optimal_p_i}
    p_a:=\argmin_{\left(1+\delta_\epsilon\right)/2< p \le 1} ~ \frac{2p}{2p-1-\delta_\epsilon}Q_a(p).%
\end{align}
denote the best quantile over $(\left(1+\delta_\epsilon\right)/2,1]$. %

The following theorem is our main result which bounds the estimation error of $\wh\bA$ from \cref{alg:1} in terms of the best error quantiles $Q_a(p_a)$, $a\in [r]$. Its  proof can be found in \cref{app_sec_proof_thm:recover}.

\begin{theorem}\label{thm:recover}
Grant conditions   \eqref{ass_orthogonal}, \eqref{ass_cover} and \eqref{ieq:cond_epsilon}. 
For any $a\in [r]$, assume 
     $Q_a(p_a) < c_0 \sqrt{n_{\min} / \bar n}$ with $c_0$ given in \eqref{cond_eps_ik}. Then there exists 
    some  permutation $\pi:[r]\to [r]$ such that for any $a\in [r]$,
    \begin{align}\label{rate_propose}
         \min_{\xi\in \{\pm1\}}\bigl\|\wh\ba_a - \xi\mb a^\star_{\pi(a)}\bigr\|  ~ \leq ~   \frac{2 p_a}{2 p_a - 1 - \delta_\epsilon} Q_a(p_a). 
    \end{align} 
\end{theorem} 
Theorem~\ref{thm:recover} highlights the main advantage of our aggregation procedure based on the geometric median: the final error bound for estimating each $\ba_a^\star$ depends on the quantiles of the estimation errors of its oracle-aligned local estimators. As long as more than half of these local estimators are consistent, the resulting estimator is also consistent.

To better appreciate this gain, it is useful to contrast \eqref{rate_propose} with the performance of the estimator $\bar\bA$ in \eqref{def_centers}, which simply averages the local estimators after resolving their signed permutation ambiguity. The following theorem establishes the estimation error of this averaging procedure. Its proof is given in \cref{sec:proof_kmeans_center}. 

\begin{theorem}\label{thm:kmeans_center} 
 Under conditions   \eqref{ass_orthogonal}, \eqref{ass_cover} and \eqref{ieq:cond_epsilon}, 
there exists some permutation $\pi:[r]\to[r]$  such that 
\begin{align}\label{rate_center}
        \min_{\xi\in \{\pm 1\}}\bigl\|
        \bar\ba_a-\xi\ba_{\pi(a)}^\star
    \bigr\|
    ~ \le ~ 
    C\left(\sqrt{\epsilon_a} +\frac{\bar{n}n_{\max}}{n_{\min}^2}\epsilon\right),\qquad \forall ~ a\in [r].
\end{align}
Here $C>0$ is some absolute constant depending only on $c$ in \eqref{ieq:cond_epsilon}. 
\end{theorem} 

Although we advocate aggregation using the geometric median, \cref{thm:kmeans_center}, which provides the estimation error for the simple average of the aligned local estimators, is also of independent interest. 

As shown in \eqref{rate_center}, since each  $\bar \ba_a$ is the average of all local estimators within the estimated cluster, the rate in \eqref{rate_center} is largely determined by the worst local estimator among all clients. In the presence of severe heterogeneity, where multiple clients, or even a single client, provide poor or inconsistent estimators, the averaged estimators can converge slowly or even become inconsistent. 

Both our methods and theory up to this point apply more generally to settings where each client estimates a target set of vectors up to a signed permutation. In the next section, we specialize to particular ICA models and derive more explicit rates for the error bounds in \eqref{rate_propose} and \eqref{rate_center}.

\subsection{Explicit error rates under the ICA model}\label{sec_theory_ICA}

As revealed by the previous section, the error of the aggregated estimator, whether based on the geometric median or the simple average, depends on the quantiles of the recovery errors $\epsilon^{(k)}_i$ of the local estimators across clients. In this section, we provide more concrete results by focusing on a well-studied class of ICA models, as mentioned in \cref{sec_method_ICA}. 

For ease of presentation, we assume that each client $k$ observes data $\bY^{(k)}$ generated from all $r$ signals mixed through the columns of $\bA^\star$, although the extension to heterogeneous $r_k$ is straightforward.
\begin{assumption}\label{assu:complete_clients}
For every client $k\in[K]$, we have $r_k=r$, so that $\bA^{(k)} = \bA^\star \in \bbO_{r,r}$. 
\end{assumption}
Under \cref{assu:complete_clients}, one has 
$
    n_1 = \cdots = n_r 
    = K. 
$ 
The main working condition on latent signal $\bX^{(k)}$ is given below. We refer to the discussion following the Assumption 1 of \cite{bing2025optimal} and to \cite{rohe2020vintage} for further details. The sub-Gaussian assumption can be relaxed to suitable finite-moment conditions; see, for instance, \citep{auddy2023Large}. 
\begin{assumption}\label{ass_X}
    For each client $k\in[K]$, the entries of $\bX^{(k)}\in\RR^{r\times N_k}$ in model \eqref{def:model} are mutually independent, and each entry has zero mean, variance $\sigma^2>0$, and excess kurtosis $\kappa>0$. Each entry is further assumed to be sub-Gaussian\footnote{A random variable $W$ is said to be sub-Gaussian with sub-Gaussian constant $\gamma$ if, for any $t\in \mathbb R$, $\EE[\exp(tW)]\le \exp(t^2\gamma^2/2)$.} with some sub-Gaussian constant $\gamma<\infty$.
\end{assumption}

Note that \cref{ass_X} does not imply independence of the data across different clients. 
Under model \eqref{def:model} and \cref{assu:complete_clients,ass_X}, \cite{auddy2023Large} establishes the following minimax lower bound for estimating the columns of $\bA^\star$ from the local data $\bY^{(k)}$:
\begin{align}\label{optimal_rate}
    \inf_{\wt{\bA}}
    \sup_{\bA^\star\in \bbO_{r,r}}  
    \E^{(k)} \min_{\bP\in \cP_r}  \bigl\|
        \wt{\bA}- \bA^\star \bP
        \bigr\|_{2,\infty}^2
     \gtrsim 
      {\frac{r}{N_k}},\qquad 
    \inf_{\wt{\bA}}
    \sup_{\bA^\star\in \bbO_{r,r}} 
    \E^{(k)} \min_{\bP\in \cP_r} \bigl\|
        \wt{\bA}- \bA^\star \bP
        \bigr\|_F^2 
      \gtrsim 
      {\frac{r^2}{N_k}}.
\end{align}
Here $\cP_r$ denotes the set of all $r\times r$ signed permutation matrices, the expectation $\E^{(k)}$ is taken with respect to the law of $\bY^{(k)}$, and the infimum is taken over all estimators $\wt{\bA}$ based on $\bY^{(k)}$. %
Furthermore, both rates in \eqref{optimal_rate} can be achieved, up to logarithmic factors, by computationally efficient algorithms, including the deflationary procedures studied in \cite{auddy2023Large} and \cite{bing2025optimal}. These methods recover the columns sequentially and employ practical initialization strategies despite the inherent non-convexity of the ICA formulation \eqref{eq:ICA_form}. Motivated by these results, we impose the following condition on the errors $\epsilon_i^{(k)} $ of local estimators, given in \eqref{eqn:def_data_A}. Let $T := \max_{k\in[K]} N_k \vee K$.
\begin{assumption}\label{ass_epsilon}
    With probability  $1-1/T$, one has  
     $
        \epsilon_i^{(k)}  \lesssim \sqrt{r\log(T)/N_k}
     $ 
      for all $(k,i)\in \cI$.
\end{assumption}
\cref{ass_epsilon} is satisfied by the deflationary FastICA estimator of \cite{auddy2023Large} under 
\begin{equation}\label{cond_Nk}
    N_k \ge Cr^2\log^2 T
\end{equation}
for some sufficiently large constant $C>0$. Up to logarithmic factors, this sample-size requirement is further shown therein to be necessary for all low-degree polynomial algorithms. \cref{ass_epsilon} is also satisfied by the deflationary estimator of \cite{bing2025optimal} under $N_k \ge Cr^3\log T$; with some additional data splitting, this condition can be improved to $N_k \ge C r^2\log T$.

Recall from \eqref{def_epsilon_ell} and \eqref{def_epsilon_all}. \cref{assu:complete_clients,ass_epsilon} readily give
\begin{equation}\label{explicit_order_snr}
    \epsilon_a \asymp {1\over K}\sum_{k=1}^K {r\log T\over N_k},\qquad \epsilon \asymp {1\over K}\sum_{k=1}^K {r^2\log T\over N_k}.
\end{equation}
Moreover, $\delta_\epsilon$ given in \eqref{def_delta_eps} is of order  $\epsilon$ and the $p$-quantile of estimation errors in
\eqref{def:p_quantile_i} becomes that of  $\sqrt{r\log(T)/N_1},\ldots,\sqrt{r\log(T)/N_K}$.
Similar to \eqref{def:optimal_p_i} we define 
\begin{align*}
    p^\star:=\argmin_{ (1+ \epsilon)/2 < p \le 1} \frac{2p}{2p-1- \epsilon}Q\left(
                p; \left\{ \sqrt{r\log T\over N_1},\ldots,\sqrt{r\log T\over N_K} \right\} \right).
\end{align*}
The following corollary follows immediately from \cref{thm:recover,thm:kmeans_center}.  
\begin{corollary}[Explicit error rates under the ICA model]\label{cor:dictionary}
Grant \cref{assu:complete_clients,ass_X,ass_epsilon} as well as condition \eqref{cond_Nk}. 
    Then with probability at least $1-1/T$,  the GM estimator  $\wh{\mb A}$  of \cref{alg:1} satisfies: 
    \begin{align}\label{rate_final}
        \min_{\bP \in \cP_r}     \bigl\|\wh \bA  - \bA^\star \bP\bigr\|_F^2  ~ \lesssim ~      {Q\left(p^\star; \left\{{\frac{r^2\log T}{N_1}},\ldots,{\frac{r^2\log T}{N_K}}\right\}\right)},  
    \end{align}
while the averaged estimator  $\bar{\mb A}$ in \eqref{def_centers} satisfies 
\[
      \min_{\bP \in \cP_r}  \bigl\|
        \bar\bA - \bA^\star \bP
    \bigr\|_F^2
    ~ \lesssim ~ 
   {1\over K}\sum_{k=1}^K\left( {r^2 \log T\over N_k}  +  {r^5\log^2 T\over N_k^2}\right).
\]
\end{corollary}
Condition \eqref{cond_Nk} ensures \eqref{ieq:cond_epsilon} and $Q_a(p) \le c_0 \sqrt{n_{\min} / \bar n}$ in \cref{thm:recover}. We remark that it does not imply all local estimators have vanishing estimation errors, ignoring the logarithmic factors. Indeed, when there exist some  $k$ such that $C_1 r^2 \le N_k   \le C_2 r^2$ for sufficiently large $C_1\le C_2$, the local estimator $\wt{\bA}^{(k)}$ of that client cannot be consistent under the Frobenius norm in view of \eqref{optimal_rate}. Consequently, the error bounds of the averaged estimator in \eqref{rate_center} are not vanishing. By contrast, as revealed in \eqref{rate_final}, the GM estimator is consistent as long as more than $1/2 + o(1)$ of the clients have sample sizes satisfying $N_k / r^2 \to \infty$.

\subsection{Near-optimality  under the homogeneous ICA setting}
\label{sec:optimal_aggregate_rate_ica}

The preceding results demonstrate the robustness of the GM estimator under heterogeneous federated ICA. In this section, we further benchmark its optimality when the clients have homogeneous local data, thereby showing that this robustness comes without a loss of statistical efficiency. To this end, we assume that all clients have sample sizes of the same order. 
\begin{assumption}\label{assu:sample_order_N}
   There exists some  constant $c>0$ such that $N := \min_{k\in [K]} N_k \ge c \max_{k\in [K]} N_k$.
\end{assumption}

Under \cref{assu:sample_order_N}, \cref{cor:dictionary} yields that 
\[
    \EE \min_{\bP \in \cP_r} \bigl\|\wh \bA-\bA^\star \bP\bigr\|_F^2
    ~\lesssim~
    \frac{r^2\log T}{N}.
\]
In view of \eqref{optimal_rate}, when the number of clients $K$ remains fixed, this rate cannot be improved even by a central server with access to all local data. Thus, for fixed $K$, the proposed GM estimator is minimax optimal, up to a logarithmic factor, in the homogeneous ICA setting. Moreover, without additional assumptions, allowing $K\to\infty$ does not in general lead to a faster rate.

Under the additional assumption that the local data are independent across clients, however, the above error bound for the GM estimator does not exploit this independence and therefore does not capture the statistical gain as $K\to\infty$. To provide intuition, in the absence of signed-permutation ambiguity, averaging $K$ independent local estimators would reduce the stochastic error by a factor of $K^{-1/2}$, a gain not captured by the preceding error bound. In the remainder of this section, we establish a stronger result showing that, in addition to its robustness under heterogeneity, the GM estimator achieves the rate $r^2\log N/(NK)$ as $K\to\infty$.

To establish this stronger result, we further require each local estimator to admit a first-order stochastic representation, as specified in the following assumption. For each $(k,i)\in\cI$, recall $\sigma^\star(k,i)$ and $\xi^\star_{(k,i)}$ from \eqref{def_sigma_star} and \eqref{def_xi_star}, respectively.

\begin{assumption}\label{assu:tilde_A_optimal}
    Assume that the local data $\bY^{(k)}$ are independent across $k\in [K]$. 
     For each $k\in[K]$ and $i\in [r]$, under the notation in \cref{ass_X}, its $i$-th local estimator satisfies the decomposition 
    $$
        \wt\ba_i^{(k)} - \xi^\star_{(k,i)}~  \ba_{\sigma^\star(k,i)}^\star  =  \frac{1}{3\kappa N_k} 
    \sum_{t=1}^{N_k}
    x_{it}^{(k)3}\!\!\! \sum_{j\in [r]\setminus \{i\}} \!\! x_{jt}^{(k)} \ba_j^\star   + \bv_i^{(k)}, 
    $$ 
    where  the higher-order term $\bv_i^{(k)}$ satisfies $\P\{\|\bv_i^{(k)}\| \lesssim r^{3/2}\log(T)/ N_k,~ \forall  (k,i)\in \cI\} \ge 1-T^{-1}$. 
\end{assumption}
The first term on the right-hand side is a zero-mean stochastic term that is asymptotically normal at the root-$N_k$ rate. It is easy to see that the higher-order term $\bv_i^{(k)}$   is negligible under condition \eqref{cond_Nk}.  
\Cref{assu:tilde_A_optimal} can be verified, for instance, when $\wt\ba_1^{(k)},\ldots,\wt\ba_r^{(k)}$ are obtained by solving a variant of the one-column recovery problem \eqref{eq:ICA_form} in \cite{bing2025optimal} $r$ times and taking a stationary point from each run, with initializations as specified in their Section 5.2. Details are provided in \cref{lem:client_expansion} of \cref{sec:aggre_proof_local_exp}.

The following theorem gives the error rates of the GM estimator under the aforementioned setting. Its proof can be found  in \cref{sec:proof_thm:optimal_aggre_rate}. 
\begin{theorem}\label{thm:optimal_aggre_rate}
    Under \cref{assu:sample_order_N,assu:complete_clients,assu:tilde_A_optimal,ass_X},  assume $r\ge 3$ and condition \eqref{cond_Nk}. Then with probability at least $1-1/T$, %
    the  GM estimator $\wh{\mb A}$ satisfies:
    \begin{align}\label{rate_GM_homo}
            \min_{\bP \in \cP_r}  \bigl\|\wh \bA - \bA^\star \bP\bigr\|_F^2 ~ \lesssim ~  \frac{r^2\log(T)}{NK}+\frac{r^3\log(T)}{N^{3/2}}.
        \end{align}
        Furthermore,  if $N\ge K^2r^2$ holds additionally, then the above bound simplifies to  
        \begin{align}\label{rate_GM_homo_simp}
           \min_{\bP \in \cP_r}   \bigl\|\wh \bA - \bA^\star \bP\bigr\|_F^2  ~ \lesssim ~  \frac{r^2\log(N)}{NK}.
        \end{align} 
\end{theorem}

The two terms in the bound of \eqref{rate_GM_homo} arise from different sources. The first term, $r^2\log(T)/(NK)$, is the leading stochastic error and exhibits the $K^{-1}$ improvement from aggregating independent local estimators. The second term, $r^3\log(T)/N^{3/2}$, arises from the higher-order term in \cref{assu:tilde_A_optimal} and does not benefit from cross-client aggregation.  The latter is dominated by the former under the additional condition $K\le \sqrt{N}/r$. Similar restrictions on the number of clients arise in distributed estimation, where averaging reduces the first-order stochastic error while higher-order local errors may not enjoy the same improvement; see, e.g., \citet{zhang2013communication,rosenblatt2016optimality}.

In view of \eqref{optimal_rate}, the rate in \eqref{rate_GM_homo_simp} is near-optimal in the minimax sense, as it matches, up to logarithmic factors, the optimal rate of a central server with access to all client data of size $NK$. This result also complements \cref{thm:recover}: while \cref{thm:recover} establishes the robustness of the GM estimator to client heterogeneity, \cref{thm:optimal_aggre_rate} here shows that such robustness does not come at the cost of statistical efficiency in the homogeneous regime with independent client data. 

A key technical ingredient of proving \cref{thm:optimal_aggre_rate} is a new
concentration inequality of the geometric median of independent random vectors, stated and proved in \cref{sec:aggre_proof_aggregation}. Unlike existing concentration results for the geometric median
\citep{minsker2015geometric,minsker2024geometric}, our bound allows non-identically distributed random vectors which are approximately spatially centered at the origin, and subject to deterministic perturbations. This result is what
allows the robust geometric-median estimator to retain the usual
$K^{-1}$ aggregation gain, and is of independent interest in its own right.

\section{Experimental results}\label{sec:exp}
\subsection{Simulation studies}\label{sec:sim}

We generate the simulated data as follows. The ground-truth mixing matrix $\mb A^\star\in\mathbb R^{r\times r}$ is constructed by first generating a matrix with i.i.d. standard Gaussian entries and then projecting it onto the set of orthogonal matrices under the Frobenius norm. For each client $k\in[K]$, the entries of the ground-truth source matrix $\mb X^{(k)\star}\in\mathbb R^{r\times N_k}$ are generated as i.i.d. copies of  $X=B W$, where $B\sim\operatorname{Bernoulli}(0.1)$ and $W\sim\mathcal N(0,1)$. Here we choose $r=10$ during the simulation.
To introduce heterogeneity across clients, we partition the $K$ clients into normal and corrupted clients according to their local sample sizes. Each normal client has $N_k=5000$ samples, whereas each corrupted client has a smaller local sample size. In our experiments, we vary both the number of corrupted clients and their local sample size to investigate the effect of client heterogeneity on the aggregation procedures.

For each client $k\in[K]$, the local estimator $\wt{\mb A}^{(k)}$ is computed using the \textit{rotatefactors} function in MATLAB R2026a, which implements an ICA procedure for solving \eqref{eq:ICA_form}. Since $k$-means clustering depends on its random initialization, we run $k$-means 10 times with independent random initializations and retain the solution with the smallest objective value. We measure the estimation error of an estimator $\wh{\mb A}$ by $\min_{\mb P\in \cP_r}\|\wh{\mb A}-\mb A^\star\mb P\|_F$.
 
 We compare the following methods. 
\begin{itemize}[noitemsep]
\item {\bf SRF-ICA} is the proposed method: it first performs spectral $k$-means using the sign-invariant affinity matrix, then aligns signs within each estimated cluster, and finally applies the geometric median. 
\item {\bf SRF-ICA-Noalignment} is an ablation of SRF-ICA in which the within-cluster sign-alignment step is removed; namely, it uses the same spectral $k$-means clustering but directly applies the geometric median to the raw atoms in each cluster. This baseline is included to isolate the effect of Step~3 in \cref{alg:1}. 
\item {\bf SF-ICA} uses the same spectral clustering and sign-alignment steps as SRF-ICA, but replaces the geometric median by the arithmetic mean within each cluster.
\item {\bf Naive Baselines} adopt simple mean and simple median to aggregate local atoms directly according to their column indices without resolving the signed permutation ambiguity.
\end{itemize}

\begin{figure}[ht]
    \centering
    \includegraphics[width=1.0\linewidth,height=.25\textheight]{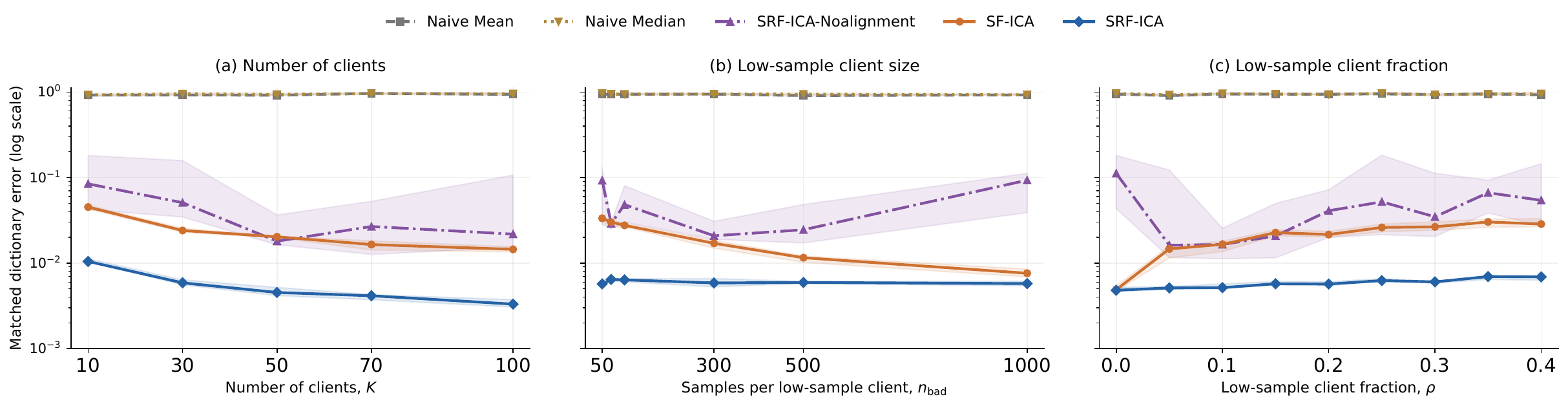}
    \caption{The averaged estimation error of our algorithm (SRF-ICA) with other four competing methods.
}
    \label{fig:error}
\end{figure}

We evaluate the recovery error of the five methods under three heterogeneous settings: (1) varying the total number of clients $K \in \{10,30,50,70,100\}$, (2) varying the sample size of corrupted clients in $\{50,70,100,300,500,1000\}$, and (3) varying the corrupted-client ratio in $\{0,0.05,0.1,0.15,\ldots,0.4\}$. Unless otherwise specified, we set $K=30$, the sample size of each corrupted client to $300$, and the proportion of corrupted clients to $0.1$. Each data point in \cref{fig:error} reports the average estimation error over $20$ independent trials. The $y$-axis is shown on a logarithmic scale.

The results show that SRF-ICA consistently achieves the smallest recovery error across all three panels. Compared to SRF-ICA-Noalignment, the proposed SRF-ICA has a clear advantage, which confirms the importance of the within-cluster sign-alignment step: although the sign-invariant affinity matrix can cluster atoms with opposite signs together, directly applying the geometric median without sign alignment can suffer from the cancellation due to the remaining sign ambiguity in the estimated cluster, as discussed in \cref{sec_method_sign}. Compared with SF-ICA, SRF-ICA benefits from using the geometric median rather than the arithmetic mean after alignment, with the improvement becoming more pronounced as the sample size of corrupted clients decreases or the proportion of corrupted clients increases. This behavior is consistent with the robustness guarantees established in \cref{thm:kmeans_center,cor:dictionary}. In addition, the estimation error of SRF-ICA decreases as the number of clients increases, consistent with the statistical gain from aggregating independent local estimators established in \cref{thm:optimal_aggre_rate}. Finally, the simple mean and simple median baselines yield errors close to $1$ across all settings, indicating that direct index-wise aggregation fails without first resolving the permutation ambiguity. This demonstrates the importance of the clustering step analyzed in \cref{lem:mis_class}.

\subsection{Federated ICA on MNIST}
\label{sec:mnist-real-data}

We next evaluate SRF-ICA on a dataset of $m=70{,}000$ MNIST handwritten-digit images. Each $28\times 28$ grayscale image is vectorized into $\mb z_i\in\mathbb R^{d}$ with $d=784$ and $i\in[m]$, and we seek to recover a dictionary of $r$ latent image patterns, represented by the columns of a loading matrix $\bLambda\in\mathbb R^{d\times r}$. This formulation is motivated by the approximate low-rank factor model
\[
\mb Z=\bLambda\mb X+\mb E,
\]
where $\mb Z\in\RR^{d\times m}$ contains the $m$ vectorized images, $\mb X$ contains the image-specific coefficients, and $\mb E$ represents residual variation. Each column of $\bLambda$ represents a dictionary pattern, while the corresponding row of $\mb X$ determines how strongly that pattern is expressed across images. Unlike in the synthetic experiments, the true $\bLambda$ is unknown for MNIST. We therefore construct a \emph{centralized reference dictionary} using the full dataset and use it as a benchmark for evaluating the dictionaries obtained through federated estimation and aggregation.

Our dictionary estimation procedure uses principal component analysis (PCA) followed by Varimax rotation, motivated by \citet{rohe2020vintage}. PCA first estimates the $r$-dimensional subspace spanned by the columns of $\bLambda$. When the rows of $\bLambda$ have independent and appropriately standardized coordinates, $d^{-1}\bLambda^\T\bLambda$ approaches the identity matrix as $d$ grows, so that the normalized dictionary have  approximately orthogonal columns. Consequently, the leading $r$ left singular vectors $\mb V\in\bbO_{d,r}$ of $\mb Z$ can be viewed, up to estimation error, as an orthogonal rotation of the normalized $\bLambda$, that is, $\mb V^\T\approx d^{-1/2}\mb A\bLambda^\T$ for some $\mb A\in\bbO_{r,r}$. Under the additional condition that the entries of $\bLambda$  are leptokurtic, \citet{rohe2020vintage} show that the rotation $\bA$ can be identified using the Varimax criterion. Such leptokurtosis is closely related to sparse representations of natural images, which in turn give rise to spatially localized dictionary elements \citep{olshausen1996emergence,bell1997independent}. Since the Varimax rotation criterion is equivalent to the FastICA criterion in \eqref{eq:ICA_form} \citep{bing2025optimal}, we solve the latter using the same algorithm as in \cref{sec:sim} to estimate $\bA$ and recover the individual dictionary elements.

Specifically, let $\bZ\in\mathbb R^{d\times m}$ denote the centered data matrix containing all $m=70{,}000$ images. We let $\mb V\in\bbO_{d,r}$ denote its leading $r$ left singular vectors. We then apply symmetric FastICA to the rows of $\mb V$ by solving \eqref{eq:ICA_form} with $\bY^{(k)}$ replaced by $\mb V^\T$. Let $\widehat{\mb A}\in\bbO_{r,r}$ denote the estimated ICA mixing matrix. The centralized reference dictionary is then obtained by rotating the column of $\mb V$  according to 
\[
\mb\Lambda_{\mathrm{ref}} = \mb V\widehat{\mb A} \in \bbO_{d,r}.
\]

\begin{figure}[!ht]
    \centering
    \includegraphics[width=0.6\linewidth]
    {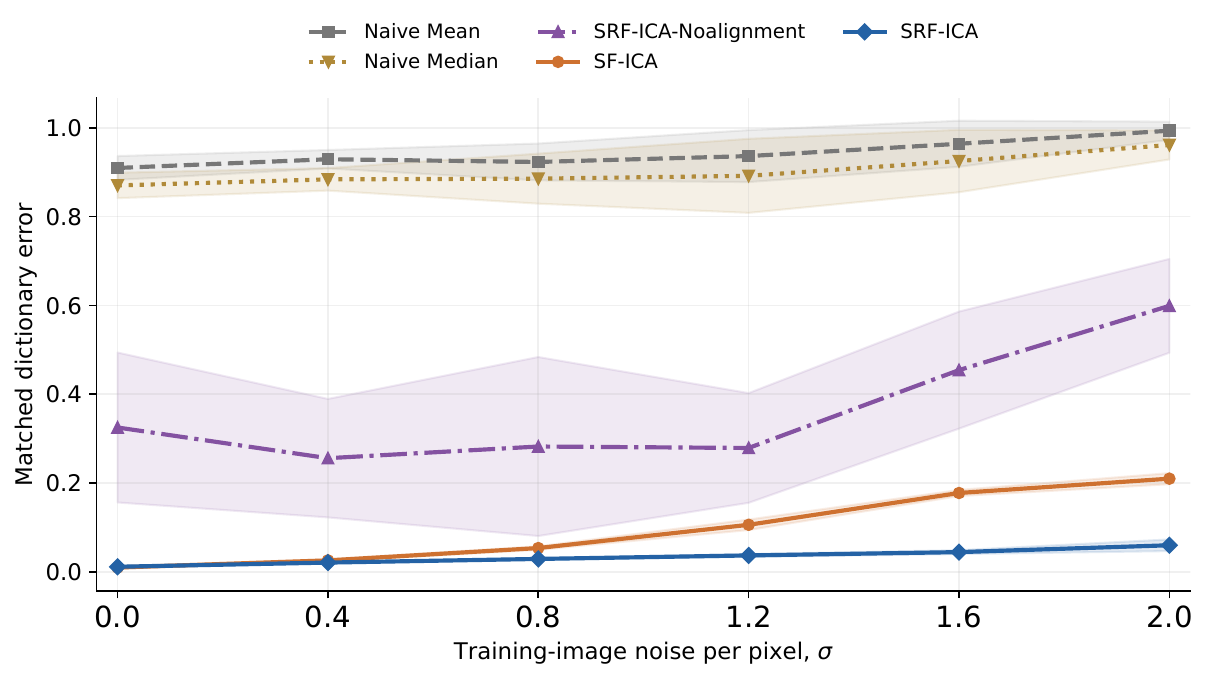}
    \caption{Average estimation error of SRF-ICA and the four competing methods on the MNIST dataset.}
    \label{fig:mnist_noise_error}
\end{figure}

To evaluate federated aggregation under client heterogeneity, we construct $K=10$ clients with balanced label distributions, so that each client contains the same proportions of digits $0,\ldots,9$. Since digit $5$ has only $6{,}313$ images, we let all clients have the same local sample size $N_k=6{,}310$ to maintain these proportions, while four of the ten randomly selected clients are corrupted by additive Gaussian noise. Specifically, for each image $\mb z_i$ assigned to a corrupted client, we replace it by
\[
\widetilde{\mb z}_i=\operatorname{clip}_{[0,1]}\left(\mb z_i+\bs\varepsilon_i\right),\qquad
\bs\varepsilon_i\sim\mathcal N(0,\sigma^2\mb I_d),
\]
where $\operatorname{clip}_{[a,b]}(\bv):=\min\{b,\max\{\bv,a\}\}$ is applied element-wise to any vector $\mb v$. The six normal clients use the original images. All client datasets are disjoint, and we vary the corruption level as $\sigma\in\{0,0.2,0.4,\ldots,2\}$. For a given run, the client partition, the identities of the four corrupted clients, and the algorithmic random seeds are held fixed across all values of $\sigma$, so that changes in performance are  attributed to that of client heterogeneity. Each client independently applies the aforementioned PCA-FastICA procedure to its local data matrix $\bZ^{(k)}$. Let $\mb V^{(k)}$ and $\wh{\mb A}^{(k)}$ denote the resulting left singular vectors and FastICA estimate, respectively. Client $k$ communicates the local dictionary
$\wt{\mb\Lambda}^{(k)}=\mb V^{(k)}\wh{\mb A}^{{(k)}}$.
The central server then aggregates the local dictionaries $\wt{\mb\Lambda}^{(1)},\ldots,\wt{\mb\Lambda}^{(K)}$ using the five methods introduced in \cref{sec:sim}. %
Each aggregated dictionary $\wt{\mb\Lambda}$ is evaluated by its discrepancy from the centralized reference:
$
\min_{\mb P\in\cP_r}
\|\wt{\mb\Lambda}-\mb\Lambda_{\mathrm{ref}}\mb P\|_F/\sqrt{r}.
$
We repeat the entire experiment over five independent runs. For each noise level, we report the mean discrepancy of each method with error bars corresponding to one standard error. \cref{fig:mnist_noise_error} plots the recovery error as a function of $\sigma$ on a logarithmic vertical scale.

\cref{fig:mnist_noise_error} demonstrates the advantage of SRF-ICA
under heterogeneous client data.  As the noise level $\sigma$ increases,
SRF-ICA achieves the smallest recovery error among the five methods, with the
advantage becoming particularly pronounced for $\sigma\ge 0.8$.
The substantial gap over the naive mean and median confirms that resolving the
signed-permutation ambiguity is essential, consistent with the spectral
clustering guarantee in \cref{lem:spectral_kmeans} and the signed-permutation
alignment result in \cref{prop:sign_alignment}.  The improvement over
SRF-ICA-Noalignment further isolates the benefit of the sign-alignment step.
Finally, compared with SF-ICA, which uses the same clustering and alignment
but replaces the geometric median by an arithmetic mean, SRF-ICA degrades much
more slowly as the corrupted clients become noisier.  This behavior agrees with
\cref{thm:recover,thm:kmeans_center}: the error of the geometric-median
estimator is controlled by a majority quantile of the local estimation errors,
whereas arithmetic averaging is more sensitive to low-quality local
estimators. 

\begin{figure}[!ht]
    \centering
    \includegraphics[width=1\linewidth]{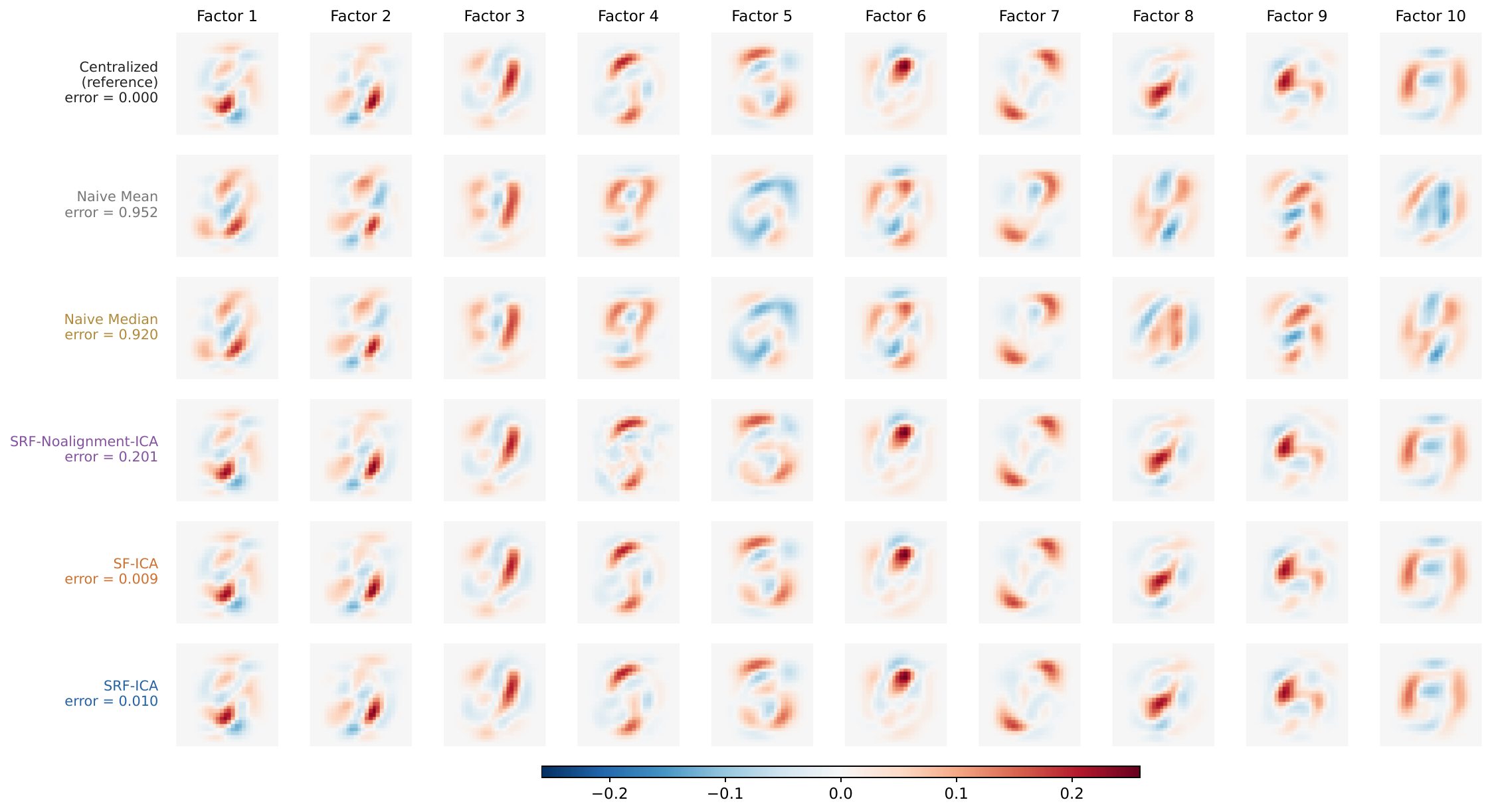}
    \caption{
Visualization of the ten recovered factors when $\sigma=0$.
From top to bottom, the rows correspond to the ground truth, the two naive baselines,
SRF-ICA-Noalignment, SF-ICA, and SRF-ICA respectively.
}
    \label{fig:mnist_noise_0_vis}
\end{figure}

\begin{figure}[!ht]
    \centering
    \includegraphics[width=1\linewidth]{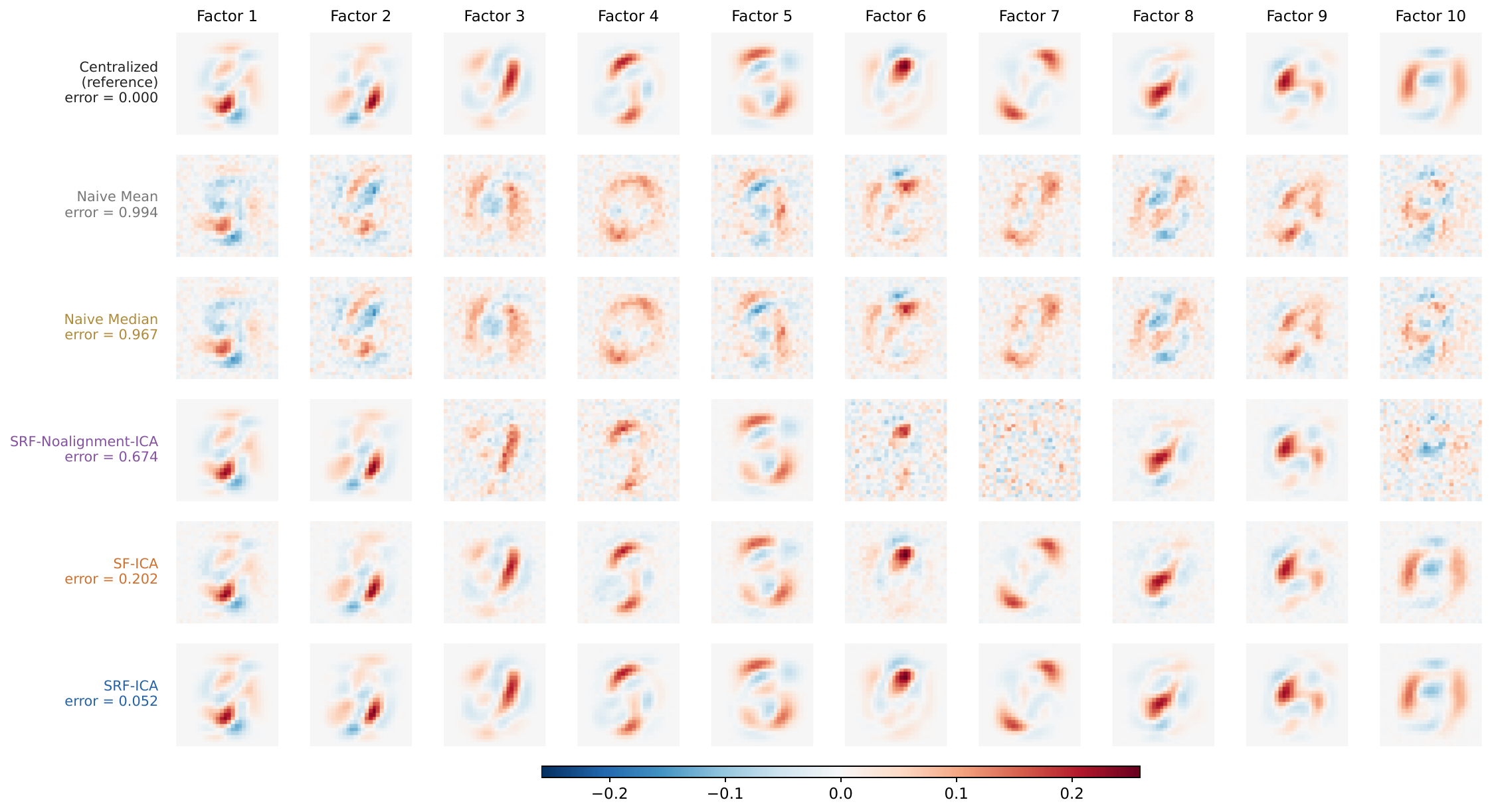}
    \caption{
Visualization of the ten recovered factors when $\sigma=2$.
From top to bottom, the rows correspond to the ground truth, the two naive baselines,
SRF-ICA-Noalignment, SF-ICA, and SRF-ICA respectively.
}
    \label{fig:mnist_noise_2_vis}
\end{figure}

In \cref{fig:mnist_noise_0_vis,fig:mnist_noise_2_vis} we further visualize the ten recovered columns of $\mb\Lambda_{\text{ref}}$ and the estimates of aforementioned methods under $\sigma=0$ and $\sigma=2$, respectively. Even in the noiseless setting of \cref{fig:mnist_noise_0_vis}, the two naive baselines fail to recover several factors, especially factors $5$, $6$, $8$, and $9$, and have large recovery errors. SRF-ICA-Noalignment approximately recovers all ten factors, but some details are missing compared with the reference dictionary, especially for factors $4$ and $5$. In contrast, both SF-ICA and SRF-ICA recover all ten factors clearly with smaller errors.

\begin{figure}[!ht]
    \centering
    \includegraphics[width=.8\linewidth]{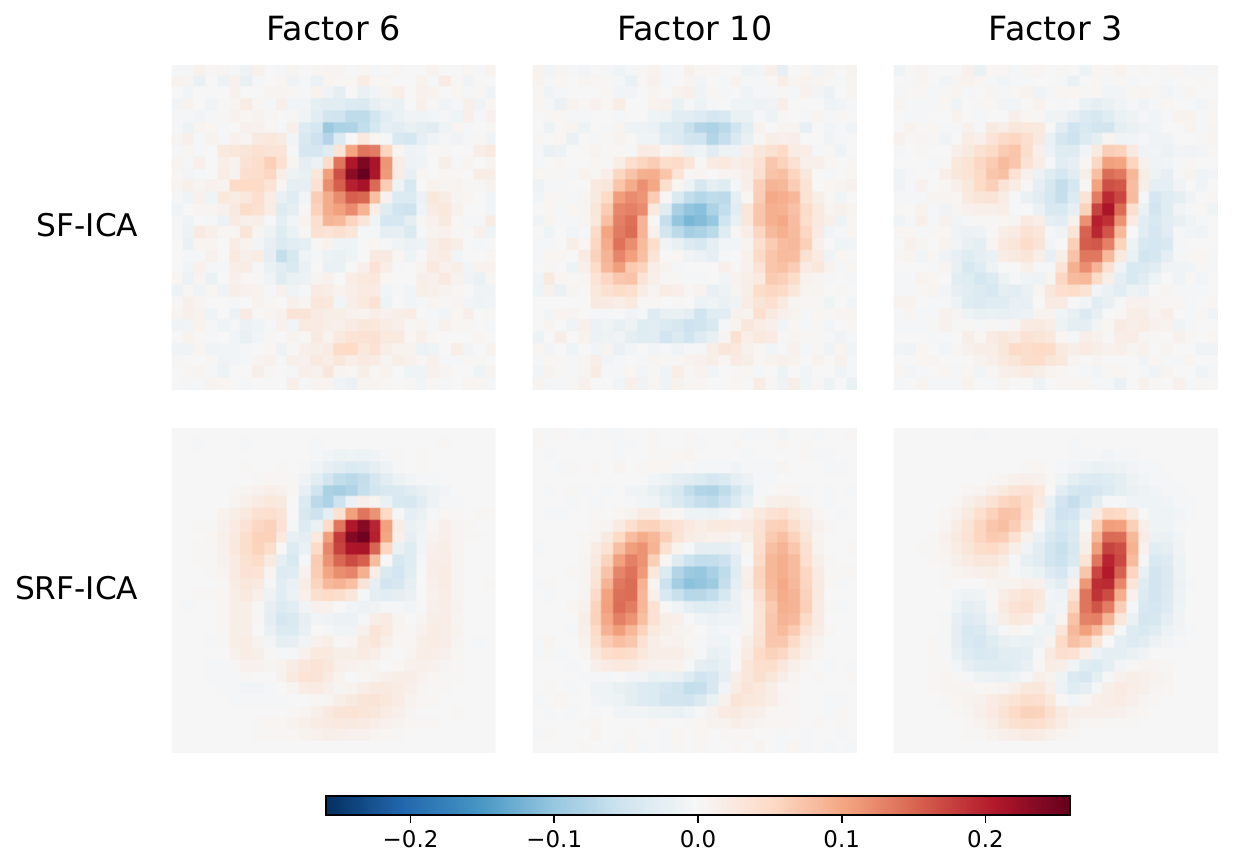}
    \caption{
Enlarged comparison of factors $3$, $6$, and $10$ of SF-ICA and SRF-ICA from
\cref{fig:mnist_noise_2_vis}.
}
    \label{fig:mnist_noise_2_vis_detail}
\end{figure}

When the noise level increases to $\sigma=2$, the results in \cref{fig:mnist_noise_2_vis} further show the advantage of SRF-ICA. The two naive baselines still fail to recover several factors, while SRF-ICA-Noalignment completely fails to recover factors $6$, $7$, and $10$. Both SF-ICA and SRF-ICA successfully recover all ten factors, but SRF-ICA preserves more details and gives visually cleaner results. We also provide enlarged comparisons of SRF-ICA and SF-ICA for factors $3$, $6$, and $10$ in \cref{fig:mnist_noise_2_vis_detail} to show these differences more clearly.

{\small
\bibliographystyle{abbrvnat}
\bibliography{ncvx,fl}
}
\newpage

\appendix

We provide a section-by-section proof in \cref{app_sec_proof_km,app_sec_proof_GM,app_sec_proof_optimality}. \cref{sec:aggre_proof_aggregation} states a new concentration inequality for the geometric median under perturbations. \cref{app_sec_approx} discusses settings in which either the $k$-means problem in \eqref{kmeans_spec} or the geometric median problem in \eqref{def_GM} is solved approximately.

\section{Proofs of \cref{sec_theory_km}}\label{app_sec_proof_km}

\subsection{Proof of the equivalence between two $k$-means formulations}\label{app_sec_proof_kmeans_equiv}

We prove in this section that exact $k$-means applied to the columns of $\mb Q$ is equivalent, up to a permutation of the cluster labels, to that applied to the columns of $\mb M^{(r)}$. Recall that $\mb m^{(r)}_{(k,i)}$ denotes the column of $\mb M^{(r)}$ indexed by $(k,i)\in \mathcal I$. Consider the following  $k$-means problem on $\mb M^{(r)}$:

\begin{align}\label{def:hat_sigma}
        \wh{\mb c}_1,\ldots,\wh{\mb c}_r,\wh\sigma
    =
    \argmin_{\substack{\mb m_1,\ldots,\mb m_r\in\bb R^R\\ \sigma:\mathcal I \to [r]}}
    \sum_{(k,i)\in \mathcal I}
    \bigl\|\mb m^{(r)}_{(k,i)}-\mb m_{\sigma(k,i)}\bigr\|^2.
\end{align} 
 Let  $ \wt{\sigma}(\cdot)$ be given in \eqref{kmeans_spec}. The following lemma shows that $\wt \sigma$ is equivalent to $\wh \sigma$ up to a group permutation.  
\begin{lemma}\label{lem:rank_kmeans}
   There exists some permutation $\pi_0:[r]\to [r]$ such that $\pi_0(\wt{\sigma}(k,i))=\wh{\sigma}(k,i)$ for all $(k,i)\in \mathcal{I}$.
\end{lemma}
\begin{proof}
    For the $k$-means problem in \eqref{def:hat_sigma}, let 
    \[
        \wh \cC_a := \{(k,i)\in \mathcal I: \wh \sigma(k,i) = a\},\qquad  \forall ~ a\in [r].
    \]
    We start by noting that 
    \[
        \wh{\mb c}_a := {1\over |\wh \cC_a|}  \sum_{(k,i)\in \wh \cC_a} \mb m^{(r)}_{(k,i)},\qquad \forall ~ a\in [r]
    \]
    which implies   $\bU_r \bU_r^\T \wh{\mb c}_a  = \wh{\mb c}_a$ with $\bU_r \in \bbO_{R,r}$ containing the first $r$ eigenvectors of $\bM$. It therefore suffices to search  $\mb m_1,\ldots,\mb m_r$ in \eqref{def:hat_sigma} within the column space  of $\bU_r$. For any such $\mb m_1,\ldots,\mb m_r$ and any $\sigma: \mathcal I \to [r]$, one has
    \begin{align*}
        \sum_{(k,i)\in \mathcal I} \bigl\|\mb m^{(r)}_{(k,i)}-\mb m_{{\sigma}(k,i)}\bigr\|^2&= \sum_{(k,i)\in \mathcal I}\bigl\|\mb U_r^\top\mb m^{(r)}_{(k,i)}-\mb U_r^\top\mb m_{\sigma(k,i)}\bigr\|^2\nonumber\\
    &=\sum_{(k,i)\in \mathcal I}\bigl\|\bq_{(k,i)}-\mb U_r^\top \mb m_{\sigma(k,i)}\bigr\|^2. 
    \end{align*}
    which is equivalent to solving \eqref{kmeans_spec}, as by the same reasoning, the optimal $\wt \btheta_1,\ldots,\wt\btheta_r$ therein belong to the column space of $\bQ = \bU_r^\T \bM$. This completes the proof.
\end{proof}

\subsection{Proof of \cref{lem:k_means}}\label{app_sec_proof_lem:kmeans}

\begin{proof}
By \cref{lem:rank_kmeans}, it remains to analyze \eqref{def:hat_sigma} as once the desired bounds are proved for $\wh\sigma$, the same conclusions follow for $\wt\sigma$. For simplicity and also without loss of generality, we assume $\pi_0$ in \cref{lem:rank_kmeans} to be identity.  Recall the true partition corresponding to $\sigma^\star$ in \eqref{def_sigma_star}:
\[
    \cC_a^\star
    =
    \left\{
        (k, i) \in \mathcal{I}: \sigma^\star(k, i) = a
    \right\},
    \qquad |\cC_a^\star| = n_a,
    \qquad \forall ~ a\in [r].
\]
Recall \eqref{def_barn} and 
\[
    \mb M^\star= {1 \over \sqrt{\bar n}}~ \mb Z^\top\mb Z.
\]
It is easy to see that $\mb M^\star$ has only $r$ distinct columns, denoted by $\mb m_1^\star,\ldots,\mb m_r^\star\in\RR^{R}$, and these columns satisfy 
\begin{equation}\label{def_m_star_diff}
    \|\mb m_a^\star\|^2 = {n_a \over \bar n},\qquad   \|\mb m_a^\star-\mb m_b^\star\|^2= {n_a + n_b \over \bar n},\qquad\forall ~  a\ne b \in [r].
\end{equation}
We also note that 
\begin{equation}\label{eigen_M_star}
    {n_{\min} \over \sqrt{\bar n}} \le \lambda_r(\bM^\star) \le \lambda_1(\bM^\star) \le {n_{\max} \over \sqrt{\bar n}}.
\end{equation}
Moreover, for every $(k,i)\in \cC_a^\star$, one has   $\mb m^{\star}_{(k,i)} = \mb m_a^\star$. Further  recall from \eqref{def_eta_a}
\begin{align}
    \eta = {1\over \bar n}\sum_{a=1}^r \sum_{(k,i)\in \cC_a^\star}\|\mb m^{(r)}_{(k,i)}-\mb m_a^\star\|^2,
    \qquad
    \eta_a = \frac{1}{n_a}\sum_{(k,i)\in \cC_a^\star}\|\mb m^{(r)}_{(k,i)}-\mb m_a^\star\|^2,
    \qquad    a\in [r].\nonumber
\end{align}

\noindent We first analyze the estimated centers $\wh{\mb c}_1,\ldots,\wh{\mb c}_r$ by proving: for all $b\in [r]$,
\begin{align}\label{rate_center_raw}
    \min_{a\in [r]} \|\wh{\mb c}_{a} - \mb m_b^\star\| \le 2\sqrt{\bar n\over n_b}\sqrt{\eta}.
\end{align}
We prove this by contradiction:
Suppose there exists  $b\in [r]$ such that $\min_{a\in [r]}\|\wh{\mb c}_a-\mb m_b^\star\| >2\sqrt{(\bar n /  n_b)\eta}$. Then, by triangle inequality, we have
\begin{align}
   \sum_{(k,i)\in \cC^\star_b}  \min_{a\in [r]} \|\mb m^{(r)}_{(k,i)}  -  \wh{\mb c}_a\|^2
   &\ge \sum_{(k,i)\in \cC^\star_b} \min_{a\in [r]}
    \left(\frac{1}{2}\|\wh{\mb c}_a - \mb m_b^\star\|^2 -  \|\mb m^{(r)}_{(k,i)}-\mb m_b^\star\|^2 \right)\nonumber\\
    &>   2\bar n \eta -   \sum_{(k,i)\in \cC^\star_b} \|\mb m^{(r)}_{(k,i)}-\mb m_b^\star\|^2
    &&\text{by } |\cC^\star_b| = n_b\nonumber\\
    & \ge  \bar n \eta
    &&\text{by }\bar n \eta =  \sum_b n_b \eta_b.\label{ieq:upper_sum_distance}
\end{align}
However, this is a contradiction because one feasible solution to the $k$-means problem in \eqref{def:hat_sigma} is given by the centers $\mb m_1^\star,\ldots,\mb m_r^\star\in \RR^{R}$ with the corresponding assignment $\sigma^\star$, for which
\begin{align}\label{ieq:prof_lem:k_means}\nonumber
  \bar n\eta <  \sum_{(k,i)\in \cC^\star_b}  \min_{a\in [r]} \|\mb m^{(r)}_{(k,i)}  -  \wh{\mb c}_a\|^2 & \le \sum_{(k,i)\in \mathcal I}  \min_{a\in [r]} \|\mb m^{(r)}_{(k,i)}  -  \wh{\mb c}_a\|^2 \\\nonumber
     & \le \sum_{(k,i)\in \mathcal I}  \min_{a\in [r]} \|\mb m^{(r)}_{(k,i)}  -   \mb m_a^\star\|^2 \\\nonumber 
     &\le \sum_{b=1}^r \sum_{(k,i)\in \cC^\star_b} \|\mb m^{(r)}_{(k,i)} -\mb m_b^\star\|^2\\
     & = \bar n \eta.
\end{align}
This proves \eqref{rate_center_raw}.  Furthermore, we note that, under
$$
      \eta \le {1\over 8}{n_{\min}^2 \over  \bar n^2}. 
$$
and using  \eqref{def_m_star_diff} with triangle inequality, 
there must exist a permutation $\pi^\star:[r]\to[r]$ such that $\wh{\mb c}_{\pi^\star(a)}$ is closest to $\mb m_a^\star$ for all $a\in [r]$. 

 We proceed to prove \eqref{up_bd_sm}. Without loss of generality, we assume the above permutation $\pi^\star$ is the identity.
Pick any $a\ne b\in [r]$. Define
\begin{equation}\label{def_N_hat_ab}
    \wh N_{a\to b}
    :=
    \left\{
        (k,i)\in \mathcal I: \sigma^\star(k,i)=a,\ \wh\sigma(k,i)=b.
    \right\}
\end{equation}
as the set of points in $\cC^\star_a$ that are misclustered to the estimated cluster $b$. Note that
\[
    s_a(\wh\sigma)= \frac{1}{n_a}\sum_{b\in [r]\setminus \{a\}}|\wh N_{a\to b}|.
\]
For any $(k,i)\in \wh N_{a\to b}$, we have
\begin{align*}
    \|\mb m^{(r)}_{(k,i)} - \mb m_a^\star\|
    &\ge \|\mb m^{(r)}_{(k,i)} - \wh{\mb c}_a\| - \|\wh{\mb c}_a - \mb m_a^\star\|\\
    &\ge  \frac{1}{2}\|\wh{\mb c}_a - \wh{\mb c}_{b}\| - \|\wh{\mb c}_a - \mb m_a^\star\|\\
    &\ge  \frac{1}{2}\|\mb m_a^\star - \mb m_b^\star\| - 2 \left(\|\wh{\mb c}_a - \mb m_a^\star\| \vee \|\wh{\mb c}_{b}-\mb m_{b}^\star\|\right).
\end{align*}
where the second inequality is due to
\[
    \|\wh{\mb c}_a - \wh{\mb c}_{b}\|
    \le \|\mb m^{(r)}_{(k,i)} - \wh{\mb c}_a\| + \|\mb m^{(r)}_{(k,i)} - \wh{\mb c}_{b}\|
    \le 2\|\mb m^{(r)}_{(k,i)} - \wh{\mb c}_a\|.
\]
Since \eqref{rate_center_raw} and 
\[
     {\eta \le {1\over 128} {n_{\min}^2 \over  \bar n^2}}
\]
ensure that
\[
    \left(\|\wh{\mb c}_a - \mb m_{a}^\star\| \vee \|\wh{\mb c}_{b} - \mb m_{b}^\star\|\right)
    \le 2\sqrt{ \bar n\eta \over n_a \wedge n_b}
    \le \frac{1}{8}\|\mb m_{a}^\star-\mb m_{b}^\star\|,
\]
we conclude that
\begin{equation}\label{ieq:lower_devi_node_center}
    \|\mb m^{(r)}_{(k,i)} - \mb m_{a}^\star\| \ge \frac{1}{4}\|\mb m_{a}^\star-\mb m_{b}^\star\|.
\end{equation}
Therefore, by \eqref{def_eta_a},
\begin{align*}
    n_a \eta_a
    = \sum_{(k,i)\in\cC_a^\star}\|\mb m^{(r)}_{(k,i)} - \mb m_{a}^\star\|^2
    &\ge \sum_{(k,i)\in \bigcup_{b\in [r]\setminus \{a\}}\wh N_{a\to b}} \frac{1}{16}\|\mb m_{a}^\star-\mb m_{b}^\star\|^2\\
    &= \sum_{b \in [r]\setminus \{a\}} |\wh N_{a\to b}| ~  \frac{1}{16}\|\mb m_{a}^\star-\mb m_{b}^\star\|^2\\
    &= \frac{1}{16}\sum_{b \in [r]\setminus \{a\}} |\wh N_{a\to b}| {n_a + n_b \over \bar n} &&\text{by \eqref{def_m_star_diff}}.
\end{align*}
It then follows that
\[
    s_a(\wh\sigma)=\frac{1}{n_a}\sum_{b\in [r]\setminus \{a\}}|\wh N_{a\to b}|
    \le {8 \bar n\over n_{\min}}\,\eta_a.
\]
Finally, since $\wh\sigma$ and $\wt\sigma$ are identical up to a permutation of labels by \cref{lem:rank_kmeans}, the same bound also holds for $s_a(\wt\sigma)$. This completes the proof.
\end{proof}

\subsection{Proof of \cref{lem:spectral_kmeans}}\label{app_sec_proof_lem:spec}
\begin{proof}
    It remains to prove \eqref{ieq:eta_upperbound}, as
    \eqref{up_bd_sm} then follows immediately from \cref{lem:k_means} and 
    \begin{equation}\label{def_cond_eps}
        32\sqrt{2\epsilon}  ~ \le ~  {n_{\min} \over \sqrt{\bar n  n_{\max} }}.
    \end{equation}
    To bound from above $\eta$, by definition in \eqref{def_eta_a} and using the fact that $\bM^{(r)}$ is the best rank-$r$ approximation to $\bM$ under the Frobenius norm, we obtain 
    \begin{align*}
    \sqrt{\eta} =\frac{1}{\sqrt{\bar n}}\|\mb M^{(r)}-\mb M^\star\|_F &\le \frac{1}{\sqrt{\bar n}}\|\mb M^{(r)}-\mb M \|_F + \frac{1}{\sqrt{\bar n}}\|\mb M -\mb M^\star\|_F  \leq \frac{2}{\sqrt{\bar n}}\|\mb M-\mb M^\star\|_F.
    \end{align*}
    Write
    \begin{align}\label{def:delta_c}
         \bs\Delta_c:=\wt\bA_c-\bA^\star \bZ \mb\Xi \in \RR^{r\times R}
    \end{align}
    so that, by \eqref{def_epsilon_all}, $\|\bs\Delta_c\|_F^2 = \bar n\epsilon$. By observing that 
    \begin{align}
 \frac{2}{\sqrt{\bar n}}\|\mb M-\mb M^\star\|_F&=\frac{2}{\bar n}\left\|
     \abs{\wt{\mb A}_c^\top\wt{\mb A}_c}- \abs{\mb \Xi\mb Z^\top\mb Z\mb \Xi}
    \right\|_F \nonumber\\
    &\le
    \frac{2}{\bar n}\left(
   2 \Bigl\|
    \bs\Delta_c^\top\mb A^\star\mb Z\mb \Xi 
    \Bigr\|_F + \Bigl\| \bs\Delta_c^\top\bs\Delta_c
    \Bigr\|_F 
    \right) \nonumber\\
    &\le
    \frac{2}{\bar n}
    \Bigl(
    2\|\bs\Delta_c\|_F\|\mb Z\|_{\op}
    +
    \|\bs\Delta_c\|_F^2
    \Bigr) \nonumber\\
    &\le
    \frac{4 \sqrt{n_{\max}}}{\bar n}\|\bs\Delta_c\|_F+\frac{2}{\bar n}\|\bs\Delta_c\|_F^2,
\end{align}
the proof is complete. 
\end{proof}

\subsection{Additional results on the k-means clustering}

For $a\in [r]$,
\cref{lem:spectral_kmeans,lem:k_means} are concerned about $s_a(\wt \sigma)$, which controls the proportion of misclustered points in $\cC_a^\star$. 
The following lemma provides additional guarantees on the proportion of misclustered points among $\wt \cC_a^\star$. Recall that $\pi^\star$ minimizes \eqref{def:spec_esti}. Let 
\begin{equation}\label{def:rho_e}
    \rho_\epsilon := 32 {\bar n^2 \over n_{\min}}  \left(2 \sqrt{\epsilon ~ {n_{\max}\over \bar n}}+\epsilon\right)^2.
\end{equation}

\begin{lemma}\label{lem_misclustering}
Under condition \eqref{ieq:cond_epsilon}, for each $a\in [r]$, one has 
\[
    |\wt \cC_{\pi^\star(a)} \setminus \cC_a^\star| \le  \rho_\epsilon 
\]
as well as 
\[
    |\cC_a^\star\setminus \wt \cC_{\pi^\star(a)} | \le \rho_\epsilon.
\]
\end{lemma}
\begin{proof}
    Assuming the relevant permutations to be identity, the first statement follows from 
    \begin{align*}\nonumber
   |\wt \cC_{a} \setminus \cC_a^\star|=  \sum_{b\in [r]\setminus\{a\}} \sum_{(k,i)\in \cC_b^\star} 1\{\wt\sigma(k,i) = a\} & \overset{\eqref{def:s_max}}{\le } \sum_{b\in [r]\setminus\{a\}} n_b ~ s_b(\wt \sigma)\\\nonumber
    &\le  {8\bar n\over n_{\min}}   \sum_{b\in [r]\setminus\{a\}} n_b ~ \eta_b &&\text{by \cref{lem:spectral_kmeans}}\\\nonumber 
    &\le  {8\bar n^2 \over n_{\min} }  \eta &&\text{by \eqref{def_eta_a}}\\
    &\le   \rho_\epsilon &&\text{by \eqref{ieq:eta_upperbound}}.
\end{align*} 
For the second statement, observe that 
\begin{align*}
    |\cC_a^\star \setminus  \wt \cC_{a} |= \sum_{b\in [r]\setminus\{a\}} \sum_{(k,i)\in \cC_a^\star} 1\{\wt\sigma(k,i) = b\} = n_a s_a(\wt \sigma) \le {8\bar n\over n_{\min}}   n_a ~ \eta_a \le   {8\bar n^2 \over n_{\min} }  \eta.
\end{align*}
This completes the proof. 
\end{proof}

\subsection{Proof of \cref{prop:sign_alignment}}\label{app_sec_proof_lem_sign}

The statement in  (i) is the result of the following lemma which shows that any local estimator $\wt\ba_i^{(k)}$ with its error $\epsilon^{(k)}_i$ satisfying \eqref{cond_eps_ik} is correctly clustered by $\wt \sigma$ in \eqref{kmeans_spec}. 

\begin{lemma}[Accurate local estimators are correctly clustered]
\label{lem:A_good_not_misclustered}
Under condition \eqref{ieq:cond_epsilon}, there exists some permutation  $\pi:[r]\to[r]$ such that for any atom $(k,i)\in \cI$ with its error satisfying \eqref{cond_eps_ik}, one has  
\[
    \sigma^\star(k,i)=\pi(\wt\sigma(k,i)).
\] 
\end{lemma}
\begin{proof}
Pick any atom $(k,i)$ and let $a:=\sigma^\star(k,i)$.   We aim to prove $a = \pi^\star(\wt \sigma(k,i))$. By \cref{lem:rank_kmeans}, it suffices to show this holds for $\wh\sigma$ given in \eqref{def:hat_sigma} up to the permutation $\pi^\star \circ \pi_0$. 
Without loss of generality, we assume both permutations  to be the identity, and prove that  $\wh \sigma(k,i) = a$. It suffices to show that $\mb m^{(r)}_{(k,i)}$ is strictly closer to the estimated center  $\wh{\mb c}_a$
 than to any other estimated centers. 
 
Recall  $\mb M^\star$ in \eqref{def_M_star} and let
$\mb m^\star_a$ denote the population column corresponding to cluster
$a$. Let $\bU^\star$ contain the  $r$ eigenvectors of $\bM^\star$. 

We first upper bound the distance between  
$\mb m^{(r)}_{(k,i)}$ and its population counterpart $\mb m^\star_a$. Since $\bM^{(r)} = \bU_r\bU_r^\T \bM$, we find 
\begin{equation}
\begin{aligned}
    \|\mb m^{(r)}_{(k,i)}-\mb m^\star_a\| & = \|\bU_r \bU_r^\T \mb m_{(k,i)}-\bU^\star \bU^{\star\T}\mb m^\star_a\|\\
    &\le
    \|\bU_r \bU_r^\T (\mb m_{(k,i)}-\mb m^\star_a)\|
    +
    \|(\bU_r \bU_r^\T -\bU^\star \bU^{\star\T}) \mb m^\star_a\|  \\
    &\le
    \|\mb m_{(k,i)}-\mb m^\star_a\|
    +
    \|\bU_r \bU_r^\T -\bU^\star \bU^{\star\T}\|_{\op} \|\mb m^\star_a\|.
\end{aligned}
\label{eq:Mr_column_decomp}
\end{equation}
By 
invoking the Davis-Kahan theorem, we obtain
\begin{align}\nonumber
   \|\bU_r \bU_r^\T -\bU^\star \bU^{\star\T}\|_{\op} \|\mb m^\star_a\|
    ~ &\le 
    {2\|\mb M-\mb M^\star\|_{\op} \over \lambda_r(\bM^\star)} \sqrt{n_a \over \bar n}\\\nonumber
    &\le {2\sqrt{\bar n \eta} \over n_{\min} /\sqrt{\bar n}} \sqrt{n_a\over \bar n} &&\text{by  \eqref{def_m_star_diff} \& \eqref{eigen_M_star}}\\
    &\le   \underbrace{\left(8\sqrt{n_{\max}\epsilon}+4\epsilon\sqrt{\bar n}\right)\frac{\sqrt{n_a}}{n_{\min}}}_{:=~ \omega_a(\epsilon)}  &&\text{by }\eqref{ieq:eta_upperbound}.
    \label{eq:best_rank_residual}
\end{align}
To find a bound for
$\|\mb m_{(k,i)}-\mb m^\star_a\|$, recall  that
\[
   \mb M
    =
    \frac{1}{\sqrt{\bar n}}
    \left|
        \wt{\mb A}_c^\top \wt{\mb A}_c
    \right|,
    \qquad  \wt{\mb A}_c
    =
    [\wt{\mb A}^{(1)},\ldots,\wt{\mb A}^{(K)}]. 
\]  
Let $\bs\delta_c(k,i)$ denote the column  corresponding to atom $(k,i)$ of $\bs\Delta_c$, given in \eqref{def:delta_c}. We have 
\[
    \wt{\mb a}^{(k)}_i
    =
    \xi^\star_{(k,i)}~ \ba^\star_a
    +
    \bs\delta_c(k,i),\qquad \text{with}\quad \|\bs\delta_c(k,i)\|
    =
    \epsilon_i^{(k)}.
\]
This gives  
\[
\begin{aligned}
    \|\mb m_{(k,i)}-\mb m^\star_a\|
    &=
    \frac{1}{\sqrt {\bar{n}}}
    \left\|
        \bigl|
            \wt{\mb A}_c^\top
            \wt{\mb a}^{(k)}_i
        \bigr|
        -
        \left|
             \bA_c^{\star\top}
            \xi^\star_{(k,i)}\ba^\star_a
        \right|
    \right\|   \le
    \frac{1}{\sqrt {\bar{n}}}
    \left\|
        \wt{\mb A}_c^\top
        \wt{\mb a}^{(k)}_i
        -
        \bA_c^{\star\top}
        \xi^\star_{(k,i)}\ba^\star_a
    \right\| .
\end{aligned}
\]
Now noting that
\[
\begin{aligned}
    \wt{\mb A}_c^\top
    \wt{\mb a}^{(k)}_i
    -
    \bA_c^{\star\top}
    \xi^\star_{(k,i)}\ba^\star_a
    &=
    \wt{\mb A}_c^\top
    \left(
        \xi^\star_{(k,i)}\ba^\star_a
        +
        \bs\delta_c(k,i)
    \right)
    -
    \bA_c^{\star\top}
    \xi^\star_{(k,i)}\ba^\star_a  \\
    &=
    \wt{\mb A}_c^\top \bs\delta_c(k,i)
    +
    (\wt{\mb A}_c-\bA_c^\star)^\top
    \xi^\star_{(k,i)}\ba^\star_a  \\
    &=
    \wt{\mb A}_c^\top \bs\delta_c(k,i)
    +
    \bs\Delta_c^\top
    \xi^\star_{(k,i)}\ba^\star_a &&\text{by \eqref{def:delta_c}},
\end{aligned}
\]
we conclude 
\[
\begin{aligned}
    \|\mb m_{(k,i)}-\mb m^\star_a\|
    &\le
    \frac{1}{\sqrt {\bar{n}}}
    \|\wt{\mb A}_c^\top \bs\delta_c(k,i)\|
    +
    \frac{1}{\sqrt {\bar{n}}}
    \|\bs\Delta_c^\top  \ba^\star_a\|  \le
    \frac{\|\wt{\mb A}_c\|_{\op}}{\sqrt{\bar n}}
    \|\bs\delta_c(k,i)\|
    +
    \frac{\|\bs\Delta_c\|_{\op}}{\sqrt{\bar n}}
    \|\ba^\star_a\| .
\end{aligned}
\]
Since $\mb A^\star$ is orthogonal with $\|\ba^\star_a\|=1$, we obtain
\[
    \frac{\|\bs\Delta_c\|_{\op}}{\sqrt{\bar n}}\le \sqrt{\epsilon},
    \qquad
    \frac{\|\wt{\mb A}_c\|_{\op}}{\sqrt{\bar n}}\leq\frac{\|\bA_c^\star\|_{\op}+\|\bs\Delta_c\|_{\op}}{\sqrt{\bar n}}
    \le
    \sqrt{\frac{n_{\max}}{\bar n}}+\sqrt{\epsilon}
\]
so that  
\begin{equation}
    \|\mb m_{(k,i)}-\mb m^\star_a\|
    \le
    (\sqrt{\frac{n_{\max}}{\bar n}}+\sqrt{\epsilon})\epsilon_i^{(k)}
    +
    \sqrt{\epsilon}.
    \label{eq:M_column_bound_by_delta}
\end{equation}
Combining \eqref{eq:Mr_column_decomp}, \eqref{eq:best_rank_residual}, and
\eqref{eq:M_column_bound_by_delta} yields
\begin{equation}
    \|\mb m^{(r)}_{(k,i)}-\mb m^\star_a\|
    \le
    (\sqrt{\frac{n_{\max}}{\bar n}}+\sqrt{\epsilon})\epsilon_i^{(k)}
    +
    \sqrt{\epsilon}
    +
    \omega_a(\epsilon) .
    \label{eq:Mr_column_bound_by_delta}
\end{equation}
Further invoking   the  
perturbation bound for centers of the k-means in \eqref{rate_center_raw} yields
\begin{equation}
    \left\|
        \wh{\mb c}_{a}
        -
        \mb m^\star_a
    \right\|
    \le 2\sqrt{\bar n\over n_a}\sqrt{\eta}
   \le 
    \frac{n_{\min}}{n_a}\omega_a(\epsilon).
    \label{eq:center_bound_Mr}
\end{equation}
Together, we have 
\[
\begin{aligned}
    \|
        \mb m^{(r)}_{(k,i)}
        -
        \wh{\mb c}_a
    \|
    &\le
    \|\mb m^{(r)}_{(k,i)}-\mb m^\star_a\|
    +
    \|\mb m^\star_a-\wh{\mb c}_a\|  \le
   (\sqrt{\frac{n_{\max}}{\bar n}}+\sqrt{\epsilon})\epsilon_i^{(k)}
    +
    \sqrt{\epsilon}
    +
    \omega_a(\epsilon) +\frac{n_{\min}}{n_a}\omega_a(\epsilon).
\end{aligned}
\]

On the other hand, for any $b\neq a$, the population centers satisfy
\[
    \|\mb m^\star_a-\mb m^\star_b\|=\sqrt{{n_a + n_b \over \bar n}} .
\]
Triangle inequality leads to
\[
\begin{aligned}
    \|
        \mb m^{(r)}_{(k,i)}
        -
        \wh{\mb c}_b
    \|
    &\ge
    \|\mb m^\star_a-\mb m^\star_b\|
    -
    \|\mb m^{(r)}_{(k,i)}-\mb m^\star_a\|-\|\mb m^\star_b-\wh{\mb c}_b\| \\
    &\ge
    \sqrt{{n_a + n_b \over \bar n}}-(\sqrt{\frac{n_{\max}}{\bar n}}+\sqrt{\epsilon})\epsilon_i^{(k)}
    -
    \sqrt{\epsilon}
    -
    \omega_a(\epsilon)-\frac{n_{\min}}{n_b}\omega_b(\epsilon) .
\end{aligned}
\]
Thus, the atom $(k,i)$ is correctly clustered if 
\[
    (\sqrt{\frac{n_{\max}}{\bar n}}+\sqrt{\epsilon})\epsilon_i^{(k)}
    +
    \sqrt{\epsilon}
    +
    \left(1+\frac{n_{\min}}{2n_a}\right)\omega_a(\epsilon)+\frac{n_{\min}}{2n_b}\omega_b(\epsilon)
    <
    \frac{1}{2}\sqrt{{n_a + n_b \over \bar n}}.
\]
Using  condition \eqref{ieq:cond_epsilon} and the  definition of $\omega_a(\epsilon)$ in \eqref{eq:best_rank_residual}, a sufficient condition is
\begin{align}
    (\sqrt{\frac{n_{\max}}{\bar n}}+\sqrt{\epsilon}) \epsilon_i^{(k)}~ <\ \frac{1}{2}\sqrt{{n_a+n_b}\over \bar{n}}-\sqrt{c\over 2}\sqrt{{n_a+n_b}\over \bar{n}}-c'\left(\sqrt{n_a\over \bar {n}}+\sqrt{n_b\over \bar {n}}\right) 
\end{align} 
for some constant $c'$ only  depending on $c$ defined in \eqref{ieq:cond_epsilon}. Here we also use $\epsilon<c n_{\min}/\bar{n}<c(n_a+n_b)/(2\bar n)$. Since this condition is satisfied by \eqref{cond_eps_ik} together with choosing 
sufficient small $c$ in \eqref{ieq:cond_epsilon},  the proof is complete. 
\end{proof} 

\begin{proof}[Proof of (ii) in \cref{prop:sign_alignment}]
Pick any $a\in [r]$. For any $(k,i)\in \cC^\star_a$ such that  
\[
    \wt{\mb a}^{(k)}_i
    =
    \xi^\star_{(k,i)}\ba^\star_a+\mb e_i^{(k)}
\]
where $\|\mb e_i^{(k)}\|= \epsilon_i^{(k)}$ satisfies \eqref{cond_eps_ik}, \cref{lem:A_good_not_misclustered} ensures that $(k,i) \in \wt \cC_{\pi^\star(a)} = \wt \cC_a$ by assuming $\pi^\star (a) = a$ for simplicity. Then if we can show that 
\[
    \left|
    \left\langle
        \wt{\mb u}_a,\ba^\star_a
    \right\rangle
    \right|
    ~ \overset{(*)}{>}~  \epsilon_i^{(k)} = 
    \|\mb e_i^{(k)}\|
\]
one can deduce that 
\[
\begin{aligned}
    \operatorname{sgn}
    \bigl(
        \bigl\langle
            \wt{\mb u}_a,\wt{\mb a}^{(k)}_i
        \bigr\rangle
    \bigr)  =
    \xi^\star_{(k,i)}
    \operatorname{sgn}\left(
        \left\langle
            \wt{\mb u}_a,\ba^\star_a
        \right\rangle
    \right)
    =:
    \xi^\star_{(k,i)}~ \varsigma_a .
\end{aligned}
\]
As a result, by the definition of the sign-aligned atom,
\[
\begin{aligned}
    \widecheck{\mb a}^{(k)}_i =
    \operatorname{sgn}
    \bigl(
        \bigl\langle
            \wt{\mb u}_a,\wt{\mb a}^{(k)}_i
        \bigr\rangle
    \bigr)
    \wt{\mb a}^{(k)}_i                                   
    &=
    \xi^\star_{(k,i)} \operatorname{sgn}\left(
        \left\langle
            \wt{\mb u}_a,\ba^\star_a
        \right\rangle
    \right)
    \left(
        \xi^\star_{(k,i)} \ba^\star_a+\mb e_i^{(k)}
    \right)                                                     \\
    &=
    \operatorname{sgn}\left(
        \left\langle
            \wt{\mb u}_a,\ba^\star_a
        \right\rangle
    \right) \left( \ba^\star_a+\xi^\star_{(k,i)}  \mb e_i^{(k)}\right) .
\end{aligned}
\]
so that the claim in $(ii)$ follows:
\[
    \bigl\|
        \widecheck{\mb a}^{(k)}_i
        -
        \operatorname{sgn}\left(
        \left\langle
            \wt{\mb u}_a,\ba^\star_a
        \right\rangle
    \right) \ba^\star_a
    \bigr\|
    =
    \|\mb e_i^{(k)}\|
    =
    \widecheck{\epsilon}_i^{(k)}.
\]

To prove $(*)$, we bound from below $\left|
    \left\langle
        \wt{\mb u}_a,\ba^\star_a
    \right\rangle
    \right|$. 
Recall  $\wt{\mb u}_a$ is the leading left singular vector of 
$
    \wt{\mb A}_{\wt\cC_a}:=
    [\wt{\mb a}^{(k)}_i: (k,i)\in\wt\cC_a] 
$, equivalently
the leading eigenvector of $ \wt{\mb A}_{\wt\cC_a} \wt{\mb A}_{\wt\cC_a}^\top$. We first bound the difference of this matrix from its population-level counterpart
$n_a \ba^\star_a \ba_a^{\star\top}$. Split the set of $\wt\cC_a$ into two parts
\begin{align}\label{def:G_a_B_a}
        G_a:=\wt\cC_a\cap \cC^\star_{a},
    \qquad
    B_a:=\wt\cC_a\setminus \cC^\star_{a},\qquad 
    D_a:=\cC^\star_{a}\setminus \wt\cC_a.
\end{align}
to obtain 
\begin{align*} 
    \left\|
         \wt{\mb A}_{\wt\cC_a} \wt{\mb A}_{\wt\cC_a}^\top
        -
        n_a \ba^\star_a \ba_a^{\star\top}
    \right\|_{\op} &\le~  \left\|
         \sum_{(k,i)\in G_a}
\wt{\mb a}^{(k)}_i
\wt{\mb a}_i^{(k)\top}
-
|G_a|\ba^\star_a \ba_a^{\star\top} 
    \right\|_{\op} + \left\|
        \sum_{(k,i)\in B_a}
\wt{\mb a}^{(k)}_i
\wt{\mb a}_i^{(k)\top} 
    \right\|_{\op}+\left\|
|D_a|\ba^\star_a \ba_a^{\star\top}\right\|_{\op}.
\end{align*}
The first term on the right-hand-side is bounded from above by  
\[ 
2\sum_{(k,i)\in G_a}\|\mb e_i^{(k)}\|
+
\sum_{(k,i)\in G_a}\|\mb e_i^{(k)}\|^2  \le ~ 
2n_a\sqrt{\epsilon_a}+n_a\epsilon_a
\]
where we use the definitions of $\epsilon^{(k)}_i$ in \eqref{eqn:def_data_A} and $\epsilon_a$ in \eqref{def_epsilon_ell} together with the Cauchy-Schwarz inequality.
For the other two terms, note that  
\[
    \left\|
        \sum_{(k,i)\in B_a}
\wt{\mb a}^{(k)}_i
\wt{\mb a}_i^{(k)\top} 
    \right\|_{\op} \le  |B_a|,\qquad\textit{and}\quad\left\|
|D_a|~\ba^\star_a \ba_a^{\star\top}\right\|_{\op}=\abs{D_a}.
\] 
Since
\cref{lem_misclustering} gives $|B_a| \le  \rho_\epsilon$ and $|D_a| \le  \rho_\epsilon$
with $\rho_\epsilon$ given in \eqref{def:rho_e},
we conclude that  
\[
    \left\|
         \wt{\mb A}_{\wt\cC_a} \wt{\mb A}_{\wt\cC_a}^\top
        -
        n_a \ba^\star_a \ba_a^{\star\top}
    \right\|_{\op}
     \le ~ 
     2n_a \sqrt{\epsilon_a}+n_a \epsilon_a+2\rho_\epsilon.
\]
Note that condition \eqref{ieq:cond_epsilon} ensures 
\begin{align*}
    \epsilon_a < \frac{\bar n}{n_a} \epsilon <   c\frac{n_{\min}}{n_{\max}}< c
\end{align*}
as well as 
\begin{align*}
    {\rho_\epsilon\over n_a} &= {32\bar n^2 \over n_{\min} n_a}  \left(2 \sqrt{\epsilon ~ {n_{\max}\over \bar n}}+\epsilon\right)^2    %
     \leq c',
\end{align*} 
we have
\[
    \left\|
        {1\over n_a} \wt{\mb A}_{\wt\cC_a} \wt{\mb A}_{\wt\cC_a}^\top
        -
        \ba^\star_a \ba_a^{\star\top}
    \right\|_{\op}
    <
    {\sqrt{3}\over 4}.
\] 
Since the Davis--Kahan's theorem yields
\[
    \sin\angle(\wt{\mb u}_a,\ba^\star_a)
    \le
    2\left\|
        \frac{1}{n_a} \wt{\mb A}_{\wt\cC_a} \wt{\mb A}_{\wt\cC_a}^\top
        -
        \ba^\star_a \ba_a^{\star\top}
    \right\|_{\op}
    <
    {\sqrt{3}\over 2},
\]
we obtain 
\[
    \left|
        \left\langle
            \wt{\mb u}_a,\ba^\star_a
        \right\rangle
    \right|
    >
    \sqrt{1-{3\over 4}}
     =
    \frac12 > \epsilon^{(k)}_i
\]
as claimed in $(*)$. This completes the proof.  
\end{proof}

\section{Proofs of \cref{sec_theory_GM,sec:combined}}\label{app_sec_proof_GM}

\subsection{Proof of \cref{lem:quantile_preservation_sign}}\label{sec:proof_quantile_preservation_sign}

\begin{proof} 
By the definition of the empirical $p$-quantile, at least a $p$-fraction of atoms in $\wt{\mathcal C}_a$ satisfy
\[
    \epsilon_i^{(k)}\le \wt Q_a(p).
\]
Since $\wt Q_a(p)<c_0 \sqrt{n_{\min} / \bar n}$, invoking \cref{prop:sign_alignment} and \cref{lem:A_good_not_misclustered} implies that all these $p$-fraction of atoms are correctly clustered and sign-aligned. In particular,  for all such atoms, \eqref{def:check_epsilon_sign} gives
\[
    \widecheck\epsilon_i^{(k)}
    =
    \epsilon_i^{(k)}
    \le \wt Q_a(p).
\]
Hence at least a $p$-fraction of atoms in $\wt{\mathcal C}_a$ have sign-aligned error no larger than $\wt Q_a(p)$, which gives
\begin{align}\label{ieq:upper_check_bar_quantile}
        \widecheck{Q}_a(p)\le \wt Q_a(p).
\end{align}
This completes the proof.
\end{proof}

\subsection{Proof of \cref{lem:mis_class}}\label{app_sec_proof_lem:mis_class}
\begin{proof}
Pick  any $a\in[r]$. The definition of $G_a$, $B_a$ in \eqref{def:G_a_B_a} and that of $\rho_\epsilon$ in \eqref{def:rho_e} gives that
\[
\begin{aligned}
    |\wt{\cC}_a|
    &=
    |G_a|
    +
    |B_a|   \leq n_a(1+\frac{\rho_\epsilon}{n_a}).
\end{aligned}
\]  
By the definition of the $p$-th sample quantile, at least $pn_a$ atoms in  $\cC_a^\star$ satisfy
$
    \epsilon_i^{(k)}\le Q_a(p) .
$
Since $ Q_a(p)<c_0 \sqrt{n_{\min} / \bar n}$, \cref{lem:A_good_not_misclustered} implies that all
these atoms are correctly classified by $\wt\sigma$, and hence they belong
to $\wt{\cC}_a$ (assuming $\pi^\star$ being identity). Therefore, inside $\wt{\cC}_a$, at least $pn_a$ atoms have
 estimation errors no larger than $Q_a(p)$. Since
\[
    |\wt{\cC}_a|\le n_a(1+\frac{\rho_\epsilon}{n_a}),
\]
these atoms form at least a fraction
\[
    \bar p_{a}:=\frac{p}{1+ \rho_\epsilon/ n_a}
\]
of the estimated cluster, implying 
\[
    \wt Q_a(\bar p_a)\le  Q_a(p).
\]
Now using 
$
    \wt Q_a(\bar p_a)<c_0 \sqrt{n_{\min} / \bar n}
$
and applying \cref{lem:quantile_preservation_sign} give 
\[
    \widecheck Q_a(\bar p_a) \le \wt Q_a(\bar p_a)\le Q_a(p).
\]
Since the above holds for all $\bar p  = p/(1+\delta_\epsilon) \le \bar p_a$,   the proof is complete.
\end{proof}

\subsection{Proof of \cref{thm:recover}}\label{app_sec_proof_thm:recover}
\begin{proof}
Pick any $a\in [r]$ with $p_a$ given in \eqref{def:optimal_p_i}.
   It suffices to show that  
\begin{align*}  
    \bigl\|\wh\ba_a - \varsigma_a\mb a^\star_{\pi(a)}\bigr\| \leq \frac{2 \bar p_a}{2\bar p_a - 1} Q_{a}(p_a).
\end{align*}  
for some permutation $\pi$ and all 
$1/2 < \bar p_a \le p_a /  (1+\delta_\epsilon)$.
Without loss of generality, we assume $\pi$ to be identity. Recall the definition of $\wh\bA$ and $\widecheck{Q}_{a}(p)$. Invoking 
\cref{lem:gm_robust} gives
\begin{align}\label{ieq:upper_q}
   \bigl\|\wh\ba_a - \varsigma_a\mb a^\star_{a}\bigr\|\leq \inf_{p\in (1/2, 1]}\frac{2p}{2p-1}\widecheck{Q}_{a}(p)\le \frac{2\bar p_a}{2\bar p_a-1}\widecheck{Q}_{a}(\bar p_a).
\end{align}
 Further invoking \cref{lem:mis_class} yields 
 \begin{align*}
    \widecheck{Q}_{a}(\bar p_a)\leq Q_a(p_a),
\end{align*}
which, together with  \eqref{ieq:upper_q}, completes the proof.
\end{proof}

\subsection{Proof of \cref{thm:kmeans_center}}\label{sec:proof_kmeans_center}

\begin{proof}
We first recall the results we have proved for the k-means clusters $\wt \cC_1,\ldots, \wt\cC_r$. Assuming $\pi^\star$ to be identity,  for each $a\in [r]$, one has 
\[
    G_a:=\wt{\cC}_a\cap \cC_a^\star,
    \qquad
    B_a:=\wt{\cC}_a\setminus \cC_a^\star.
\]
Recall  $\rho_\epsilon$ from \eqref{def:rho_e}. \cref{lem_misclustering} gives 
\begin{equation}\label{lb_td_C_a}
    |B_a|
    \le
    \rho_\epsilon,
    \qquad
    |\wt{\cC}_a| \ge |G_a| \ge n_a - |\cC^\star_a \setminus \wt \cC_a| 
    \ge
    n_a(1-\frac{\rho_\epsilon}{n_a}).
\end{equation}
Since atoms in $G_a$ may still have different signs from the oracle ones after the sign-alignment step, we define
\begin{equation}\label{def_H_a}
    H_a
    :=
    \left\{
        (k,i)\in G_a:
        \epsilon_i^{(k)}<\sqrt{n_{\min}\over 4\bar n}
    \right\} 
\end{equation}
so that $H_a$ contains the correctly clustered atoms which are also sign aligned.   \cref{prop:sign_alignment} thus gives  
\[
    \bigl\|
        \widecheck{\ba}^{(k)}_i - \varsigma_a\ba_a^\star
    \bigr\|
    = \epsilon_i^{(k)},
    \qquad
    (k,i)\in H_a .
\]
Let $R_a:=G_a\setminus H_a$.  For each $(k,i) \in R_a\cup B_a$, by triangular inequality, we have
\[
    \bigl\|
        \widecheck{\ba}^{(k)}_i
        -
        \varsigma_a\ba_a^\star
    \bigr\|
    \le
    \bigl\|\widecheck{\ba}^{(k)}_i\bigr\|
    +
    \bigl\|\ba_a^\star\bigr\|
    =
    2 .
\]
Using the definition
\[
    \bar\ba_a
    =
    \frac{1}{|\wt{\cC}_a|}
    \sum_{(k,i)\in\wt{\cC}_a}
    \widecheck{\ba}^{(k)}_i
\]
gives
\begin{align} \label{bd_bar_a_diff}
    \bigl\|
        \bar\ba_a-\varsigma_a\ba_a^\star
    \bigr\|
    &\le
    \frac{1}{|\wt{\cC}_a|}
    \sum_{(k,i)\in H_a}
    \bigl\|
        \widecheck{\ba}^{(k)}_i
        -
        \varsigma_a\ba_a^\star
    \bigr\|   +
    \frac{1}{|\wt{\cC}_a|}
    \sum_{(k,i)\in R_a\cup B_a}
    \bigl\|
        \widecheck{\ba}^{(k)}_i
        -
        \varsigma_a\ba_a^\star
    \bigr\|  \\\nonumber
    &\le
    \frac{1}{|\wt{\cC}_a|}
    \sum_{(k,i)\in H_a}
    \epsilon_i^{(k)}
    +
   \frac{ 2}{|\wt{\cC}_a|}(|R_a|+|B_a|).
\end{align} 
By Cauchy-Schwarz inequality,
\[
\begin{aligned}
    \frac{1}{|\wt{\cC}_a|}
    \sum_{(k,i)\in H_a}
    \epsilon_i^{(k)}
    &\le
    \frac{\sqrt{|H_a|}}{|\wt{\cC}_a|}
    \left(
        \sum_{(k,i)\in H_a}
        \left(\epsilon_i^{(k)}\right)^2
    \right)^{1/2} \\ 
    &\le
    \left(
        \frac{1}{|\wt{\cC}_a|}
        \sum_{(k,i)\in H_a}
        \left(\epsilon_i^{(k)}\right)^2
    \right)^{1/2}\\
    &\le \left(
        \frac{1}{|\wt{\cC}_a|}
        \sum_{(k,i)\in \cC^\star _a}
    \left(\epsilon_i^{(k)}\right)^2
    \right)^{1/2}\\
    &\le \left(
        \frac{n_a}{|\wt{\cC}_a|}\epsilon_a
    \right)^{1/2} &&\text{by \eqref{def_epsilon_ell}}.
\end{aligned}
\]
Together with  $|\wt{\cC}_a|\ge n_a - \rho_\epsilon$, we obtain
\[
    \frac{1}{|\wt{\cC}_a|}
    \sum_{(k,i)\in H_a}
    \epsilon_i^{(k)}
    \le
    \sqrt{
        \frac{\epsilon_a}{1-\rho_\epsilon/n_a}
    } .
\]

Regarding the atoms in $R_a$, note that for every $(k,i)\in R_a$, we must have
\[
    \epsilon_i^{(k)}
    \ge
    \sqrt{n_{\min}\over 8\bar{n}}.
\]
Therefore,
\[
    {n_{\min}\over 8\bar{n}} |R_a|
    \le
    \sum_{(k,i)\in R_a}
    \left(\epsilon_i^{(k)}\right)^2 .
\]
Since $R_a\subseteq\cC^\star_a$,  by the definition of $\epsilon_a$ in \eqref{def_epsilon_ell},
\[
    \sum_{(k,i)\in R_a}
    \left(\epsilon_i^{(k)}\right)^2
    \le
    \sum_{(k,i)\in\mathcal C_a^\star}
    \left(\epsilon_i^{(k)}\right)^2
    \le
    n_a\epsilon_a 
\]
implying
\[
    |R_a|\le {8\bar{n}n_a\over n_{\min}} \epsilon_a.
\]
Combining this with the bound $|B_a|\le \rho_\epsilon$ and
$|\wt{\cC}_a|\ge n_a -\rho_\epsilon$ gives
\[
    \frac{2}{|\wt{\cC}_a|}(|R_a|+|B_a|)
    \le
     \frac{2}{1- \rho_\epsilon/n_a}\left({8\bar{n}\over n_{\min}}\epsilon_a+\frac{\rho_\epsilon}{n_a}\right).
\]
Collecting the above bounds  yields
\begin{align}\label{bd_bar_a_diff_final}
        \bigl\|
        \bar\ba_a-\varsigma_a\ba_a^\star
    \bigr\|
   ~  &\le ~ 
   \sqrt{
        \frac{\epsilon_a}{1-\rho_\epsilon/n_a}
    } 
    +
    \frac{2}{1-\rho_\epsilon/n_a}\left({8\bar{n}\over n_{\min}}\epsilon_a+\frac{\rho_\epsilon}{n_a}\right)\nonumber\\
    &\lesssim ~  \sqrt{\epsilon_a} +\frac{\bar{n}n_{\max}}{n_{\min}^2}\epsilon+\frac{\bar{n}^2}{n_{\min}^2}\epsilon^2  ~ \lesssim \sqrt{\epsilon_a} +\frac{\bar{n}n_{\max}}{n_{\min}^2}\epsilon
\end{align}
where in the last step we use \eqref{def:rho_e}, \eqref{ieq:cond_epsilon} and $\epsilon_a\leq (\bar{n}/n_a)\epsilon$ to collect terms. %
This completes the proof.
\end{proof}

\section{Proofs of \cref{sec:optimal_aggregate_rate_ica}}\label{app_sec_proof_optimality}

\subsection{Verification of \cref{assu:tilde_A_optimal}: the stochastic expansion of  client-level local estimators}\label{sec:aggre_proof_local_exp}

The following lemma verifies \cref{assu:tilde_A_optimal}. 

\begin{lemma}
\label{lem:client_expansion}
Grant
\cref{assu:sample_order_N,assu:complete_clients,ass_X} as well as condition \eqref{cond_Nk}. Assume the local datasets are independent across clients.  For each $k\in [K]$, let 
$\widetilde{\ba}_1^{(k)},\ldots, \widetilde{\ba}_r^{(k)}$ be the stationary points  of running the projected gradient descent algorithm in \citet[Eqn.(2.5)]{bing2025optimal} $r$ times using $\bY^{(k)}$, with initializations as specified in their Section 5.2. Then for each $i\in [r]$, one has 
we have \begin{align}\label{eq:expan_a_u_v}
    \widetilde{\ba}_{i}^{(k)}-\xi^\star_{(k,i)}\mb a_{\sigma^\star(k,i)}^\star=\mb u_i^{(k)}   + \bv_i^{(k)}.
\end{align}
where  
\begin{align}
\label{def:U_ik}
    \bu_i^{(k)}
    :=
    \frac{1}{3\kappa N_k} 
    \sum_{t=1}^{N_k}
    x_{it}^{(k)3} \sum_{j\in [r]\setminus \{i\}} \ba_j^\star x_{jt}^{(k)}.
\end{align}
and 
the higher-order term $\bv_i^{(k)}$ satisfies 
\begin{equation}\label{bd_v_ki}
    \P\left\{\|\bv_i^{(k)}\| \lesssim {r^{3/2}\log(T)\over N_k},~ \forall  (k,i)\in \cI\right\} \ge 1-{1\over T}.
\end{equation}
\end{lemma}

\begin{proof}
Fix $k\in [K]$ and let $N_k \gtrsim K$ as otherwise one can replace $\log(N_k)$ by $\log(K)$ in the bound and $1-N_k^{-1}$ in the probability tail by $1-K^{-1}$.
For notational convenience, define the sequences
\begin{align}
\label{def:rate_a_l_N}
    a_{N_k}:=\sqrt{\frac{r\log N_k}{N_k}},
    \qquad
     \ell_{N_k}:=\sqrt{\frac{r^2\log N_k}{N_k}}.
\end{align} 
Without loss of generality, assume $\xi^\star_{(k,i)} = 1$ and $\sigma^\star(k,i) = i$ for $i\in [r]$. 
By  \citet[Theorems 3 and 8]{bing2025optimal},  we could verify that for each $i\in [r]$, $\widetilde{\ba}_{i}^{(k)}$ is a stationary point  near $\mb a_i^\star$ satisfying 
\begin{align}\label{rate_a_Nk}
    \|\widetilde{\ba}_{i}^{(k)}- \mb a_i^\star\| \leq a_{N_k}
\end{align} 
with probability at least $1-T^{-2}$. 
To show \eqref{eq:expan_a_u_v}, we start by recalling the empirical objective $F_k(\bq)$ of solving one-column of \eqref{eq:ICA_form} and its popular counterpart $h(\bq)$:  for any $\bq\in\Sp^{r-1}$,
\begin{align}
\label{def:local_varimax_objective}
    F_k(\bq)
    :=
    -\frac{1}{12N_k}
    \sum_{t=1}^{N_k}
    (\bq^\top\by_t^{(k)})^4,
    \qquad h(\mb q):=-\frac{\kappa}{4}\norm{\mb A^{\star\top} \bq}{4}^4.
\end{align}
With $\bP_{\bq}^{\perp}
    :=
    \bI-\bq\bq^\top$,  their Riemannian gradients are
\begin{align}
\label{def:grad_F}
    \grad F_k(\bq)
    &=
    -\frac{1}{3N_k}
    \bP_{\bq}^{\perp}
    \sum_{t=1}^{N_k}
    (\bq^\top\by_t^{(k)})^3\by_t^{(k)},
    \\
\label{def:grad_h}
    \grad h(\bq)
    &=
    -\kappa
    \bP_{\bq}^{\perp}
    \bA^\star
    (\bA^{\star\top}\bq)^{\circ3}.
\end{align}
Further note that the fact that $\widetilde{\ba}_{i}^{(k)}$ is the stationary point of $F_k(\cdot)$ implies 
\begin{align}
    \mb A^{\star\top}\left(\grad h(\widetilde{\ba}_{i}^{(k)})+\grad F_k(\widetilde{\ba}_{i}^{(k)})-\grad h(\widetilde{\ba}_{i}^{(k)})\right)=0.\nonumber
\end{align}
By letting %
\begin{equation}\label{def_zeta}
    \bs\zeta := \bs\zeta(i):=\mb A^{\star\top}\widetilde{\ba}_{i}^{(k)} \in \Sp^{r-1}
\end{equation}
one can deduce that $\zeta_i > 0$. By further writing 
\begin{align}\label{def_beta_ik}
\bs\beta_{-i,k}(\bq):= \frac{1}{\kappa}
    \bA_{-i}^{\star\top}
    \left\{
        \grad h(\bq)
        -
        \grad F_k(\bq)
    \right\},
\end{align}
invoke \citet[equation B.22]{bing2025optimal} to have \begin{align}\label{eq:statioanry_cubic_zeta}
    \bs\zeta_{-i}^{\circ 3} -\norm{\bs \zeta}{4}^4\bs\zeta_{-i} +\bs\beta_{-i,k}(\widetilde{\ba}_{i}^{(k)})=0.
\end{align}
where $\bv^{\circ 3} = \{v_i^3\}_{i\in [d]}$ for $\bv\in \RR^d$. %
By the fact that $\bA^\star \in \bbO_{r,r}$, we also have
\begin{align*}
\widetilde{\ba}_{i}^{(k)}-\mb a_i^\star&=\mb A^\star\bs\zeta-\mb a_i^\star\nonumber\\
&=\mb A^\star_{-i}\bs\zeta_{-i}+(\zeta_i-1)\mb a_i^\star\nonumber\\
&={1\over \|\bs\zeta\|_4^4}\mb A^\star_{-i} \left(\bs\zeta_{-i}^{\circ 3}+\bs\beta_{-i,k}(\widetilde{\ba}_{i}^{(k)})\right)+(\zeta_i-1)\mb a_i^\star\nonumber\\
&={1\over \|\bs\zeta\|_4^4}  \mb A^\star_{-i}\bs\zeta_{-i}^{\circ 3}+{1\over \|\bs\zeta\|_4^4}\mb A^\star_{-i}\bs\beta_{-i,k}(\mb a_i^\star)+{1\over \|\bs\zeta\|_4^4}\mb A^\star_{-i}\left[\bs\beta_{-i,k}(\widetilde{\ba}_{i}^{(k)})-\bs\beta_{-i,k}(\ba_i^\star)\right]+(\zeta_i-1)\mb a_i^\star.
\end{align*}
By the definition of $\mb u_i^{(k)}$ in \eqref{def:U_ik},  we observe $\mb A^\star_{-i}\bs\beta_{-i,k}(\mb a_i^\star)=\mb u_i^{(k)}$ so that 
\begin{align*}
    \bv_i^{(k)} &:= \widetilde{\ba}_{i}^{(k)}-\mb a_i^\star-\mb u_i^{(k)}\nonumber\\
    &=\left({1\over \|\bs\zeta\|_4^4}-1\right)\mb u_i^{(k)}+{1\over \|\bs\zeta\|_4^4}\mb A^\star_{-i}\bs\zeta_{-i}^{\circ 3}+{1\over \|\bs\zeta\|_4^4}\mb A^\star_{-i}\left[\bs\beta_{-i,k}(\widetilde{\ba}_{i}^{(k)})-\bs\beta_{-i,k}(\ba_i^\star)\right]+(\zeta_i-1)\mb a_i^\star.
\end{align*}
It remains to prove \eqref{bd_v_ki} by bounding from above  
\begin{align*} 
\|\bv_i^{(k)}\| &\leq {1 \over \|\bs\zeta\|_4^4}\left\{\left(1-\norm{\bs\zeta}{4}^4\right)  \|\mb u_i^{(k)}\| +  \norm{\bs \zeta_{-i}^{\circ 3}}{}+  \bigl\|\bs\beta_{-i,k}(\widetilde{\ba}_{i}^{(k)})-\bs\beta_{-i,k}(\ba_i^\star)\bigr\| \right\} +\abs{\zeta_i-1}\\
&\lesssim \left(1-\norm{\bs\zeta}{4}^4\right)  \|\mb u_i^{(k)}\| +  \|\bs\zeta_{-i}\|^3 +  \bigl\|\bs\beta_{-i,k}(\widetilde{\ba}_{i}^{(k)})-\bs\beta_{-i,k}(\ba_i^\star)\bigr\| +1 -  \zeta_i \\
&\le (1-\zeta_i) \|\mb u_i^{(k)}\| +  (1-\zeta_i)^{3/2} +  \bigl\|\bs\beta_{-i,k}(\widetilde{\ba}_{i}^{(k)})-\bs\beta_{-i,k}(\ba_i^\star)\bigr\| +1 -  \zeta_i 
\end{align*}
where the second step uses $\norm{\bs \zeta}{4}^{4}\geq 1/9$ from  \citet[equation B.26]{bing2025optimal} and the last step uses $1-\|\bs\zeta\|_4^4 \le 1-\zeta_i^4 \lesssim   1-\zeta_i$ as well as $\|\bs\zeta_{-i}\|^2= \|\bs\zeta\|^2 - \zeta_i^2 \le 2(1-\zeta_i)$ from \eqref{def_zeta}. By \eqref{def_zeta} and \eqref{rate_a_Nk}, we further have 
$$
     1-\zeta_i ~ =~  {1\over 2} \|\widetilde{\ba}_{i}^{(k)}-\ba_i^\star\|^2 ~ \overset{}{\le} ~ {1 \over  2}a_{N_k}^2,
$$
implying 
\[
   \|\bv_i^{(k)}\| ~ \lesssim ~ a_{N_k}^2\|\mb u_i^{(k)}\| +\bigl\|\bs\beta_{-i,k}(\widetilde{\ba}_{i}^{(k)})-\bs\beta_{-i,k}(\ba_i^\star)\bigr\|+a_{N_k}^2.
\]
Let  
\begin{align}\label{def:s_k}
            S_k(\mb q)
        :=
        \frac1{N_k}
        \sum_{t=1}^{N_k}
        \left[
        (\mb q^\top \by_t^{(k)})^3\by_t^{(k)}
        -
        \mathbb E\{(\mb q^\top \by_t^{(k)})^3\by_t^{(k)}\}
        \right]
\end{align}
so that 
\begin{equation}\label{equiv_uki}
     \bu_i^{(k)} = {1\over  3\kappa}\bP_{\ba_i^\star}^\perp S_k(\ba_i^\star)
\end{equation}
and 
\[
     \bs\beta_{-i,k}(\bq)-\bs\beta_{-i,k}(\ba_i^\star) = {1\over 3\kappa}\bA^{\star^\T}_{-i}  \left(  \mb P_{\mb q}^\perp S_k(\mb q) -  \mb P_{\ba_i^\star}^\perp S_k(\ba_i^\star)\right).
\]
The rate in \eqref{bd_v_ki} follows by invoking \cref{lem:oracle_gradient_increment}. 

This proves the claim for each fixed client $k\in [K]$. The claim follows by the union bound argument. 
\end{proof}

The following lemma provides controls on quantities related with $S_k$.
\begin{lemma}
\label{lem:oracle_gradient_increment}
Pick any \(k\in [K]\) and \(i\in[r]\). Under $\bY^{(k)} = \bA^\star \bX^{(k)}$ with $\bA^\star \in \bbO_{r,r}$, \cref{ass_X} and $N_k \ge r^2\log(N_k)$,  with probability at least $1-N_k^{-2}$, the following holds uniformly over
\(\mb q\in\mathbb S^{r-1}\),
\[
        \norm{S_k(\ba_i^\star)}{}
        \le
        C a_{N_k},
\]
\[
        \norm{S_k(\mb q)-S_k(\ba_i^\star)}{}
        \le
        C\norm{\mb q-\ba_i^\star}{}\ell_{N_k},
\]
and
\[
        \norm{ \mb P_{\mb q}^\perp S_k(\mb q)- \mb P_{\ba_i^\star}^\perp S_k(\ba_i^\star)  }{}
        \le
        C\norm{\mb q-\ba_i^\star}{}\ell_{N_k}.
\]
Here $C>0$ is some absolute constant, depending only on $\gamma$, $\kappa$ and $\sigma$.  
\end{lemma}

\begin{proof}
The first claim follows from \citet[lemma $22$]{bing2025optimal} with
$\mb q=\mb q_0=\ba_i^\star$.

 For the second claim, write
$
        \bs \delta=\mb q-\ba_i^\star.
$
so that 
\[
        (\mb q^\top \by_t^{(k)})^3
        -
        (\ba_i^{\star\top} \by_t^{(k)})^3
        =
        (\bs \delta^\top \by_t^{(k)})^3
        +
        3(\bs \delta^\top \by_t^{(k)})^2(\ba_i^{\star\top} \by_t^{(k)})
        +
        3(\bs \delta^\top \by_t^{(k)})(\ba_i^{\star\top} \by_t^{(k)})^2 .
\]
Invoke the last three terms $\rII, \rIII$, and $\rIV$ in the proof of \citet[Lemma 22]{bing2025optimal} to obtain
\[
        \norm{S_k(\mb q)-S_k(\ba_i^\star)}{}
        \le
        C\norm{\bs \delta}{}\ell_{N_k}
        =
        C\norm{\mb q-\ba_i^\star}{}\ell_{N_k} .
\]
\noindent
For the third claim, write for short 
\[
        G_k(\mb q) = - \mb P_{\mb q}^\perp S_k(\mb q).
\] 
We have
\[
\begin{aligned}
        G_k(\mb q)-G_k(\ba_i^\star)
        &=
        -  \left\{
        \mb P_{\mb q}^\perp S_k(\mb q)
        - \mb P_{\ba_i^\star}^\perp S_k(\ba_i^\star)
        \right\}                                                     \\
        &=
        -  \mb P_{\mb q}^\perp\{S_k(\mb q)-S_k(\ba_i^\star)\} - 
        (\mb P_{\mb q}^\perp-\mb P_{\ba_i^\star}^\perp)S_k(\ba_i^\star).
\end{aligned}
\]
Therefore
\[
\begin{aligned}
        \norm{G_k(\mb q)-G_k(\ba_i^\star)}{}
        &\le
        C\norm{S_k(\mb q)-S_k(\ba_i^\star)}{}
        +
        C\left\|\mb P_{\mb q}^\perp-\mb P_{\ba_i^\star}^\perp\right\|_{\op}
        \norm{S_k(\ba_i^\star)}{}                                           \\
        &\le
        C\norm{\mb q-\ba_i^\star}{}\ell_{N_k}
        +
        C\norm{\mb q-\ba_i^\star}{}a_{N_k}.
\end{aligned}
\]
Since \(\ell_{N_k}\ge a_{N_k}\), we obtain
\[
        \norm{G_k(\mb q)-G_k(\ba_i^\star)}{}
        \le
        C\norm{\mb q-\ba_i^\star}{}\ell_{N_k}.
\]
This proves the lemma.
\end{proof}

\subsection{Proof of \cref{thm:optimal_aggre_rate}}\label{sec:proof_thm:optimal_aggre_rate}

\begin{proof}
    Based on the stochastic expansion of local estimators in \cref{assu:tilde_A_optimal}, the main ingredient of the proof is to apply  the new concentration inequalities of the geometric median derived in \cref{sec:aggre_proof_aggregation}. For simplicity, we assume $N\ge K$, so that $T=N$. Otherwise, the same result holds with $\log N$ replaced by $\log K$, and the resulting probability becomes $1-K^{-1}$.
    
    Recall the definition of $\varsigma_a$ in \cref{prop:sign_alignment}. It suffices to show that with probability tending to one, 
    \begin{align}\label{ieq:bound_rate_A_column}
        \bigl\|\wh \ba_a - \varsigma_a\ba_{\pi(a)}^\star\bigr\|\lesssim \sqrt{\frac{r\log(N)}{KN}}.
    \end{align}
    for any choice of $a\in [r]$. Without loss of generality we assume $\varsigma_a=1$ as well as $\pi(a) = a$. 
    
    Recall from \cref{alg:1} that 
    \[
          \wh \ba_a-\ba_a^\star =
    \argmin_{\mb q\in\RR^r}
    \sum_{(k,i)\in\wt{\mathcal C}_a}
    \left\|\mb q-\bigl(\widecheck{\mb a}^{(k)}_i-\ba_a^\star\bigr)\right\|.
    \]
    First invoking \cref{lem:oracle_gradient_increment} and \eqref{equiv_uki} together with a union bound gives
    \[
        \max_{(k,i)\in \cI}
        \epsilon_i^{(k)}
        \lesssim
        \sqrt{\frac{r\log N}{N}}.
    \]
    with probability $1-N^{-1}$. Consequently, for sufficiently large $N$,
    condition \eqref{cond_eps_ik} is met and \cref{prop:sign_alignment} ensures that all local atoms are correctly clustered, \begin{equation}\label{res_clusters}
        \widetilde{\cC}_a = \cC_{a}^\star ,\quad \forall ~ a\in [r],
    \end{equation}
    and sign-aligned
    \begin{equation}\label{res_sign}
        \widecheck\epsilon_i^{(k)} = \epsilon_i^{(k)},\quad \forall ~ (k,i)\in \cI.
    \end{equation} 
    As a result of \eqref{res_clusters}, 
    \[
         \wh \ba_a-\ba_a^\star = \GMed{\{\widecheck{\mb a}^{(k)}_i-\ba_a^\star\}_{(k,i)\in \cC_a^\star}}
    \]
    while \cref{assu:tilde_A_optimal} together with 
    \eqref{res_sign} ensures   
    \begin{align}\label{decomp_wc_a}
        \widecheck{\mb a}^{(k)}_i-\ba_a^\star=\bu_i^{(k)} +
        \mb v_i^{(k)}.
    \end{align}
    We then aim to invoke \cref{lem:spatial_median_rate} to bound $\|\text{GM}(\{\widecheck{\mb a}^{(k)}_i-\ba_a^\star\}_{(k,i)\in \cC_a^\star})\|$ by verifying its conditions. 
    
    For any vector $\bx\in \R^r$, define its projection onto the unit sphere as 
    \begin{equation}\label{def_proj_sphere}
        \Pi_{\Sp}(\mb{x})=  \mb{x}/\|\mb{x}\|,\quad \text{for }\mb{x}\neq0
    \end{equation}
    and $\Pi_{\Sp}(\bx) = 0$ otherwise.   By using \cref{lem:spatial_sign_bias} under \cref{ass_X},  condition 
    \eqref{eq:approx_spatial_centering} holds with 
    \begin{align}
        \frac{1}{K}\left\|\sum_{k=1}^K \E\left[\Pi_{\Sp}(\mb u_i^{(k)})\right]\right\|\lesssim \sqrt{\frac{r}{N}} =: \nu.
    \end{align}
    By further invoking \cref{lem:normalized_leading_second_moment} and  the sub-Gaussianity in \cref{ass_X},  condition \eqref{ieq:condition_fourth_z} holds with $C_3 = (r-2)/(r-1)$ and $C_4 = C_4(\gamma)$. Finally, it is easy to see that $\E[\|\bu_i^{(k)}\|^2] \asymp r/N$. We thus invoke \cref{lem:spatial_median_rate} with $b=\sqrt{r/N}$, $\rho=r^{3/2}\log(N)/N$ and $\nu=\sqrt{r/N}$ to obtain: 
    \begin{align}
        \|\wh{\mb a}^{(k)}_i-\ba_a^\star \| ~ &\lesssim  
        \sqrt{\frac{r\log(K)}{NK}}  +\sqrt{\frac{r^2\log(N)}{N^{3/2}}} + {r^{3/2}\log N\over N}\nonumber\\
        &\asymp \sqrt{\frac{r\log(K)}{NK}}+\sqrt{\frac{r^2\log(N)}{N^{3/2}}}.
    \end{align}
    The last step also uses $ r^2\log^2(N) \le N$ to collect terms. This completes the proof. 
\end{proof}

\subsection{Technical lemmas used in the proof of \cref{thm:optimal_aggre_rate}}

Recall the definition of $\bu_i^{(k)}$ in \cref{lem:client_expansion} and $\Pi_{\Sp}(\cdot)$ in the proof of \cref{lem:spatial_median_rate}, we have the following lemma:

\begin{lemma}
\label{lem:spatial_sign_bias}
Under \cref{assu:complete_clients,ass_X} and independence of $\bX^{(k)}$ across $k\in [K]$, one has  
\[
    \max_{i\in[r]} \left\| \frac1K\sum_{k=1}^K \E[\Pi_{\Sp}(\bu_i^{(k)})]
    \right\|  ~ \le ~  \frac{C}{K}
    \sum_{k=1}^K\sqrt{r\over N_k},
\]
where $C>0$ is some constant depending only on $\gamma$. 
\end{lemma}

\begin{proof} 
Pick any \(i\in[r]\) and \(k\in[K]\), denote 
\begin{align}\label{def:T_x}
        \mb t_{i,k}
    :=
    \frac1{\sqrt{N_k}}
    \sum_{t=1}^{N_k}
    x_{it}^{(k)3}\mb x^{(k)}_{-i,t} \in \R^{r-1}.
\end{align}
Recall the definition of $\bu_i^{(k)} = \frac{1}{3\kappa\sqrt{N_k}}\bA_{-i}^\star \mb t_{i,k}$, since
\(\bA_{-i}^{\star\top}\bA_{-i}^\star=\mb I_{r-1}\)
 we have
$
    \Pi_{\Sp}(\bu_i^{(k)})
    =
    \bA_{-i}^\star \Pi_{\Sp}(\mb t_{i,k})
$
so that 
\[
    \|\EE[\Pi_{\Sp}(\bu_i^{(k)})]\| = \|\EE[\Pi_{\Sp}(\mb t_{i,k})]\|.
\]
To bound from above the right hand side, let $\bs\theta$ be uniformly distributed on $\Sp^{r-2}$.  For any $\mb x\in\mathbb R^{r-1}$, we obtain
\begin{align}
\E_{\bs \theta}
\left[
    \bs \theta\,\sgn(\bs \theta^\top\mb x)
\right]
&=
2\E_{\bs \theta}
\left[
    \bs \theta\,
    \mathbf 1\{\bx^\T \bs \theta>0\}
\right]
\nonumber\\
&=
2\,\Pi_{\Sp}(\mb x)\,
\E_{\bs \theta}
\left[
    \Pi_{\Sp}(\mb x)^\top\bs \theta\,
    \mathbf 1\{\Pi_{\Sp}(\mb x)^\top\bs\theta>0\}
\right]\nonumber\\
&=
\Pi_{\Sp}(\mb x)\,
\E_{\bs \theta}
\left[
    |\Pi_{\Sp}(\mb x)^\top\bs \theta|
\right].
\label{eq:spatial-sign-sphere}
\end{align}
where the second equality uses the fact that $\Pi_{\Sp}(\mb x)\Pi_{\Sp}(\mb x)^\T$ is a projection and $\btheta$ is uniformly distributed. This implies  
\[
 \Pi_{\Sp}(\mb t_{i,k})  = 
\frac{
    \E_{\bs\theta}
    \left[
        \bs\theta\,
        \sgn(\bs\theta^\top\mb t_{i,k})
    \right]
}{
    \E_{\bs\theta}
    \left[
        |\Pi_{\Sp}(\mb t_{i,k})^\top\bs\theta|
    \right]
} 
\]
and by rotational invariance, the denominator equals to $\E_{\bs\theta}|\theta_1|\approx r^{-1/2}$.  We thus obtain 
\begin{align*}
\left\|\E \Pi_{\Sp}(\mb t_{i,k})\right\|_2 &\lesssim \sqrt{r} 
\left\|\E_{\bs\theta}\left[
    \bs\theta\,
    \E[\sgn(\bs\theta^\top\mb t_{i,k})]
\right]
\right\|_2 \le \sqrt{r} ~ 
\E_{\bs\theta}
\left|
    \E[\sgn(\bs\theta^\top\mb t_{i,k})]
\right|
\end{align*}
where the last step uses Cauchy-Schwarz inequality. Finally, by noting that 
\[
    \bs\theta^\top\mb t_{i,k}
    =
    \frac1{\sqrt{N_k}}
    \sum_{t=1}^{N_k}
    x_{it}^{(k)3}\bs\theta ^\top\mb x^{(k)}_{-i,t},
\]
\cref{lem:one_d_BE} invokes the one-dimensional Berry-Esseen bound to give
\begin{align}\label{ieq:bound_sign_qt}
        \left|
        \E[\sgn(\bs\theta^\top\mb t_{i,k})]
    \right|
    \lesssim
    \frac{1}{\sqrt{N_k}},
\end{align}
which further implies
\[ \max_{i\in[r]} \left\| \frac1K\sum_{k=1}^K \E[\Pi_{\Sp}(\bu_i^{(k)})]
    \right\| 
     \le
    \max_{i\in[r]}
    \frac1K
    \sum_{k=1}^K
    \left\|\E[\Pi_{\Sp}(\bu_i^{(k)})]\right\|  \lesssim
    \frac{1}{K}
    \sum_{k=1}^K \sqrt{r\over N_k}.\]
The proof is complete.
\end{proof}

\begin{lemma}\label{lem:one_d_BE}
    Grant \cref{ass_X}. For any $\mb q\in \mathbb S^{r-2}$, with $\mb t_{i,k}$ given in \eqref{def:T_x}, 
    we have \begin{align}
        \left|
        \E\left[{\rm sgn} \left(\mb q^\top\mb t_{i,k}\right)\right]
    \right|\lesssim \frac{1}{\sqrt{N_k}}.
    \end{align}
\end{lemma}
\begin{proof}
    For any fixed $\mb q\in\mathbb S^{r-2}$, define
\[
    \xi_t(\mb q)
    :=
    x_{it}^{(k)3}\,\mb q^\top\mb x_{-i,t}^{(k)},
\]
By the coordinate-wise independence in $\bX^{(k)}$ and $\|\mb q\|_2=1$,
\[
    \Var\left(\xi_t(\mb q)\right)
    =
    \E x_{it}^{(k)6}\,
    \Var(\mb q^\top\mb x_{-i,t}^{(k)})
    =
    \sigma^2\E x_{it}^{(k)6}.
\]
Moreover, there exists some $C = C(\gamma)$ such that 
\[
    \frac{\E|\xi_t(\mb q)|^3}{\Var^{3/2}\!\left(\xi_t(\mb q)\right)}
    \le C.
\]
Hence, for
\[
    Z_{i,k}(\mb q)
    :=
    \frac{\mb q^\top\mb t_{i,k}}{\sqrt{\Var\left(\xi_t(\mb q)\right)}}
    =
    \frac{1}{\sqrt{N_k}}
    \sum_{t=1}^{N_k}
    \frac{\xi_t(\mb q)}{\sqrt{\Var\left(\xi_t(\mb q)\right)}},
\]
which has mean zero,   
we invoke the one-dimensional Berry--Esseen bound to obtain
\[
    \sup_{z\in\mathbb R}
    \left|
        \P(Z_{i,k}(\mb q)\le z)-\Phi(z)
    \right|
    \le
    \frac{C_\gamma}{\sqrt{N_k}}.
\]
Let $F_q$ denote the distribution function of $Z_{i,k}(\mb q)$. Since
$\Phi(0)=1/2$,
\[
    \left|F_q(0)-\frac12\right|
    \le
    \frac{C_\gamma}{\sqrt{N_k}},
    \qquad
    \left|F_q(0-)-\frac12\right|
    \le
    \frac{C_\gamma}{\sqrt{N_k}}.
\]
Therefore,
\begin{align*}
     \left|
        \E\sgn(\mb q^\top\mb t_{i,k})
    \right| &= \left|
        \P(Z_{i,k}(\mb q)>0)
        -
        \P(Z_{i,k}(\mb q)<0)
    \right| =
    \left|
        1-F_q(0)-F_q(0-)
    \right|
    \le
    \frac{2C_\gamma}{\sqrt{N_k}},
\end{align*}
completing the proof.
\end{proof}

\begin{lemma}\label{lem:normalized_leading_second_moment}
For any $i\in[r]$, let $\mb z_k$ be given in \cref{lem:spatial_median_rate} using $\bu_i^{(k)}$.
Suppose that the coordinates of \(\mb x_t^{(k)}\) are independent,
mean zero, and have unit variance. Then
\[
    \E[\mb z_k\mb z_k^\top]
    =
    \frac{1}{r-1}
    \bA_{-i}^{\star}\bA_{-i}^{\star\top}.
\]
Consequently,
\[
    \norm{
    \E[\mb z_k\mb z_k^\top]
    }{\op}
    =
    \frac{1}{r-1}.
\]
\end{lemma}

\begin{proof}
Pick any $k\in [K]$ and recall the definition of $\mb z_k$ and $\mb u_k$ in \eqref{def:U_ik}:
\begin{align}
    \mb z_k:=\frac{\mb u_k}{\sqrt{\E\norm{\mb u_k}{}^2}}=\frac{1}{3\kappa N_k \sqrt{\E\norm{\mb u_k}{}^2}}\sum_{t=1}^{N_k}x_{it}^{(k)3}\mb A_{-i}^\star\mb x_{-i,t}^{(k)}.
\end{align}
Thus 
\begin{align}
\E[\mb z_k\mb z_k^\top]
&=
\frac{1}{
    9\kappa^2 N_k^2\,\E\|\mb u_k\|_2^2
}
\sum_{t=1}^{N_k}\sum_{s=1}^{N_k}
\E\left[
    (x_{it}^{(k)})^3(x_{is}^{(k)})^3
    \mb A_{-i}^\star
    \mb x_{-i,t}^{(k)}
    (\mb x_{-i,s}^{(k)})^\top
    \mb A_{-i}^{\star\top}
\right]
\nonumber\\
&=
\frac{1}{
    9\kappa^2 N_k^2\,\E\|\mb u_k\|_2^2
}
\sum_{t=1}^{N_k}
\E\left[(x_{it}^{(k)})^6\right]
\mb A_{-i}^\star
\E\left[
    \mb x_{-i,t}^{(k)}
    (\mb x_{-i,t}^{(k)})^\top
\right]
\mb A_{-i}^{\star\top}
\nonumber\\
&=
\frac{
    \E[(x_{it}^{(k)})^6]
}{
    9\kappa^2 N_k\,\E\|\mb u_k\|_2^2
}
\mb A_{-i}^\star
\mb A_{-i}^{\star\top}.
\end{align}
On the other hand 
\begin{align*}
\E\|\mb u_k\|_2^2  =
\tr\!\left(\E[\mb u_k\mb u_k^\top]\right) =
\frac{\E[(x_{it}^{(k)})^6]}{9\kappa^2 N_k}\tr\!\left(\mb A_{-i}^\star
    \mb A_{-i}^{\star\top}\right) =
\frac{ (r-1)\E[(x_{it}^{(k)})^6]}{9\kappa^2 N_k}.
\end{align*}
Plugging $\E\|\mb u_k\|_2^2$ back into $\E[\mb z_k\mb z_k^\T]$ completes the proof.
\end{proof}

\section{Concentration inequalities for the geometric median of independent, approximately spatially centered random vectors under perturbations}\label{sec:aggre_proof_aggregation}
The concentration and local-curvature properties of the geometric median
have been extensively studied (see, e.g.,
\citet{minsker2015geometric,minsker2024geometric}) and the classical spatial-median
literature \citep{brown1983statistical,mottonen2010asymptotic}.
The following lemma adapts these ideas to our setting, allowing the random vectors to be independent but non-identically distributed, only approximately
spatially centered, and subject to deterministic perturbations. Recall $\Pi_{\Sp}$ from \eqref{def_proj_sphere}.

\begin{lemma}[Concentration of the geometric median under perturbations]
\label{lem:spatial_median_rate}
Let $\mb{w}_k=\mb{u}_k+\mb{r}_k$ for $k\in [K]$ be random vectors in $\R^r$, where  the vectors $\mb{u}_1,\ldots,\mb{u}_K$ are independent, and satisfy 
    \begin{align}\label{eq:approx_spatial_centering}
            \left\|\frac1K\sum_{k=1}^K \E [\Pi_{\Sp}(\mb{u}_k)]\right\|\le \nu,
    \end{align}
    for some sufficiently small constant $\nu>0$.  
    By letting $b_k:= (\E\|\mb{u}_k\|^2)^{1/2}$  and  $\mb{z}_k:=\mb{u}_k/b_k$,  assume
    \begin{align}\label{ieq:condition_fourth_z}
    C_1 b \le b_k \le C_2 b,\qquad \bigl\|\E\mb{z}_k\mb{z}_k^\top\bigr\|_{\op}\le 1-C_3,
        \qquad
        \E\|\mb{z}_k\|^4\le C_4,\qquad \forall ~  k\in [K],
    \end{align} 
    for  some constants $C_1, C_2, C_4>0$, $C_3 \in (0,1)$ and some quantity $b>0$.
Further assume $\max_{k\in[K]}\|\mb{r}_k\|   \le   \rho$.
Then for some sufficiently large $K$,  one has
\[
    \P\left\{\left\|\GMed{\mb{w}_1,\ldots,\mb{w}_K}\right\|
      \lesssim
     ~ b\sqrt{\frac{\log K}{K}}+b\nu+\sqrt{b\rho} + \rho\right\} \ge 1-K^{-2}-e^{-c'K}.
\]
\end{lemma}

\begin{proof}
For any $\bv\in \RR^r$, define
\[
    \Phi_u(\bv)=\frac1K\sum_{k=1}^K\|\bv-\mb{u}_k\|,
    \qquad
    \Phi_w(\bv)=\frac1K\sum_{k=1}^K\|\bv-\mb{w}_k\|.
\]
By definition, we have 
\[
    \GMed{\mb{w}_1,\ldots,\mb{w}_K} = \argmin_{\bv \in \RR^r} ~ \Phi_w(\bv).
\]
For some large constant $C>0$, let 
\begin{equation}\label{def_alpha}
    \alpha := C \left(   b\sqrt{\frac{\log K}{K}}+b\nu + \rho +\sqrt{b\rho}\right)
\end{equation}
 Let $\Ba_r(\alpha):= \{\bv \in \RR^r: \|\bv\| \le \alpha\}$ and $\Ba^c_r(\alpha) = \{\bv \in \RR^r: \|\bv\| > \alpha\}$. We aim to show that
\begin{equation}\label{goal}
  \Phi_w(\mb{v}) > \Phi_w(0),\quad \text{ for any $\bv \in \Ba_r^c(\alpha)$.}
\end{equation}
\begingroup\emergencystretch=3em
If this holds, then since $\Phi_w(\cdot)$ is convex on $\RR^r$, we must have $\GMed{\mb{w}_1,\ldots,\mb{w}_K}\in \Ba_r(\alpha)$ so that $\|\GMed{\mb{w}_1,\ldots,\mb{w}_K}\|\le \alpha$. The claim then follows.  \par\endgroup

To prove \eqref{goal}, by noting that 
\begin{align*}
    \sup_{\mb{v}\in \RR^r}
    |\Phi_w(\mb{v})-\Phi_u(\mb{v})|
    &\le \frac1K\sum_{k=1}^K\|\mb{r}_k\|
    \le \rho,
\end{align*}
we have for all $\bv\in \RR^r$
\begin{align}
    \Phi_w(\mb v)-\Phi_w(0)
    &\ge \Phi_u(\mb{v})-\Phi_u(0)-2\rho. 
\label{eq:gm-oracle-global-growth}
\end{align}
It remains to bound from below $ \Phi_u(\mb{v})-\Phi_u(0)$ for any  $\bv \in \Ba_r^c(\alpha)$. Writing 
\[
    \mb{S}_K:=\frac1K\sum_{k=1}^K \Pi_{\Sp}(\mb{u}_k),
\]
and using \cref{lem:euclidean-norm-quadratic-lower-bound} yield
\begin{align*}
\Phi_u(\mb v)-\Phi_u(0)
&=\frac1K\sum_{k=1}^K\left(\|\mb v-\mb u_k\|-\|\mb u_k\|\right) \\
&\ge \frac1K\sum_{k=1}^K
\frac{\mb v^\top(\mb I_r-\Pi_{\Sp}(\mb u_k)\Pi_{\Sp}(\mb u_k)^\top)\mb v}
{2(\|\mb v\|+\|\mb u_k\|)}
-\mb S_K^\top\mb v \\
&\ge \frac1K\sum_{k=1}^K
    \frac{
        \mb v^\top
        \left(
            \mb I_r-\Pi_{\Sp}(\mb u_k)\Pi_{\Sp}(\mb u_k)^\top
        \right)
        \mb v}{2(\|\mb v\|+
             B_k)}1 \left\{ \|\mb u_k\| \le B_k \right\}
    -
    \|\mb S_K\|\|\mb v\|
\end{align*}
for some $B_k > 0$ to be specified later. To bound from below the first term on the right-hand-side, we need to bound from below the smallest eigenvalue of $\EE[\mb{\Xi}_k]$
where 
\[
    \mb{\Xi}_k:=
    \left(\mb{I}_r-\Pi_{\Sp}(\mb{u}_k)\Pi_{\Sp}(\mb{u}_k)^\top\right)
    \mathbf 1\left\{\|\mb{u}_k\|\le
   B_k\right\}.
\] 
To this end,  for any unit vector $\bv \in \Sp^{r-1}$, we have 
\begin{align}
 \mb{v}^\top \EE[\mb{\Xi}_k] \mb{v}
&=\E\left[
    \left(1-\frac{(\mb{v}^\top\mb{z}_k)^2}{\|\mb{z}_k\|^2}\right)
    \mathbf 1\left\{\|\mb{z}_k\|\le {B_k\over b_k}\right\}
\right]
\nonumber\\
&\ge \frac{b_k^2}{B_k^2}
\E\left[
    \left(\|\mb{z}_k\|^2-(\mb{v}^\top\mb{z}_k)^2\right)
    \mathbf 1\left\{\|\mb{z}_k\|\le{B_k\over b_k}\right\}
\right]
\nonumber\\
&\ge \frac{b_k^2}{B_k^2}
\left\{1-\mb{v}^\top\E(\mb{z}_k\mb{z}_k^\top)\mb{v}
-\E\left[\|\mb{z}_k\|^2
\mathbf 1\left\{\|\mb{z}_k\|> {B_k\over b_k}\right\}\right]\right\} &&\text{by $\E\|\mb{z}_k\|^2=1$}
\nonumber\\
&\ge \frac{b_k^2}{B_k^2}\left(C_3-{b_k^2 C_4 \over B_k^2}\right)  &&\text{$(*)$}
\nonumber\\
&  =\frac{C_3^2}{4C_4} &&\text{by } B_k =\sqrt{\frac{2C_4}{C_3}}\,b_k .
\label{eq:gm-curvature-pop}
\end{align}
Here the step $(*)$ uses the fourth moment condition  and
\[
\E\left[\|\mb{z}_k\|^2
\mathbf 1\left\{\|\mb{z}_k\|> B_k/b_k\right\}\right]
\le \frac{\E\|\mb{z}_k\|^4}{B_k^2/b_k^2}\le {b_k^2 C_4 \over B_k^2}.
\] 
Further invoking \cref{lem:dimension-free-curvature} gives that with probability at least $1-2e^{-cK}$ for some $c = c(C_3,C_4)>0$, 
\begin{align}
    \lambda_{\min}\left(\frac1K\sum_{k=1}^K\mb{\Xi}_k\right)
    \gtrsim \frac{C_3^2}{C_4}.
\label{eq:gm-curvature-emp}
\end{align} 
Consequently, we have 
\[
    \frac1K\sum_{k=1}^K
    \frac{
        \mb v^\top  \mb{\Xi}_k
        \mb v}{2(\|\mb v\|+
             B_k)}  \gtrsim \frac{C_3^2}{C_4} {\|\bv\|^2\over  \|\bv\| + b\sqrt{C_4/C_3}},
\]
where the last step uses  $b_k\asymp b$. 
  
On the one hand side, we bound from above 
\[
    \|\mb{S}_K\| \le \left\|\mb{S}_K-\E\mb{S}_K\right\|
        +\left\|\E\mb{S}_K\right\|\le \left\|\mb{S}_K-\E\mb{S}_K\right\|
        + \nu.
\]
By independence over $k\in [K]$, note that 
\begin{align}
    \E\left\|\mb{S}_K-\E\mb{S}_K\right\|^2
    &=\frac1{K^2}\sum_{k=1}^K
    \E\left\|\Pi_{\Sp}(\mb{u}_k)-\E \Pi_{\Sp}(\mb{u}_k)\right\|^2
    \le \frac1K.
\end{align}
Since changing one summand indexed by $k\in[K]$ changes
$\|\mb S_K-\E\mb S_K\|_2$ by at most $2/K$, the bounded-difference
inequality \citep{mcdiarmid1989bounded} yields that 
\begin{align}
\P\left\{\|\mb{S}_K\| 
 ~ \lesssim \sqrt{\frac{\log K}{K}}+\nu\right\} \ge 1-K^{-2}.
\label{eq:spatial-sign-concentration-new}
\end{align}

Combining 
\eqref{eq:gm-curvature-emp} and \eqref{eq:spatial-sign-concentration-new} yields that, for any $\bv \in \Ba^c_r(\alpha)$,
\begin{align}
    \Phi_u(\mb v)-\Phi_u(0)  
    & ~ \ge  ~ c(C_3,C_4)  {\|\bv\|^2\over b+\|\bv\|} 
    - C\left( \nu+ \sqrt{\frac{\log K}{K}} \right)
    \|\mb v\| .
\label{eq:gm-oracle-growth-U}
\end{align}  
The above argument is a localized quadratic-growth argument for the
geometric-median objective, analogous to the one developed in
\citet[Theorem~2.3]{minsker2024geometric}. Here we extend it under
our moment and angular non-degeneracy conditions.

In view of  \eqref{eq:gm-oracle-global-growth}, it suffices to show that for any $\|\bv\|_2 > \alpha$,
\begin{align}
    \left\{c(C_3,C_4)   - C\left( \nu+ \sqrt{\frac{\log K}{K}} \right)\right\}\|\bv\|^2 
    - \left\{C\left( \nu + \sqrt{\frac{\log K}{K}} \right)b + 2\rho\right\}
    \|\mb v\|   -2\rho b > 0.
\label{eq:gm-oracle-quadratic}
\end{align}
Since both $\nu$ and $\log K/K$ are sufficiently small, it is easy to see that the above is met as long as we choose
\[
    \alpha \ge C'\left\{
    b\left(\nu+\sqrt{\frac{\log K}{K}}\right)
    +\rho+\sqrt{b\rho}
    \right\},
\]
for some large $C'>0$ depending on $C_3$ and $C_4$.  Choosing $\alpha$ as the above order completes the proof. 
\end{proof}

The following technical lemmas are used in the proof of \cref{lem:spatial_median_rate}.

\begin{lemma}\label{lem:euclidean-norm-quadratic-lower-bound}
Let \(\mb x,\mb u\in\mathbb R^d\) with \(\bx\neq 0\), and recall the definition of $\Pi_{\mathbb S}$ in \cref{def_proj_sphere}.
Then
\begin{align}
    \norm{\mb u-\mb x}{}
    -
    \norm{\mb x}{}
    +
    \innerprod{\Pi_{\mathbb S}(\mb x)}{\mb u}
    \ge
    \frac{
    \mb u^\top
    \left(
    \mb I_d-\Pi_{\mathbb S}(\mb x)\Pi_{\mathbb S}(\mb x)^\top
    \right)
    \mb u
    }{
    2\left(
    \norm{\mb x}{}+\norm{\mb u}{}
    \right)
    }.
\end{align}
Equivalently,
\[
    \norm{\mb u-\mb x}{}
    -
    \norm{\mb x}{}
    +
    \innerprod{\Pi_{\mathbb S}(\mb x)}{\mb u}
    \ge
    \frac{
    \norm{
    \left(
    \mb I_d-\Pi_{\mathbb S}(\mb x)\Pi_{\mathbb S}(\mb x)^\top
    \right)\mb u
    }{}^{2}
    }{
    2\left(
    \norm{\mb x}{}+\norm{\mb u}{}
    \right)
    }.
\]
\end{lemma}

\begin{proof}
Write \(\mb s:=\Pi_{\mathbb S}(\mb x)\) for short. We have
\begin{align}
    \norm{\mb u-\mb x}{}
    -
    \norm{\mb x}{}
    +
    \innerprod{\mb s}{\mb u}
    &=
    \frac{
        \norm{\mb u-\mb x}{}^2
        -
        \left(
            \norm{\mb x}{}-\innerprod{\mb s}{\mb u}
        \right)^2
    }{
        \norm{\mb u-\mb x}{}
        +
        \norm{\mb x}{}
        -
        \innerprod{\mb s}{\mb u}
    }
    \nonumber\\
    &=
    \frac{
        \norm{\mb u}{}^2
        -
        \innerprod{\mb s}{\mb u}^2
    }{
        \norm{\mb u-\mb x}{}
        +
        \norm{\mb x}{}
        -
        \innerprod{\mb s}{\mb u}
    } &&\text{by }\mb x=\norm{\mb x}{}\mb s
    \nonumber\\
    &=
    \frac{
        \mb u^\top
        \left(
            \mb I_d-\mb s\mb s^\top
        \right)\mb u
    }{
        \norm{\mb u-\mb x}{}
        +
        \norm{\mb x}{}
        -
        \innerprod{\mb s}{\mb u}
    }.\label{eq:expan_proj_s}
\end{align} 
By triangle inequality and Cauchy-Schwarz inequality,
\begin{align*}
    \norm{\mb u-\mb x}{} +
    \norm{\mb x}{} - \innerprod{\mb s}{\mb u}
    &\leq \norm{\mb u}{}+\norm{\mb x}{}+\norm{\mb x}{}+\norm{\mb s}{}\norm{\mb u}{} =2\left(\norm{\mb u}{}+\norm{\mb x}{}\right).
\end{align*}
Last step comes from $\norm{\mb s}{}=1$.
We thus finish the proof by plugging above inequality back to \eqref{eq:expan_proj_s}:\begin{align}
        \norm{\mb u-\mb x}{}
    -
    \norm{\mb x}{}
    +
    \innerprod{\mb s}{\mb u}
    \geq
    \frac{
        \mb u^\top
        \left(
            \mb I_d-\mb s\mb s^\top
        \right)\mb u
    }{
        2\left(
            \norm{\mb x}{}+\norm{\mb u}{}
        \right)
    }
\end{align} 
and noting that $\mb u^\top\left(\mb I_d-\mb s\mb s^\top\right)\mb u \ge 0$.

\end{proof}

The following lemma is used to prove \eqref{eq:gm-curvature-emp}. 
\begin{lemma} 
\label{lem:dimension-free-curvature}
For each $k\in[K]$, let
\[
    \chi_k\in\{0,1\},
    \qquad
    \mb v_k\in\mathbb R^d,
    \qquad
    \norm{\mb v_k}{}\le1,
\]
where the pairs \((\chi_k,\mb v_k)\) are independent across \(k\).
Define
\[
    \mb\Xi_k
    :=
    \chi_k
    \left(
        \mb I_d-\mb v_k\mb v_k^\top
    \right).
\]
Suppose that, for some $\lambda_\Xi>0$,
\[
    \E\mb\Xi_k
    \succeq
    \lambda_\Xi\mb I_d,
    \qquad
    k\in[K].
\]
If $K\ge 64\lambda_\Xi^{-2},$
then
\[
    \P\left[
        \lambda_{\min}
        \left(
            \frac1K\sum_{k=1}^K\mb\Xi_k
        \right)
        \ge
        \frac{\lambda_\Xi}{8}
    \right]
    \ge
    1-
    4\exp\left(
        -\frac{\lambda_\Xi^2K}{128}
    \right).
\]
\end{lemma}

\begin{proof}
We begin the proof by defining
\[
    \mb P_k
    :=
    \chi_k\mb v_k\mb v_k^\top,
    \qquad
    p_k:=\E\chi_k,
    \qquad
    \mb\Sigma_k:=\E\mb P_k,
\]
and thus let
\[
    \bar\chi
    :=
    \frac1K\sum_{k=1}^K\chi_k,
    \qquad
    \bar p
    :=
    \frac1K\sum_{k=1}^Kp_k,
    \qquad
    \bar{\mb P}
    :=
    \frac1K\sum_{k=1}^K\mb P_k.
\]
Since
\[
    p_k\mb I_d-\mb\Sigma_k
    =
    \E\mb\Xi_k
    \succeq
    \lambda_\Xi\mb I_d,
\]
we have
\begin{align}
    p_k\ge\lambda_\Xi,
    \qquad
    \mb\Sigma_k
    \preceq
    (p_k-\lambda_\Xi)\mb I_d,
    \qquad
    \tr(\mb\Sigma_k)\le p_k.
    \label{eq:curvature-pop-consequences}
\end{align}
Moreover, since
\begin{align}
     \lambda_{\min}
    \left(
    \frac1K\sum_{k=1}^K\mb\Xi_k
    \right) =\lambda_{\min}\left(\bar\chi\mb I_d-\bar{\mb P}\right)  =
    \bar\chi  - \lambda_{\max}(\bar{\mb P})
     \ge
    \bar\chi
    -
    \norm{\bar{\mb P}}{\mathrm F},
    \label{eq:curvature_frob_reduction}
\end{align}
the proof is finished by bounding from below $\bar\chi$ and bounding from above  $\norm{\bar{\mb P}}{F}$, respectively. Since Hoeffding inequality yields:
\begin{align}
    \P[\abs{\bar\chi-\bar p}<t]> 1-2\exp(-\frac{Kt^2}{2}),
\end{align}
we obtain: 
\begin{align}
    \bar\chi> \bar p-t,
\end{align}
with probability greater than $1-\exp(-Kt^2/2)$. On the other hand, by independence across \(k\),
\begin{align}
    \E\norm{\bar{\mb P}}{\mathrm F}^2
    &=
    \frac1{K^2}
    \sum_{k=1}^K
    \E\tr(\mb P_k^2)
    +
    \frac1{K^2}
    \sum_{k\neq\ell}
    \tr(\mb\Sigma_k\mb\Sigma_\ell)
    \nonumber\\
    &\leq\frac{1}{K^2}\sum_k^{K}\E\chi_k^2+\frac{1}{K^2}\sum_{k\neq l}\lambda_{\max}(\bs\Sigma_k)\tr(\bs\Sigma_l)\nonumber\\ 
    &\le
    \frac{\sum_{k=1}^K p_k}{K^2}
    +
    \frac1{K^2}
    \sum_{k\neq\ell}
    (p_k-\lambda_\Xi)p_\ell
    \nonumber\\
    &\le
    \frac{\bar p}{K}
    +
    \bar p(\bar p-\lambda_\Xi).
    \label{eq:curvature-frob-second-moment}
\end{align}
Here we use $\tr(\mb P_k^2)\le\chi_k^2\leq\chi_k$,
and $    \tr(\mb\Sigma_k\mb\Sigma_\ell)
    \le
    \lambda_{\max}(\mb\Sigma_k)
    \tr(\mb\Sigma_\ell)$ in the second step.
Combining Jensen's inequality and \eqref{eq:curvature-frob-second-moment} gives
\begin{align}
    \E\norm{\bar{\mb P}}{\mathrm F}
    &\le\sqrt{\E\norm{\bar{\mb P}}{\mathrm F}^2} \leq
    \sqrt{
        \bar p(\bar p-\lambda_\Xi)
    }
    +
    \sqrt{\frac{\bar p}{K}}.
    \label{eq:curvature-frob-mean}
\end{align}
Also, since replacing one $\mb P_k$ changes
$\norm{\bar{\mb P}}{\mathrm F}$ by at most $2/K$, an application of the bounded-difference
inequality \citep{mcdiarmid1989bounded} gives 
\[
    \P\left(
        \abs{\norm{\bar{\mb P}}{\mathrm F}-\E\norm{\bar{\mb P}}{\mathrm F}}
        >
        t
    \right)
    \le
    2\exp\left(-\frac{K t^2}{2}\right).
\]
Take the union bound of events $\{\norm{\bar{\mb P}}{\mathrm F}
        >
        \E\norm{\bar{\mb P}}{\mathrm F}+t\}$ and $\{\bar\chi>\bar p-t\}$ and combine \eqref{eq:curvature_frob_reduction} we obtain
        \begin{align}
            \lambda_{\min}\left(\frac{1}{K}\sum_{k=1}^K\Xi_k\right)&\geq \bar p-t-\left(\sqrt{\bar p\left(\bar p-\lambda_{\Xi}\right)}+\sqrt{\frac{\bar p}{K}}+t\right)\nonumber\\
            &>\bar p-\sqrt{\bar p\left(\bar p-\lambda_{\Xi}\right)}-\sqrt{\frac{1}{K}}-2t.
        \end{align}
with probability exceeding $1-4\exp(-Kt^2/2)$.
We finish the proof by taking $t=\lambda_{\Xi}/8$, using $K\geq 64\lambda_{\Xi}^{-2}$ and
\begin{align}
    \bar p-\sqrt{\bar p\left(\bar p-\lambda_{\Xi}\right)}=\frac{\lambda_{\Xi}\bar p}{\bar p+\sqrt{\bar p\left(\bar p-\lambda_{\Xi}\right)}}>\frac{\lambda_{\Xi}\bar p}{2\bar p}=\frac{\lambda_{\Xi}}{2}.
\end{align}
\end{proof}

\section{Extension to approximate solutions}\label{app_sec_approx}

\subsection{Extension to approximate $k$-means solutions}\label{sec:appro_kmeans}

The non-convex nature of the $k$-means objective function makes computing its exact solution computationally intractable due to the problem’s high time complexity. This challenge has driven the development of approximation methods, such as the $(1+\eps)$-approximation scheme in \cite{kumar2004simple} for any desired accuracy $\eps$. %
Formally, for any $\eps \ge 0$, the cluster centers $\{\wt\btheta^{(\eps)}_{ i}\}_{i \in [r]}$ and the group assignment map $\wt{\sigma}^{(\eps)}:\cI \to [r]$ corresponding to the $(1+\eps)$-approximate solution is defined as   
\begin{align}\label{def:appro_kmeans}
    \sum_{(k,i)\in \mathcal I}\bigl\|\bq_{(k,i)}-\wt{\btheta}^{(\eps)}_{\wt{\sigma}^{(\eps)}(k, i)}\bigr\|^2\leq \left(1+\epsilon\right)\sum_{(k,i)\in \mathcal I}
    \bigl\|\bq_{(k,i)}-\wt{\bs\theta}_{\wt\sigma(k,i)}\bigr\|^2.
\end{align}  
where we recall that  $\wt{\bs\theta}_1,\ldots,\wt{\bs\theta}_r$ and $\wt\sigma$ represent the  optimal solution of the $k$-means problem  in \eqref{kmeans_spec}.

Recall the within-cluster misclustering rate  $s_a(\cdot)$ given in \eqref{def:s_max} for $a\in [r]$. The following lemma states that the same bound for $s_a(\wt \sigma)$ in \eqref{up_bd_sm} holds for $s(\wt{\sigma}^{(\eps)})$ under a slightly stronger condition than \eqref{ieq:cond_epsilon}.
\begin{lemma}\label{lem:k_means_appro}
   Under conditions   \eqref{ass_orthogonal} and \eqref{ass_cover}. With $c>0$ given in \eqref{ieq:cond_epsilon}, if  
    \begin{equation}\label{ieq:cond_epsilon_approx}
    \epsilon \le  {c\over 1+\eps/2} {n_{\min}^2  \over \bar n n_{\max}}
    \end{equation}
    holds, then the conclusion in \eqref{up_bd_sm} holds Then  
    \begin{align}\label{up_bd_sm_appro}
         s_a\bigl(\wt{\sigma}^{(\eps)}\bigr) ~ \le ~  8\eta_a  ~ (\bar n / n_{\min}),\qquad \forall ~ a\in [r].
    \end{align}
\end{lemma}
\begin{proof}
Denote by $\wh{\sigma}^{(\eps)}$  the $(1+\eps)$-approximation solution to the k-means problem on $\mb M^{(r)}$; see, \cref{def:hat_sigma}. Due to the equivalence between   $\wh{\sigma}^{(\eps)}$ and $\wt{\sigma}^{(\eps)}$ (see \cref{lem:rank_kmeans}), we focus on analyzing $\wh{\sigma}^{(\eps)}$ by repeating similar arguments in \cref{app_sec_proof_lem:kmeans}. 

Let $\wh \cC_a^{(\eps)}$, $a\in [r]$, be the clusters induced by $\wh{\sigma}^{(\eps)}$. Denote the cluster center $\wh{\mb c}_a^{(\eps)}$ as 
\begin{align}
        \wh{\mb c}_a^{(\eps)}={1\over |\wh \cC_a^{(\eps)}|}  \sum_{(k,i)\in \wh \cC_a^{(\eps)}} \mb m^{(r)}_{(k,i)}. 
    \end{align}
   We first prove that for any $b\in [r]$,
   \begin{align}\label{target_approx}
        \min_{a\in [r]} \|\wh{\mb c}_a^{(\eps)}-\mb m_b^\star\|\leq 2\sqrt{1+\frac{\eps}{2}}\sqrt{\frac{\bar n}{n_b}}\sqrt{\eta}.
    \end{align}
    This is similar to \eqref{rate_center_raw}. To this end,  suppose \eqref{target_approx} does not hold. Then there is a  contradiction by observing (using similar arguments in \eqref{ieq:upper_sum_distance}) 
    \begin{align*}
        \sum_{(k,i)\in C_b^\star}\min_{a\in [r]}\|\mb m^{(r)}_{(k,i)}  -  \wh{\mb c}_a^{(\eps)}\|^2> (1+\eps)\eta\bar n, 
    \end{align*}
    as well as  
    \begin{align*}
         \sum_{(k,i)\in C_b^\star}\min_{a\in [r]}\|\mb m^{(r)}_{(k,i)}  -  \wh{\mb c}_a^{(\eps)}\|^2 
        &\leq \left(1+\eps\right)\sum_{(k,i)\in \mathcal I}  \min_{a\in [r]} \|\mb m^{(r)}_{(k,i)}  -  \wh{\mb c}_a\|^2 &&\text{by \eqref{def:appro_kmeans} }\\
        &\leq \left(1+\eps\right) \bar n\eta.
    \end{align*}
    The last step uses the optimality of the exact k-means solution and the same arguments in \eqref{ieq:prof_lem:k_means}. We have thus established \eqref{target_approx}. 
    
    Finally, by observing that the display \eqref{ieq:lower_devi_node_center} continues to hold under 
    \[
        \eta \le {1\over 128(1+\eps/2)} {n_{\min}^2\over \bar n^2}
    \]
    which, by \eqref{ieq:eta_upperbound} and \eqref{ieq:cond_epsilon}, is ensured by 
    \eqref{ieq:cond_epsilon_approx}, repeating the same arguments after \eqref{ieq:lower_devi_node_center} completes the proof.
\end{proof}

Let $\wh\bA^{(\eps)}$  be the output of \cref{alg:1} using the $(1+\eps)$-approximate solution in place of the exact k-means solution.  The following theorem states its estimation error bound is the same for the one using the exact k-means solution. Recall the definition of $p_a$ in \eqref{def:optimal_p_i} and $\delta_{\epsilon}$ in \eqref{def_delta_eps}.  
\begin{theorem}\label{thm:recover_appro_kmeans}
Assume   \eqref{ass_orthogonal} and \eqref{ass_cover}. For any $\eps \ge 0$, grant condition  \eqref{ieq:cond_epsilon_approx}. 
For any $a\in [r]$, assume 
     $Q_a(p_a) < c_0 \sqrt{n_{\min} / \bar n}$. Then there exists 
    some  permutation $\pi:[r]\to [r]$ such that  
    \begin{align}\label{rate_propose_kmeans_appro}
         \min_{\xi\in \{\pm1\}}\bigl\|\wh\ba_a^{(\eps)} - \xi\mb a^\star_{\pi(a)}\bigr\|  ~ \leq ~   \frac{2 p_a}{2 p_a - 1 - \delta_\epsilon} Q_a(p_a),\qquad \forall ~ a\in [r].
    \end{align}  
\end{theorem} 

The only price to pay for using a $(1+\eps)$-approximate solution to the $k$-means problem is the slightly stronger condition \eqref{ieq:cond_epsilon_approx}. Since $\eps$ can be made arbitrarily small, this price is very mild.

\subsection{Extension to approximate geometric median solutions}\label{sec:approx_kmedian}

For $\gamma\ge 0$, we consider a $(1+\gamma)$-approximation solution to the geometric median problem in \eqref{def_GM}, that is, for each $a\in [r]$,
\begin{equation}\label{def_GM_approx}
  \sum_{(k,i)\in\wt\cC_a}
    \bigl\|
        \wh\ba_a^{(\gamma)}-\widecheck{\mb a}^{(k)}_i
    \bigr\| 
    ~ \le  ~ 
    (1+\gamma) \min_{\mb q\in\RR^r}
    \sum_{(k,i)\in\wt\cC_a}
    \bigl\|
        \mb q-\widecheck{\mb a}^{(k)}_i
    \bigr\|.
\end{equation}
The following theorem is an analogue of \cref{thm:recover} that states the estimation error bound for the aggregated estimator based on $(1+\gamma)$-approximate GM solution. We see that it has an additive effect on $\gamma$, which is then multiplied by the estimation error of each column of $\bar\bA$.

\begin{theorem}\label{thm:recover_appro_kmedian}
Under the same conditions of \cref{thm:recover}, for any $\gamma\ge 0$,  there exists some  permutation $\pi:[r]\to [r]$ such that for any $a\in [r]$,
    \begin{align}\label{rate_propose_GM_appro}
         \min_{\xi\in \{\pm1\}}\bigl\|\wh\ba_a^{(\gamma)} - \xi\mb a^\star_{\pi(a)}\bigr\|  ~ \leq ~   \frac{2 p_a}{2 p_a - 1 - \delta_\epsilon} Q_a(p_a) + \frac{(1+\delta_\epsilon) \gamma}{(2 p_a-1-\delta_\epsilon)} \left( \sqrt{\epsilon_a} +\frac{\bar{n}n_{\max}}{n_{\min}^2}\epsilon\right).
    \end{align} 
\end{theorem}
\begin{proof}
    Pick any $a\in [r]$. Without loss of generality, we assume $\pi$ to be identity. 
    Using the bound for a  $(1+\gamma)$ approximation of the geometric median in \citet[Lemma 1]{chen2017distributed}, we have  
    \begin{align*}
        \|\wh{\mb a}_a^{(\gamma)}-\varsigma_a\mb a^{\star}_{a}\|&\leq \inf_{p\in (1/2, 1]}\left\{\frac{2p}{2p-1}\widecheck {Q}_a(p)+\frac{\gamma}{(2p-1)}\frac{1}{|\wt \cC_a|}\sum_{(k,i)\in \wt \cC_a}\widecheck\epsilon_i^{(k)}\right\}\nonumber\\
        &\leq \frac{2\bar p_a}{2\bar p_a-1}\widecheck{Q}_a(\bar p_a)+\frac{\gamma}{(2\bar p_a-1)}\frac{1}{|\wt \cC_a|}\sum_{(k,i)\in \wt \cC_a}\widecheck\epsilon_i^{(k)}
    \end{align*}
    for any $\bar p_a\leq p_a/(1+\delta_\epsilon)$. By the proof of \cref{thm:recover} and choosing $\bar p_a=p_a/(1+\delta_\epsilon)$, one has 
    \[
            \frac{2\bar p_a}{2\bar p_a-1}\widecheck{Q}_a(\bar p_a) \le \frac{2p_a}{2p_a-1-\delta_\epsilon}Q_a(p_a).
    \]
    It remains to bound from above 
    \[
        \frac{\gamma(1+\delta_\epsilon)}{(2 p_a-1-\delta_\epsilon)}\frac{1}{|\wt \cC_a|}\sum_{(k,i)\in \wt \cC_a}\widecheck\epsilon_i^{(k)}.
    \]
    This can be done by arguments for bounding \eqref{bd_bar_a_diff} in \cref{thm:kmeans_center}. Invoking \eqref{bd_bar_a_diff_final} gives  
    \begin{align*}
        \frac{1}{|\wt \cC_a|}\sum_{(k,i)\in \wt \cC_a}\widecheck\epsilon_i^{(k)} &\lesssim \sqrt{\epsilon_a} +\frac{\bar{n}n_{\max}}{n_{\min}^2}\epsilon.
    \end{align*}
    Combining both terms gives the desired result. 
    \end{proof}

\end{document}